\documentclass[msom]{informs3} 

\OneAndAHalfSpacedXII 
\usepackage{graphicx,caption,subcaption}

\usepackage{enumitem} 
\usepackage{soul}
\usepackage{amsfonts}
\usepackage{enumerate,verbatim}
\usepackage{algpseudocode,algorithm}
\usepackage{hyperref}       
\usepackage{url}            
\usepackage{multirow} 
\usepackage{wrapfig}
\usepackage{comment}
\usepackage{xcolor}

\newcommand{\ask}[1]{{\color{purple} [{#1}]}}

\usepackage{natbib}
 \bibpunct[, ]{(}{)}{,}{a}{}{,}%
 \def\bibfont{\small}%
\EquationsNumberedThrough    

\usepackage{ifthen}
\def\coralreport{0}

\ifthenelse{\coralreport=1}{
	\usepackage{isetechreport}
	\def\reportyear{18T} 	
	\def\reportno{005}		
	\def\revisionno{0}
	\def\originaldate{\today}
	\coraltrue
	\cvcrfalse
}{}

\MANUSCRIPTNO{MSOM-20-051.R1}
\begin{document}




\TITLE{A Deep Q-Network for the Beer Game: Deep Reinforcement Learning for Inventory Optimization}

\ARTICLEAUTHORS{
	\AUTHOR{
		Afshin Oroojlooyjadid, MohammadReza Nazari, Lawrence V.~Snyder, Martin Tak\'a\v{c}
	}%
\AFF{Department of Industrial and Systems Engineering,
	Lehigh University,
	Bethlehem, PA 18015, \EMAIL{ oroojlooy@gmail.com, \{mon314, larry.snyder\}@lehigh.edu}, Takac.MT@gmail.com}
} 

\ABSTRACT{%
{\it Problem definition:} The beer game is widely used in supply chain management classes to demonstrate the bullwhip effect and the importance of supply chain coordination. The game is a decentralized, multi-agent, cooperative problem that can be modeled as a serial supply chain network in which agents choose order quantities while cooperatively attempting to minimize the network's total cost, even though each agent only observes local information. 
\noindent {\it Academic/practical relevance:} Under some conditions, a base-stock replenishment policy is optimal. However, in a decentralized supply chain in which some agents  act irrationally, there is no known optimal policy for an agent wishing to act optimally. 
\noindent {\it Methodology:} We propose a deep reinforcement learning (RL) algorithm 
to play the beer game. Our algorithm makes no assumptions about costs or other settings. Like any deep RL algorithm, training is computationally intensive, but once trained, the algorithm executes in real time. We propose a transfer-learning approach so that training performed for one agent can be adapted quickly for other agents and settings.
\noindent {\it Results:} When playing with teammates who follow a base-stock policy, our algorithm obtains near-optimal order quantities. More importantly, it performs significantly better than a base-stock policy when other agents use a more realistic model of human ordering behavior. We observe similar results using a real-world dataset. 
Sensitivity analysis shows that a trained model is robust to changes in the cost coefficients. Finally, applying transfer learning reduces the training time by one order of magnitude.
\noindent {\it Managerial implications:} This paper shows how artificial intelligence can be applied to inventory optimization. Our approach can be extended to other supply chain optimization problems, especially those in which supply chain partners act in irrational or unpredictable ways.
Our RL agent has been integrated into a new online beer game, which has been played over 17,000 times by over 4,000 people.
}


\HISTORY{}

\maketitle


\AAnormalsizeXI

\section{Introduction}\label{sec:beer_introduction}

The beer game consists of a serial supply chain network with four agents---a retailer, a warehouse, a distributor, and a manufacturer---who must make independent replenishment decisions with limited information. The game is widely used in classroom settings to demonstrate the {\em bullwhip effect}, a phenomenon in which order variability increases as one moves upstream in the supply chain, as well as the importance of communication and coordination in the supply chain. The bullwhip effect occurs for a number of reasons, some rational \citep{LePaWh97ms} and some behavioral \citep{sterman1989modeling}. It is an inadvertent outcome that emerges when the players try to achieve the stated purpose of the game, which is to minimize costs. In this paper, we are interested not in the bullwhip effect but in the stated purpose, i.e., the minimization of supply chain costs, which underlies the decision making in every real-world supply chain. For general discussions of the bullwhip effect, see, e.g., \citet{LePaWh04,GeDiTo06}, and \citet{snyder2018fundamentals}. 

\begin{figure}
	\FIGURE
	{\includegraphics[scale=0.35]{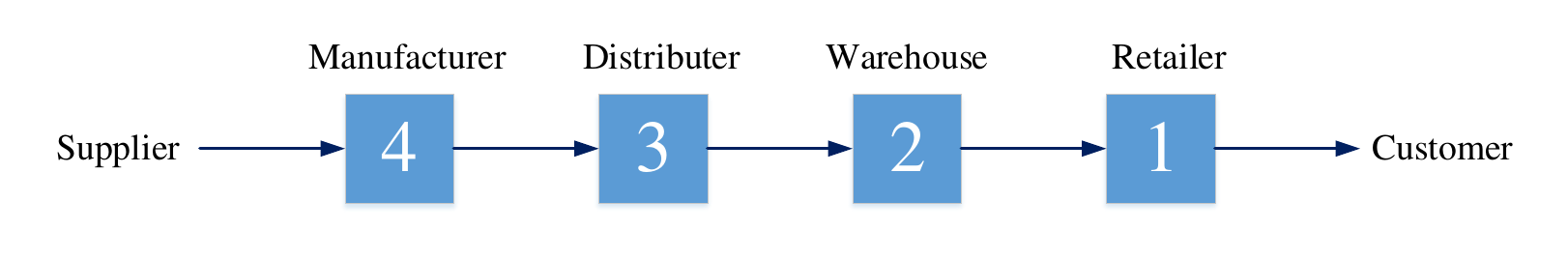}}
	{Generic view of the beer game network.\label{fig:beer_game}}	
	{}
\end{figure}

The agents in the beer game are arranged sequentially and numbered from 1 (retailer) to 4 (manufacturer), respectively. (See Figure \ref{fig:beer_game}.) 
The retailer node faces stochastic demands from its customer, and the manufacturer node has an unlimited source of supply. 
There are deterministic transportation lead times ($l^{tr}$) imposed on the flow of product from upstream to downstream, though the actual lead time is stochastic due to  stockouts upstream; there are also deterministic information lead times ($l^{in}$) on the flow of information from downstream to upstream (replenishment orders). 
Each agent may have nonzero shortage and holding costs. 

In each period of the game, each agent chooses an order quantity $q$ to submit to its predecessor (supplier) in an attempt to minimize the long-run system-wide costs, 
\begin{equation}
\sum_{t=1}^{T} \sum_{i=1}^{4} c^i_h (IL_t^i)^{+} + c^i_p (IL_t^i)^{-},
\label{eq:beer_cost}	 	
\end{equation}
where $i$ is the index of the agents; $t=1,\dots,T$ is the index of the time periods; $T$ is the time horizon of the game (which is often unknown to the players); $c^i_h$ and $c^i_p$ are the holding and shortage cost coefficients, respectively, of agent $i$; and $IL_t^i$ is the inventory level of agent $i$ in period $t$. If $IL_t^i > 0$, then the agent has inventory on-hand, and if $IL_t^i < 0$, then it has backorders. 
The notation $x^+$ and $x^-$ denotes $\max\{0,x\}$ and $\max\{0,-x\}$, respectively.

The standard rules of the beer game dictate that the agents may not communicate in any way, and that they do not share any local inventory statistics or cost information with other agents until the end of the game, at which time all agents are made aware of the system-wide cost. 
As a result, according to the categorization by \cite{claus1998dynamics}, the beer game is a decentralized, independent-learners (ILs), multi-agent, cooperative problem.

The beer game assumes the agents incur holding and stockout costs but not fixed ordering costs, and therefore the optimal inventory policy is a {\em base-stock policy} in which each stage orders a sufficient quantity to bring its inventory position (on-hand plus on-order inventory minus backorders) equal to a fixed number, called its base-stock level \citep{clark1960optimal}. 
When there are no stockout costs at the non-retailer stages, i.e., $c_p^i = 0,~i \in \{2,3,4\}$, the well known algorithm by  \cite{clark1960optimal} 
provides the optimal base-stock levels. 
To the best of our knowledge, there is no algorithm to find the optimal base-stock levels for general stockout-cost structures. 
\emph{More significantly}, when some agents do not follow a base-stock or other rational policy (as is typical in the beer game), 
the form and parameters of the optimal policy that a given agent should follow are unknown. 

In this paper, we propose a data-driven algorithm that is an extension of deep Q-networks (DQN) \citep{mnih2015human} to solve this problem. Because reward shaping is a key part of our approach, we refer to our algorithm as the {\em shaped-reward DQN} (SRDQN) algorithm. The SRDQN algorithm is customized for the beer game, but we view it also as a proof-of-concept that RL can be used to solve messier, more complicated supply chain problems than those typically analyzed in the literature. 

Like most machine learning (ML) algorithms, the training of the SRDQN algorithm can be slow, and our approach needs training data or the probability distribution of the demand. Moreover, our approach offers no theoretical guarantee of global optimality. On the other hand, once trained, our algorithm executes nearly instantaneously; it attains very good cost performance; and (as we show) it can be trained using small data sets. Moreover, our method is quite flexible and can be applied even in settings for which optimal inventory policies are unknown.

Our contributions can be summarized as follows. (See Section~\ref{sec:beer_our_contribution} for further discussion.)
\begin{itemize}
\item We present the first RL-based algorithm for the beer game that observes the game's typical rules, in particular, decentralized information and decision-making.
\item By introducing reward shaping, we adapt the classical DQN algorithm, which applies to competitive zero-sum games, to enable it to play the beer game, which is a cooperative non-zero-sum game. 
\item SRDQN is a data-driven algorithm and does not need the knowledge of demand distribution. 
\item We demonstrate numerically that the SRDQN agent is capable of learning a near-optimal policy when its co-players (teammates) act rationally, i.e.,  follow a base-stock policy.
\item More importantly, when the co-players are irrational, i.e., do not follow a base-stock policy---which mimics how humans play the beer game and is also common in actual supply chains---we show that the SRDQN agent obtains much lower costs than an agent that uses a base-stock policy, even with optimized base-stock levels. 
\item This suggests that a base-stock policy is not optimal for a stage in a serial system with non-base-stock co-players, which is, to the best of our knowledge, a new result in the literature. 
\item We show that the SRDQN agent can be sufficiently trained based on relatively few (in our case, 100) demand observations.
\item We show that the trained model is quite robust against  changes in the shortage and holding costs.
\end{itemize}

Although it is outside the scope of this paper, our hope is to learn something about the structure of the optimal inventory policy with irrational co-players by examining how the SRDQN agent plays, much as human players are learning new chess strategies by studying the way deep RL chess agents such as AlphaZero play the game \citep{Chan2017}.

The remainder of this paper is as follows. Section \ref{sec:lit_review} provides a brief summary of the relevant literature and our contributions to it. 
The details of the algorithm are introduced in Section \ref{sec:beer_dqn_for_beer_game}. Section \ref{sec:beer_numerical_experiment} provides numerical experiments, and Section \ref{sec:beer_conclusion} concludes the paper.

\section{Literature Review}\label{sec:lit_review}

\subsection{Current State of Art}\label{sec:Current_State_of_Art} 	

The beer game consists of a serial supply chain network. Under the conditions dictated by the game,  
a base-stock policy is optimal at each stage \citep{LePaWh97ms}. If the demand process and costs are stationary, then so are the optimal base-stock levels, which implies that in each period (except the first), each stage simply orders from its supplier exactly the amount that was demanded from it. If the customer demands are i.i.d. random and if backorder costs are incurred only at stage 1, then the optimal base-stock levels can be found using the exact  algorithm by \cite{clark1960optimal}; see also \cite{ChenZheng1994,GallegoZipkin1999}. 
Nevertheless, for more general serial systems neither the form nor the parameters of the optimal policy are known. Moreover, even in systems for which a base-stock policy is optimal, such a policy may no longer be optimal for a given agent if the other agents do not follow it.

There is a substantial literature on the beer game and the bullwhip effect. 
We review some of that literature here, considering both independent learners (ILs) and joint action learners (JALs) \citep{claus1998dynamics}. (ILs have no information about the other agent, whereas JALs may share some information and take central decisions.). 
See \cite{martinez2014beergame} for a thorough history of the beer game. 

In the category of ILs, 
\cite{sterman1989modeling} proposes a formula (which we call the {\em Sterman formula}) to determine the order quantity based on the current backlog of orders, on-hand inventory, incoming and outgoing shipments, incoming orders, and expected demand. 
In a nutshell, the Sterman formula attempts to model the way human players over- or under-react to situations they observe in the supply chain such as shortages or excess inventory. Note that Sterman's formula is not an attempt to optimize the order quantities in the beer game; rather, it is intended to model typical human behavior. 
\citet{sterman1989modeling} uses a version of the game in which the demand is 4 for the first four periods, and is 8 after that. Hereinafter, we refer to this demand process as $C(4,8)$ or the {\em classic} demand process. Also, he does not allow the players to be aware of the demand process.

The optimization method described in the first paragraph of this section assumes that every agent follows a base-stock policy. The hallmark of the beer game, however, is that players do not tend to follow such a policy, or {\em any} policy. Often their behavior is quite irrational, as shown by \citep{sterman1989modeling}. There is comparatively little literature on how a given agent should optimize its inventory decisions when the other agents do not play rationally \citep{sterman1989modeling, strozzi2007beer}---that is, how an individual player can best play the beer game when her teammates may not be making optimal decisions---and there are, to the best of our knowledge, no theoretical proofs of policy optimality in this setting.

Some of the beer game literature assumes the agents are JALs, i.e., information about inventory positions is shared among all agents, a significant difference compared to classical IL models. For example, 
\cite{kimbrough2002computers} propose a GA that receives a current snapshot of each agent and decides how much to order according to the $d+x$ rule. In the $d+x$ rule, agent $i$ observes $d_t^i$, the received demand/order in period $t$, chooses $x_t^i$, and then places an order of size $a_t^i = d_t^i + x_t^i$. 

In the literature on multi-echelon inventory problems, there are only a few papers that use RL. Among them, \cite{jiang2009case} propose an RL for a two-echelon serial system, in which the retailer has to optimize $(r,Q)$ or $(T,S)$ policy parameters. 
The only decision in the model is for the retailers, so it is not truly a multi-echelon problem. Their state consists of the IL and the demand, and they discretize and truncate the state and action values to avoid the curse of dimensionality. \cite{gijsbrechts2019can} solve a one-warehouse, multi-retailer (OWMR) problem (in addition to two difficult single-echelon problems) using RL. They use the asynchronous advantage actor-critic (A3C) algorithm \citep{mnih2016asynchronous} to solve the problem, with promising results compared to classical methods. Their model considers a different supply chain structure from ours (OWMR vs.\ serial) and uses a different RL methodology.

We are aware of two papers that propose RL algorithms to play versions of the beer game. 
\cite{giannoccaro2002inventory} consider a beer game with three agents with stochastic shipment lead times and stochastic demand. 
They propose a classical tabular RL algorithm to make decisions, in which the state variable is defined as the three inventory positions, each of which are discretized into 10 intervals.
The agents may use any actions in the integers on $[0,30]$.	
\cite{chaharsooghi2008reinforcement} consider the same game and solution approach except with four agents and a fixed length of 35 periods for each game.
In their proposed RL, the state variable is the four inventory positions, which are each discretized into nine intervals.
Moreover, their RL algorithm uses the $d+x$ rule to determine the order quantity, with $x$ restricted to be in $\{0,1,2,3\}$. 

Note that both of these papers assume that real-time information is {\em shared} among agents,  and that there is a {\em centralized} decision-making agent, 
significant departures from the typical beer game setting. The setting considered in these two papers can be modeled as an MDP that is relatively easy to solve. In contrast, in the typical beer game setting, and the setting we consider, each agent only observes local information, which leads to a much more complex MDP and a much more challenging setting for an RL algorithm. In addition, these papers use a tabular RL algorithm, and they assume (implicitly or explicitly) that the size of the state space is small, which is unrealistic in the beer game, since the state variable representing a given agent's inventory level can be any number in $(-\infty, +\infty)$. Thus, they map the state variable onto a small number of tiles \citep{sutton1998reinforcement}, which leads to a loss of valuable state information and therefore of accuracy of the solution. We do not use any mapping or cutting of the state variables, which leads to a more challenging problem but is capable of attaining better accuracy.

Another possible approach to tackle this problem might be classical supervised machine learning algorithms. However, these algorithms also cannot be readily applied to the beer game, since there is no historical data in the form of ``correct'' input/output pairs. Thus, we cannot use a stand-alone support vector machine or deep neural network with a training dataset 
and train it to learn the best action
(as in the approach used by \cite{oroojlooyjadid2017stock, oroojlooyjadid2020applying, ban2019big, ban2019dynamic} to solve simpler supply chain problems such as newsvendor and dynamic procurement). 
Based on our understanding of the literature, there is a large gap between solving the beer game problem effectively and what the current algorithms can handle. 
In order to fill this gap, we propose a variant of the DQN algorithm, SRDQN, to choose the order quantities in the beer game.

\subsection{Reinforcement Learning}\label{sec:RL}

Reinforcement learning \citep{sutton1998reinforcement} is an area of machine learning that has  been successfully applied to solve complex sequential decision problems. RL is concerned with the question of how an agent should choose an action to maximize a cumulative reward.
RL is a popular tool in telecommunications, robot control, and game playing, to name a few (see \cite{li2017deep} for more applications and details). 

\begin{wrapfigure}{2}{0.45\textwidth}
	\FIGURE
	{\includegraphics[scale=0.35]{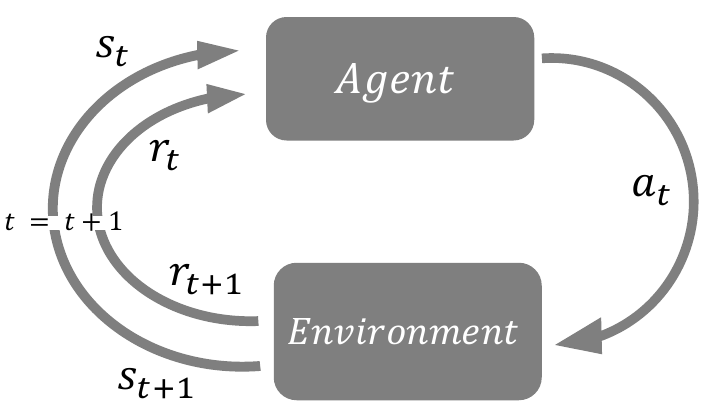}}
	{A generic procedure for RL. \label{fig:RL}}
	{}
\end{wrapfigure}

RL considers an agent that interacts with an environment. In each time step $t$, the agent observes the current state of the system, $s_t \in \mathcal{S}$ (where $\mathcal{S}$ is the set of possible states), chooses an action $a_t \in \mathcal{A}(s_t)$ (where $\mathcal{A}(s_t)$ is the set of possible actions when the system is in state $s_t$), and gets reward $r_t \in \mathbb{R}$; and then the system transitions randomly into state $s_{t+1} \in \mathcal{S}$. This procedure is known as a {\em Markov decision process} (MDP) (see Figure \ref{fig:RL}), and RL algorithms can be applied to solve this type of problem.

The matrix $P_a(s,s{'})$, which is called the {\em transition probability matrix}, provides the probability of transitioning to state $s'$ when taking action $a$ in state $s$, i.e., $P_a(s,s') = \Pr(s_{t+1}=s' \mid s_t = s, a_t=a)$.
Similarly, $R_{a} (s,s')$ defines the corresponding reward matrix. In each period $t$, the decision maker takes action $a_t = \pi_t(s_t)$ according to a given policy, denoted by $\pi_t$. 
The goal of RL is to maximize the expected discounted sum of the rewards $r_t$, when the systems runs for an infinite horizon, i.e. finding policy $\pi: \mathcal{S} \rightarrow \mathcal{A}$ that maximizes $E \left[ \sum_{t=0}^{\infty} \gamma^t R_{a_t} (s_t,s_{t+1})\right]$,
where $0 \le \gamma < 1$ is the discount factor.	
For given $P_a(s,s{'})$ and $R_{a} (s,s')$, the optimal policy can be obtained through dynamic programming 
or linear programming \citep{sutton1998reinforcement}. 

Another approach for solving this problem is {\em Q-learning}, 
a RL algorithm that obtains the {\em Q-value} for any $s \in S$ and $a = \pi(s)$, i.e., 
	\[ Q(s,a) = \mathbb{E} \left[r_{t} + \gamma r_{t+1} + \gamma^2 r_{t+2} + \dots \mid s_t = s, a_t = a; \pi \right]. \]
%
%
The Q-learning approach starts with an initial guess for $Q(s,a)$ for all $s$ and $a$ and then proceeds to update them based on the iterative formula
\begin{equation}
\label{eq:q-learning}
Q(s_t,a_t) = (1- \alpha_t)Q(s_t,a_t) + \alpha_t \left( r_{t+1} + \gamma \max \limits_{a} Q(s_{t+1},a)\right), \forall t = 1,2,\dots,
\end{equation}
where $\alpha_t$ is the learning rate at time step $t$. 
In each observed state, the agent chooses an action through an $\epsilon$-greedy algorithm with constantly decreasing $\epsilon$ to guarantee convergence to the optimal policy (see online supplement A
for more details). 
After finding optimal $Q^*$, one can recover the optimal policy as $\pi^* (s) = \argmax_a Q^*(s,a)$. 

Both of the  algorithms discussed so far (dynamic programming and Q-learning) guarantee that they will obtain the optimal policy. However, due to the curse of dimensionality, these approaches are not able to solve MDPs with large state or action spaces in reasonable amounts of time.
Many problems of interest (including the beer game) have large state and/or action spaces. 
Moreover, in some settings (again, including the beer game), the decision maker cannot observe the full state variable. This case, which is known as a {\em partially observed MDP} (POMDP), makes the problem much harder to solve than MDPs.

In order to solve large POMDPs and avoid the curse of dimensionality, it is common to approximate the Q-values in the Q-learning algorithm \citep{sutton1998reinforcement}. Linear regression is often used for this purpose \citep{li2017deep}; however, it is not powerful or accurate enough for our application. Non-linear functions and neural network approximators are able to provide more accurate approximations; on the other hand, they are known to provide unstable or even diverging Q-values due to issues related to non-stationarity and correlations in the sequence of observations.
The seminal work of \cite{mnih2015human} solved these issues by proposing {\em target networks} and utilizing {\em experience replay memory}. 
They proposed a {\em deep Q-network} (DQN) algorithm, which uses a deep neural network to obtain an approximation of the Q-function and trains it through the iterations of the Q-learning algorithm. This algorithm has been applied to many competitive games (see \cite{li2017deep}). 
Although DQN works well in practice, there has been no theoretical proof of convergence. This is mainly due to the non-convexity of the deep neural network and the fact that the DQN uses the \emph{argmax} function, which is non-smooth. In addition, the effect of the target network and the replay buffer memory on the distribution of the training samples and the gradients is unknown, and has not been studied yet. For more details, see \cite{yang2019theoretical}.

The beer game exhibits one characteristic that differentiates it from most settings in which DQN is commonly applied, namely, that there are multiple agents that cooperate in a decentralized manner to achieve a common goal. Such a problem is called a decentralized POMDP, or Dec-POMDP. Due to the partial observability and the non-stationarity of agents' local observations, Dec-POMDPs are hard to solve and are categorized as NEXP-complete problems (problems that can be solved in $2^{n^{\mathcal{O}(1)}}$ time) \citep{bernstein2002complexity}.


The beer game exhibits all of the complicating characteristics described above---large state and action spaces, partial state observations, and decentralized cooperation. Our algorithm aims to overcome these difficulties by making several adaptations to the standard DQN algorithm.

\subsection{Our Contribution}\label{sec:beer_our_contribution}


We propose a Q-learning algorithm, a data-driven algorithm for the beer game in which a DNN approximates the Q-function. Indeed, the general structure of our algorithm is based on the DQN algorithm \citep{mnih2015human}, although we modify it substantially, since DQN is designed for single-agent, competitive, zero-sum games and the beer game is a multi-agent, decentralized, cooperative, non-zero-sum game. 
In other words, DQN provides actions for one agent that interacts with an environment in a competitive setting, and the beer game is a cooperative game in the sense that all of the players aim to minimize the total cost of the system in a random number of periods. 
Also, beer game agents are playing independently and do not have any information from other agents until the game ends and the total cost is revealed, whereas DQN usually assumes the agent fully observes the state of the environment at any time step $t$ of the game. For example, DQN has been successfully applied to Atari games, but in these games the agent attempts to defeat an opponent and observes full information about the state of the system at each time step $t$.

Without shared information about rewards, applying DQN in a straightforward way to a given agent would result in a policy that optimizes that agent locally but will not result in an optimal solution for the supply chain as a whole.  (For a simple example, see online supplement 
D).
So, we propose a unified framework, SRDQN, in which the agents still play independently from one another, but in the training phase, we use reward shaping via a feedback scheme so that the SRDQN agent learns the total cost for the whole network and can, over time, learn to minimize it. 
Thus, the SRDQN agent in our model plays smartly in all periods of the game to get a near-optimal cumulative cost for any random horizon length. 

In principle, our framework can be applied to multiple SRDQN agents playing the beer game simultaneously on a team. However, to date we have designed and tested our approach only for a single SRDQN agent whose teammates are not SRDQNs, e.g., they are controlled by simple formulas or by human players. Enhancing the algorithm to train multiple SRDQN agents simultaneously and cooperatively is a topic of ongoing research. Note that multi-agent RL is significantly harder to solve than single-agent RL problems. This is due to the fact that the policy of each agent changes throughout the training, and as a result, the environment becomes non-stationary from the viewpoint of each individual agent. This violates the underlying assumption of MDP on the existence of a stationary distribution. Therefore, multi-agent problems need specialized algorithms and tricks to handle the non-stationarity. For more details, see \cite{zhang2019multi, oroojlooyjadid2019review}.


The proposed approach works very well when we tune and train the SRDQN for a given agent and a given set of game parameters (e.g., costs, lead times,  action spaces, etc.). 
However, our sensitivity analysis shows that a given trained model is fairly robust to changes in the parameters; that is, the model can be trained for one set of parameters and still perform well when playing under a (somewhat) different set of parameters. Of course, if one needs the best possible performance, in principle one needs to tune and train a new network, but this approach is time consuming.
	To avoid this, we propose using a {\em transfer learning} approach \citep{pan2010survey}  in which we transfer the acquired knowledge of one agent under one set of game parameters to another agent with another set of game parameters. 
	In this way, we decrease the required time to train a new agent by roughly one order of magnitude. 

To summarize, SRDQN is {\em a variant of the DQN algorithm for choosing actions in the beer game}. In order to attain near-optimal cooperative solutions, we develop {\em a feedback scheme as a communication framework}. The trained model is fairly robust to the changes of parameters, and finally, to simplify training agents with new settings, we use {\em transfer learning} to efficiently make use of the learned knowledge of trained agents.

\begin{wrapfigure}{2}{0.35\textwidth}
\vspace*{-10pt}
\caption{Screenshot of [{\em company name}] 
online beer game integrated with our SRDQN agent.}
\centering
\label{fig:opex_BG}
\includegraphics[width=0.3\textwidth]{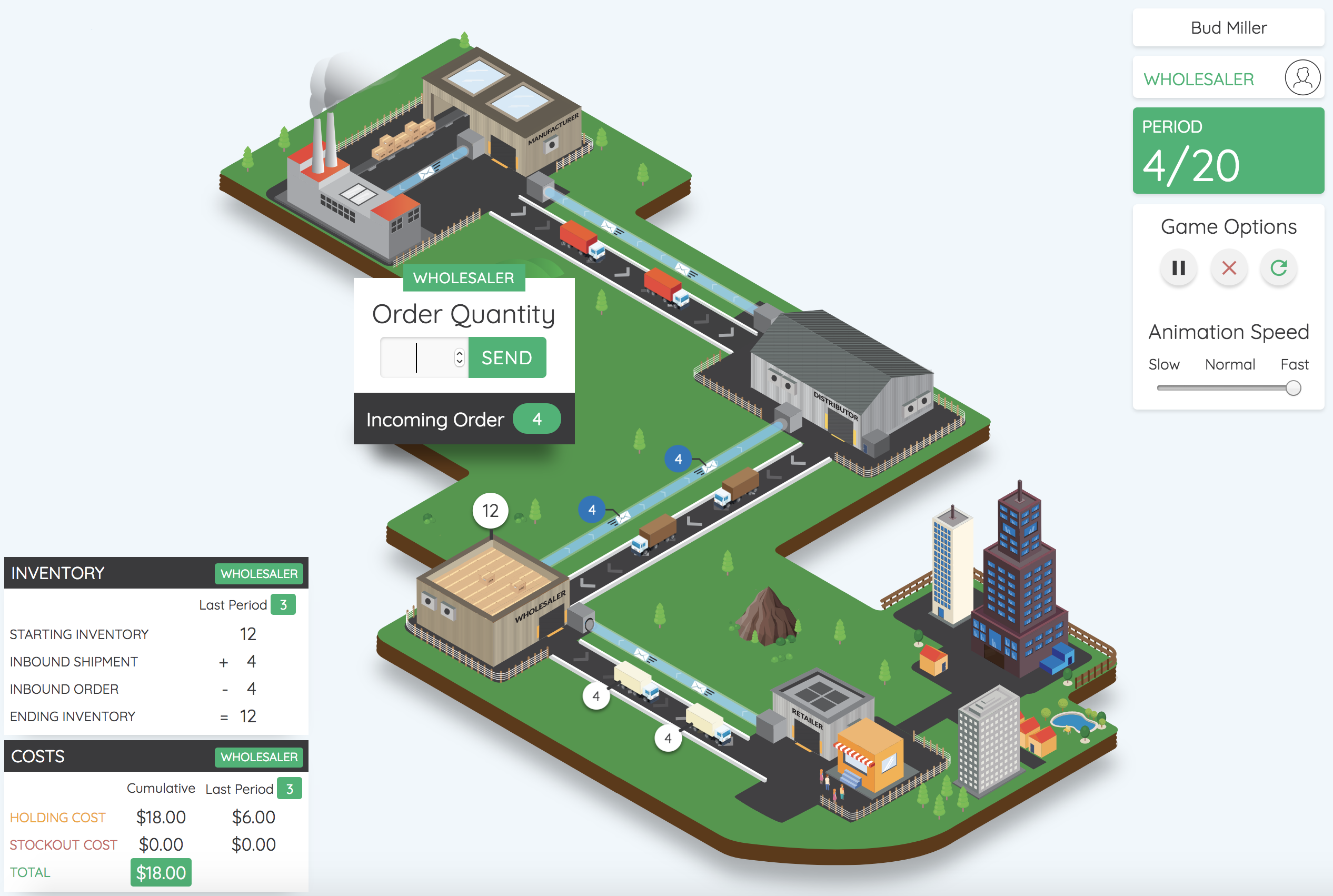}
\vspace*{-10pt}
\end{wrapfigure}

In addition to playing the beer game well, we believe our algorithm serves as a proof-of-concept that RL and other machine learning approaches can be used for real-time decision making in complex supply chain settings. 

We have integrated our algorithm into a new online beer game, developed by Opex Analytics (\url{http://beergame.opexanalytics.com/}); 
see Figure~\ref{fig:opex_BG}. 
The Opex beer game allows human players to compete with, or play on a team with, our SRDQN agent. As of this writing, the game has been played over 17,000 times by over 4,300 unique players from academia and industry. Over 1,400 of these game plays have included the SRDQN agent described in this paper.

Finally, we note that our open-source SRDQN and simulator code are available at \href{https://github.com/OptMLGroup/DeepBeerInventory-RL}{https://github.com/OptMLGroup/DeepBeerInventory-RL}.


\section{The SRDQN Algorithm}\label{sec:beer_dqn_for_beer_game}

In this section, we first present the details of our SRDQN algorithm to solve the beer game, and then describe the transfer learning mechanism.

\subsection{SRDQN: DQN with Reward Shaping for the Beer Game}

In our algorithm, an RL agent runs a Q-learning algorithm with DNN as the Q-function approximator to learn a semi-optimal policy with the aim of minimizing the total cost of the game. 
Each agent has access only to its local information and considers the other agents as parts of its environment. That is, the RL agent does not know any information about the other agents, including both static parameters such as costs and lead times, as well as dynamic state variables such as inventory levels. We propose a feedback scheme to teach the RL agent to work toward minimizing the system-wide cost, rather than its own local cost. The details of the scheme, Q-learning, state and action spaces, reward function, DNN approximator, and the algorithm are discussed below.

	\noindent {\bf State variables:} 
Consider agent $i$ in time step $t$. Let $OO_t^i$ denote the on-order items at agent $i$, i.e., the items that have been ordered from agent $i+1$ but not received yet; let $AO_t^i$ denote the size of the arriving order (i.e., the demand) received from agent $i-1$; let $AS_t^i$ denote the size of the arriving shipment from agent $i+1$; let $a_t^i$ denote the action agent $i$ takes; and let $IL_t^i$ denote the inventory level as defined in Section \ref{sec:beer_introduction}. We interpret $AO_t^1$ to represent the end-customer demand and $AS_t^4$ to represent the shipment received by agent 4 from the external supplier. 
In each period $t$ of the game, agent $i$ observes $IL_t^i$, $OO_t^i$, $AO_t^i$, and $AS_t^i$. In other words, in period $t$ agent $i$ has historical observations 
$o^i_t =\left [ \left((IL_1^i)^{+},(IL_1^i)^{-},OO_1^i,AO_1^i,RS_1^i), \dots, ((IL_t^i)^{+},(IL_t^i)^{-},,OO_t^i,AO_t^i,AS_t^i \right) \right].$
%
In addition, any beer game will finish in a finite time horizon, so the problem can be modeled as a POMDP in which each historic sequence $o^i_t$ is a distinct state and the size of the vector $o^i_t$ grows over time, which is difficult for any RL or DNN algorithm to handle.
To address this issue, we capture only the last $m$ periods (e.g., $m=3$) and use them as the state variable; thus the state variable of agent $i$ in time $t$ is
%
$s^i_t =\left[\left((IL_j^i)^{+},(IL_j^i)^{-},OO_j^i,AO_j^i,RS_j^i \right) \right]_{j=t-m+1}^t.$ See online supplement
C  
for more details about choosing the right value for $m$.

\noindent {\bf Action space:}  	
In each period of the game, each agent can order any amount in $[0,\infty)$. Since Q-learning needs a finite-size  action space, we limit the cardinality of the action space by using the $d+x$ rule for selecting the order quantity: The agent determines how much  more or less to order than its received order; that is, the order quantity is $d+x$, where $x \in [a_l , a_u]$ ($a_l, a_u \in {\mathbb Z}$).

\noindent {\bf DNN:}
In our algorithm, DNN plays the role of the Q-function approximator, providing the Q-value as the output for any pair of state $s$ and action $a$. 
Given the state and action spaces, our DNN outputs $Q(s,a)$ for every possible action $a \in \mathcal{A}$. (In the beer game, $\mathcal{A}(s)=[a_l , a_u]$ is the same for all $s$; therefore we simply use $\mathcal{A}$ to denote the action space). The DNN is trained using random mini-batches taken from the experience replay iteratively until all episodes are complete. (For more details about experience replay, see online supplement
E.)

\noindent {\bf Reward function:}
In iteration $t$ of the game, agent $i$ observes state variable $s^i_t$ and takes action $a^i_t$; one needs to know the corresponding reward value $r^i_t$ to measure the quality of action $a^i_t$. The state variable, $s^i_{t+1}$, allows us to calculate $IL^i_{t+1}$ and thus the corresponding shortage or holding costs, and we consider the summation of these costs for $r^i_t$.  
However, since there are information and transportation lead times, there is a delay between taking action $a^i_t$ and observing its effect on the reward. Moreover, the reward $r^i_t$ reflects not only the action taken in period $t$, but also those taken in previous periods, and it is not possible to decompose $r_t^i$ to isolate the effects of each of these actions. 
However, defining the state variable to include information from the last $m$ periods resolves this issue to some degree; the reward $r^i_t$ represents the reward of state $s^i_t$, which includes the observations of the previous $m$ steps.

On the other hand, the reward values $r^i_t$ are the intermediate rewards of each agent, and the objective of the beer game is to minimize the total reward of the game, 
$\sum_{i=1}^{4} \sum_{t=1}^{T} r_t^i,$ 
which the agents only learn after finishing the game. In order to add this information into the agents' experience, we use reward shaping through a feedback scheme.

\noindent {\bf Feedback scheme:}
	When an episode of the beer game is finished, all agents are made aware of the total reward. In order to share this information among the agents, we propose a penalization procedure in the training phase to provide feedback to the RL agent about the way that it has played.
Let $\omega=\sum_{i=1}^{4} \sum_{t=1}^{T} \frac{r_t^i}{T}$ and $\tau^i = \sum_{t=1}^{T} \frac{r_t^i}{T}$ be the average reward per period and the average reward of agent $i$ per period, respectively. 
%
After the end of each episode of the game (i.e., after period $T$), for each RL agent $i$ we update its observed reward in all $T$ time steps in the experience replay memory using $r_t^i = r_t^i + \frac{\beta_i}{3}(\omega - \tau^i)$, $\forall t \in \{1, \dots, T\}$,
where $\beta_i$ is a regularization coefficient for agent $i$. With this procedure, agent $i$ gets appropriate feedback about its actions and learns to take actions that result in  minimum total cost, not locally optimal solutions. 

\noindent {\bf The algorithm:} 
Our algorithm to get the policy $\pi$ to solve the beer game is provided in Algorithm \ref{alg:dqn_beer_game}. The algorithm, which is based on DQN \citep{mnih2015human}, finds weights $\theta$ of the DNN network to minimize the Euclidean distance between $Q(s,a;\theta)$ and $y_j$, where $y_j$ is the prediction of the Q-value that is obtained from target network $Q^{-}$ with weights $\theta^{-}$. 
Every $C$ iterations, the weights $\theta^{-}$ are updated by $\theta$. 
Moreover, the actions in each training step of the algorithm are obtained by an $\epsilon$-greedy algorithm, which is explained in online supplement 
A.
%

{\SingleSpacedXI
\begin{algorithm}[]
	\caption{SRDQN for Beer Game}
	\label{alg:dqn_beer_game}
	\begin{algorithmic}[1]
		\Procedure{SRDQN} {}
		\State Initialize Experience Replay Memory $E_i = \left[~\right],~ \forall i$
		\For{$Episode=1 : n$}		
		\State Reset $IL$, $OO$, $d$, $AO$, and $AS$ for each agent	
		\For{$t=1 : T$}			
		\For{$i=1 : 4$}
		\State $\text{With probability } \epsilon \text{ take random action } a_t$,
		\State otherwise set $a_t = \underset{a}{\text{argmin }} Q \left(s_t,a;\theta \right)$
		\State Execute action $a_t$, observe reward $r_t$ and state $s_{t+1}$ 
		\State Add $(s^i_t,a^i_t,r^i_t,s^i_{t+1})$ into $E_i$
		\State Get a mini-batch of experiences $(s_j,a_j,r_j,s_{j+1})$ from $E_i$ 
		\State Set $y_j = \begin{cases}
		r_j & \text{if it is the terminal state} \\
		r_j + \min \limits_{a} Q(s_{j+1},a;\theta^{-}) & \text{otherwise}
		\end{cases}$ 
		\State Run forward and backward step on the DNN with loss function $\left (y_j - Q \left (s_j,a_j;\theta \right) \right)^2$ 
		\State Every $C$ iterations, set $\theta^{-} = \theta$
		\EndFor		
		\EndFor	
		\State Run feedback scheme, update experience replay of each agent
		\EndFor	
		\EndProcedure
	\end{algorithmic}
\end{algorithm}
}

In the algorithm, in period $t$ agent $i$ takes action $a^i_t$, satisfies the arriving demand/order $AO^i_{t-1}$, observes the new demand $AO^i_t$, and then receives the shipments $AS^i_t$. This sequence of events, which is explained in detail in online supplement 
J, 
results in the new state $s_{t+1}$. Feeding $s_{t+1}$ into the DNN network with weights $\theta^{-}$ provides the corresponding Q-value for state $s_{t+1}$ and all possible actions. The action with the smallest Q-value is our choice. Finally, at the end of each episode, the feedback scheme runs and distributes the total cost among all agents. 

\noindent {\bf Evaluation procedure:} In order to validate our algorithm, we compare the results of SRDQN to those obtained using the optimal base-stock levels (when possible) by \cite{clark1960optimal},\footnote{Our implementation uses the version of the Clark--Scarf algorithm presented by \cite{ChenZheng1994,GallegoZipkin1999}, but we refer to it as the ``Clark--Scarf'' algorithm throughout.} 
 as well as models of human beer-game behavior by \cite{sterman1989modeling}. (Note that none of these methods attempts to do exactly the same thing as our method. The methods by \cite{clark1960optimal} optimizes the base-stock levels assuming all players follow a base-stock policy---which beer game players do not tend to do---and the formula by \cite{sterman1989modeling} models human beer-game play, but they do not attempt to optimize.) The details of the training procedure and benchmarks are described in Section \ref{sec:beer_numerical_experiment}.

	\subsection{Transfer Learning}\label{sec:transfer_learning}
	Transfer learning \citep{pan2010survey} has been an active and successful field of research in machine learning and especially in image processing.
	In transfer learning, there is a {\em source} dataset {\tt S} and a trained neural network to perform a given task, e.g. classification, regression, or decisioning through RL.
	Training such networks may take a few days or even weeks. 
	So, for similar or even slightly different {\em target} datasets {\tt T}, one can avoid training a new network from scratch and instead use the same trained network with a few customizations. 
	The idea is that most of the learned knowledge on dataset {\tt S} can be used in the target dataset with a small amount of additional training. 
This idea works well in image processing (e.g. \cite{rajpurkar2017chexnet}) and considerably reduces the training time.

	In order to use transfer learning in the beer game, assume there exists a source agent $i \in \{1,2,3,4\}$ with trained network $S_i$ (with a fixed size on all agents), parameters $P_1^i = \{|\mathcal{A}_1^j|, c^j_{p_1}, c^j_{h_1}\}$, observed demand distribution $D_1$, and co-player (i.e., teammate) policy $\pi_1$. The weight matrix $W_i$ contains the learned weights such that $W_i^q$ denotes the weight between layers $q$ and $q+1$ of the neural network, where $q \in \{0,\dots,nh\}$, and $nh$ is the number of hidden layers.
	The aim is to train a neural network $S_j$ for target agent $j \in \{1,2,3,4\}$, $j\ne i$. 
	We set the structure of the network $S_j$ the same as that of $S_i$, and initialize
	$W_j$ with $W_i$, making the first $k$ layers not trainable. 
	Then, we train neural network $S_j$ with a small learning rate.	Note that, as we get closer to the final layer, which provides the Q-values, the weights become less similar to agent $i$'s and more specific to each agent. 
	Thus, the acquired knowledge in the first $k$ hidden layer(s) of the neural network belonging to agent $i$ is transferred to agent $j$, in which $k$ is a tunable parameter. Following this procedure, we test the use of transfer learning in six cases to transfer the learned knowledge of source agent $i$ to target agent $j$ with: 
	\begin{itemize} 
		\item[1.] $j \neq i$ in the same game. 
		
		\item[2.] $\{|\mathcal{A}_1^j|, c^j_{p_2}, c^j_{h_2}\}$, i.e., the same action space but different cost coefficients. 
		
		\item[3.] $\{|\mathcal{A}_2^j|, c^j_{p_1}, c^j_{h_1}\}$, i.e., the same cost coefficients but different action space. 
		
		\item[4.] $\{|\mathcal{A}_2^j|, c^j_{p_2}, c^j_{h_2}\}$, i.e., different action space and cost coefficients. 
		
		\item[5.] $\{|\mathcal{A}_2^j|, c^j_{p_2}, c^j_{h_2}\}$, i.e., different action space and cost coefficients, as well as a different demand distribution $D_2$. 

		\item[6.] $\{|\mathcal{A}_2^j|, c^j_{p_2}, c^j_{h_2}\}$, i.e., different action space and cost coefficients, as well as a different demand distribution $D_2$ and co-player policy $\pi_2$.  

	\end{itemize}
	
	Unless stated otherwise, the demand distribution and co-player policy are the same for the source and target agents. 	
	Transfer learning could also be used when other aspects of the problem change, e.g., lead times, state representation, and so on.
	This avoids having to tune the parameters of the neural network for each new problem, which considerably reduces the training time.
	However, we  still need to decide how many layers should be trainable, as well as to determine which agent can be a base agent for transferring the learned knowledge. Still, this is  computationally much cheaper than finding each network and its hyper-parameters from scratch.

\section{Numerical Experiments}\label{sec:beer_numerical_experiment}

In Section~\ref{sec:uniform_0_2_result}, we discuss a set of numerical experiments that uses a simple demand distribution and a relatively small action space: 
	\begin{itemize}
		\item $d_0^t \in \mathbb{U}[0,2]$, $\mathcal{A} = \{-2,-1,0,1,2\}$.
	\end{itemize} 

After exploring the behavior of SRDQN under different co-player (teammate) policies, in Section~\ref{sec:result:literature_cases} we test the algorithm using three well-known cases from the literature, which have larger possible demand values and action spaces: 
\begin{itemize}
	\item $d_0^t \in \mathbb{U}[0,8]$, $\mathcal{A} = \{-8,\dots, 8\}$ \citep{croson2006behavioral}
	\item $d_0^t \in \mathbb{N}(10,2^2)$, $\mathcal{A} = \{-5,\dots, 5\}$ \citep[adapted from][]{chen2000stationary}
	\item $d_0^t \in C(4,8)$, $\mathcal{A} = \{-8,\dots, 8\}$ \citep{sterman1989modeling}.
\end{itemize} 
Section \ref{sec:results:basket_dataset} presents the results on two real world datasets, and Section \ref{sec:results:sensitivity_analysis} provides sensitivity analysis on the trained models.
After analyzing these cases, in Section \ref{sec:results:transfer_learning} we provide the results obtained using transfer learning for each of the six proposed cases.
As noted above, we only consider cases in which a single SRDQN plays with non-SRDQN agents, e.g., simulated human co-players. In each of the cases listed above, we consider three types of policies that the non-SRDQN co-players follow: (i) base-stock policy, (ii) Sterman formula, (iii) random policy. In the random policy, agent $i$ also follows a $d+x$ rule, in which $a_i^t \in \mathcal{A}$ is selected randomly and with equal probability, for each $t$.

We test values of $m$ in $\{5,10\}$ and $C \in \{5000, 10000\}$. 
Considering the state and action spaces, our DNN network is a fully connected network with shape $[5m, 180, 130, 61, a_u - a_l + 1]$.
We trained each agent on at most 60000 episodes and used a replay memory $E$ equal to the one million most recently observed experiences. 
The $\epsilon$-greedy algorithm starts with $\epsilon = 0.9$ and linearly reduces it to $0.1$ in the first $80\%$ of iterations. 

In the feedback mechanism, 
the appropriate value of the feedback coefficient $\beta_i$ heavily depends on $\tau_j$, the average reward for agent $j$, for each $j\neq i$. 
For example, when $\tau_i$ is one order of magnitude larger than $\tau_j$, for all $j\ne i$, agent $i$ needs a large coefficient to get more feedback from the other agents. 
Indeed, the feedback coefficient has a similar role as the regularization parameter $\lambda$ has in the lasso loss function; the value of that parameter depends on the $\ell$-norm of the variables, but there is no universal rule to determine the best value for $\lambda$. Similarly, proposing a simple rule or value for each $\beta_i$ is not possible, as it depends on $\tau_i$, $\forall i$. For example, we found that a very large $\beta_i$ does not work well, since the agent tries to decrease other agents' costs rather than its own. Similarly, with a very small $\beta_i$, the agent learns how to minimize its own cost instead of the total cost. Therefore, we used a cross validation approach to find good values for each $\beta_i$. 

There are three types of co-players: base-stock, Sterman, and random. When the single player uses the SRDQN algorithm, we call the three cases {\tt BS-SRDQN}, {\tt Strm-SRDQN}, and {\tt Rand-SRDQN}, respectively. We compare the results of these cases to the results when the agent instead follows a base-stock policy; we call those cases {\tt BS-BS}, {\tt Strm-BS}, and {\tt Rand-BS}. In the {\tt -BS} cases, the single agent uses the optimal (or near-optimal) base-stock level for that case. In particular, in {\tt BS-BS} the agent's optimal base-stock level is provided by the algorithm by \citet{clark1960optimal}, whereas in {\tt Strm-BS} and {\tt Rand-BS}, we numerically optimize the base-stock level of the single base-stock player, since the literature does not provide guidance on how to set the base-stock level in this case. To do so, we evaluated each integer base-stock level in the range $[\mu - 10\sigma, \mu + 10\sigma]$ (for normal demands) or $[-25d_l,25d_u]$ (for uniform demands), where $\mu$ and $\sigma$ are the mean and average of the normal demand distribution and $d_l$ and $d_u$ are the lower and upper bound of the uniform demand distribution. For each base-stock level, we simulated 50 episodes of the game and calculated the total system-wide cost for each. We then selected the base-stock level that obtains the minimum average cost over the 50 episodes. 

We compare SRDQN with a base-stock policy in order to demonstrate two main things:
	\begin{itemize}
	\item When co-players use base-stock policies, the SRDQN agent can attain performance that is close to that of the optimal base-stock policy, even though that policy is found using algorithms (e.g., \cite{clark1960optimal}) that assume that all information about all players (their policies, cost coefficients, etc.)\ is available, whereas the SRDQN agent only knows its own local information and learns by experimentation. (We observe a similar effect for random co-players. Although the optimal policy in this setting is, to the best of our knowledge, unknown, it is plausible that a base-stock policy is optimal.)
	\item With Sterman co-players, the SRDQN agent significantly outperforms a base-stock policy, even with an optimized base-stock level. The optimal policy with Sterman co-players has not (to the best of our knowledge) been investigated in the literature, and our results suggest that a base-stock policy is not optimal in this case. 
	\end{itemize}

\subsection{Basic Cases}\label{sec:uniform_0_2_result}
In this section, we test our approach using a beer game setup with the following characteristics. Information and shipment lead times, $l^{tr}_j$ and $l^{in}_j$, equal 2 periods at every agent. Holding and stockout costs are given by $c_h=[2,2,2,2]$ and $c_p = [2,0,0,0]$, respectively, where the vectors specify the values for agents $1,\ldots,4$.  The demand is an integer uniformly drawn from $\{0,1,2\}$. Additionally, we assume that agent $i$ observes the arriving shipment $AS_{t}^i$ when it chooses its action for period $t$. We relax this assumption later. 
We use $a_l = -2$ and $a_u =2$, so that there are 5 outputs in the neural network. 
(Later, we expand these to larger action spaces.) 

For the cost coefficients and other settings specified for this beer game, it is optimal for all players to follow a base-stock policy, and we use this policy 
as a benchmark and a lower bound on the SRDQN cost. The vector of optimal base-stock levels is $[8,8,0,0]$, and the resulting average cost per period is $2.0705$, though these levels may be slightly suboptimal due to rounding. This cost is allocated to stages 1--4 as $[2.0073, 0.0632, 0.03, 0.00]$\label{opt_cost}. 
We present the results for systems with co-players that use base-stock and Sterman policies in Sections \ref{sec:results:dnn_vs_base_stock} and \ref{sec:results:dnn_vs_formula}, respectively. 

\subsubsection{SRDQN Plus Base-Stock Policy ({\tt BS-SRDQN})}\label{sec:results:dnn_vs_base_stock} 
\begin{figure}
	\caption{Total cost (upper figure) and normalized cost (lower figure) for {\tt BS-SRDQN} and {\tt BS-BS} cases. Caption indicates the role played by SRDQN.} 
	\label{fig:dqn_vs_three_optimal}
	\centering
	\begin{subfigure}{0.24\textwidth}
		\centerline{ \includegraphics[scale=0.16]{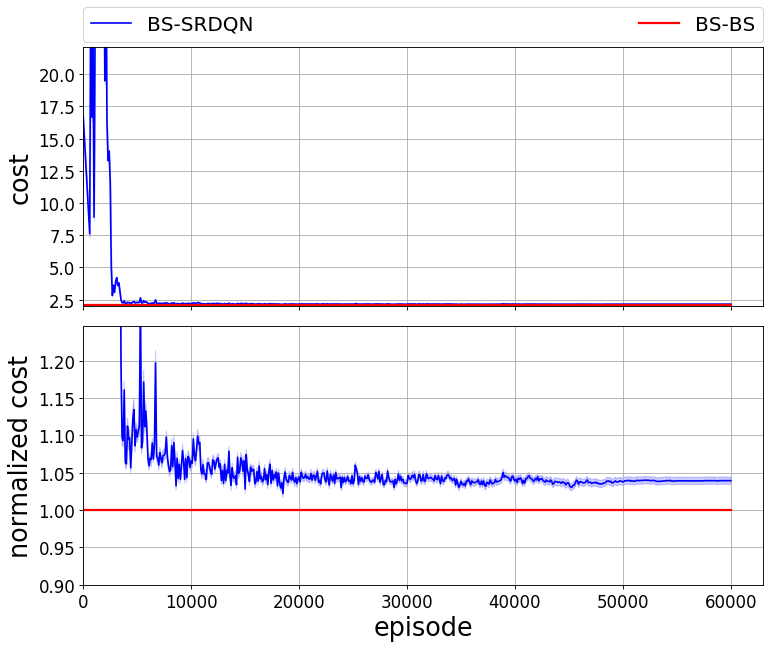}} 
		\vspace{-10pt}
		\caption{Retailer \label{fig:4Agent:vs_optm:A_DNN_Retailer}  		}
	\end{subfigure}
	\begin{subfigure}{0.24\textwidth}
		\centerline{ \includegraphics[scale=0.16]{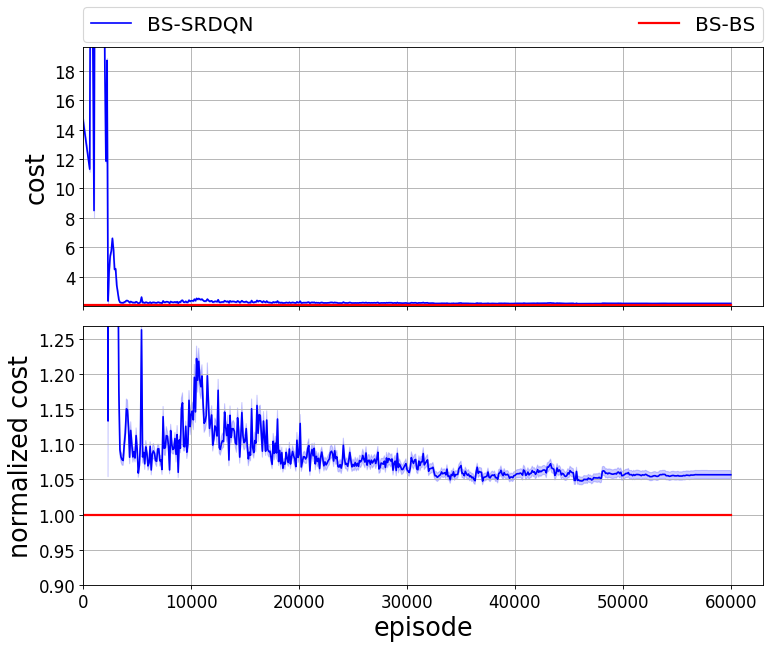}}
		\vspace{-10pt}
		\caption{Warehouse \label{fig:4Agent:vs_optm:A_DNN_Warehouse}  	}
	\end{subfigure}		
	\begin{subfigure}{0.24\textwidth}
		\centerline{ \includegraphics[scale=0.16]{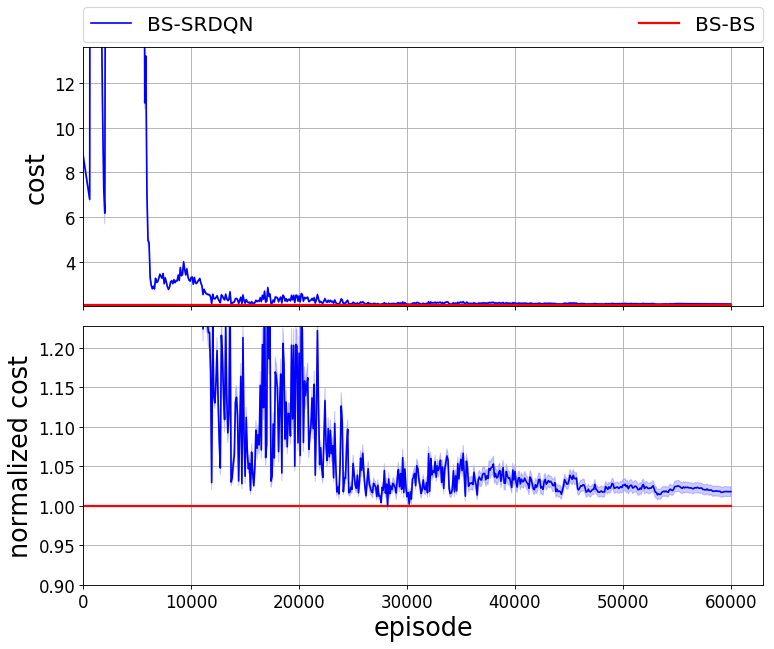}}
		\vspace{-10pt}
		\caption{Distributor \label{fig:4Agent:vs_optm:A_DNN_Distributer}}
	\end{subfigure}	
	\begin{subfigure}{0.24\textwidth}
		\centerline{ \includegraphics[scale=0.16]{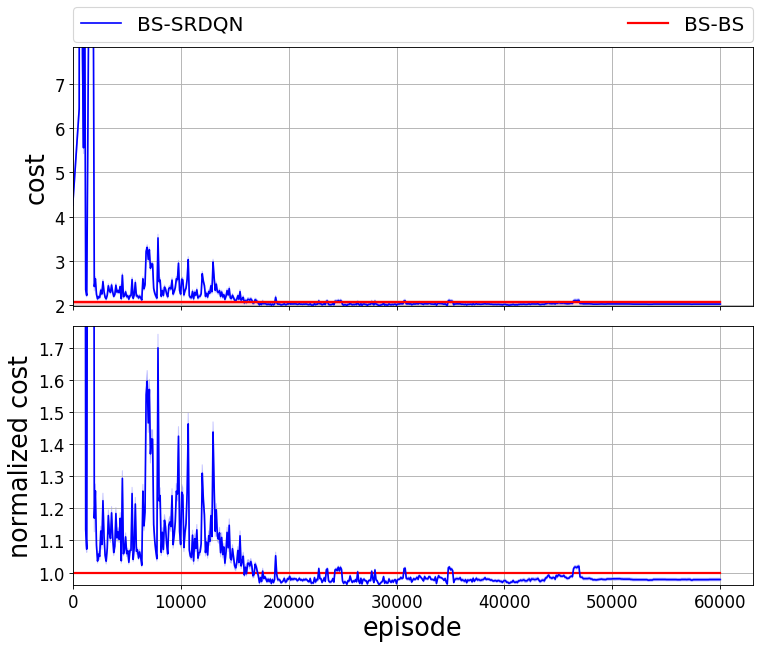}}
		\vspace{-10pt}
		\caption{Manufacturer \label{fig:4Agent:vs_optm:A_DNN_Manufacturer}}
	\end{subfigure}	
\end{figure}

We consider four cases, with the SRDQN playing the role of each of the four players and the co-players using a base-stock policy. 
The results of all four cases are shown in Figure \ref{fig:dqn_vs_three_optimal}. Each plot shows the training curve, i.e., the evolution of the average cost per game as the training progresses. In particular, the horizontal axis indicates the number of training episodes, while the vertical axis indicates the total cost per game. The caption indicates the role played by the SRDQN. After every 100 episodes of the game and the corresponding training, the cost of 50 validation points (i.e., 50 new games), each with 100 periods, are obtained and their average plus a 95\% confidence interval are plotted. (The confidence intervals, which are light blue in the figure, are quite narrow, so they are difficult to see.) The red line indicates the cost of the case in which all players follow a base-stock policy, which, recall, is optimal in this case. In each of the sub-figures, there are two plots; the upper plot shows the cost, while the lower plot shows the normalized cost, in which each cost is divided by the corresponding {\tt BS-BS} cost; essentially this is a ``zoomed-in'' version of the upper plot. 
We trained the network using values of $\beta_i \in \{5, 10, 20, 50, 100, 200\}$, each for at most 60000 episodes. Figure~\ref{fig:dqn_vs_three_optimal} plots the results from the best $\beta_i$ value for each agent; we present the full results using different $\beta_i$, $m$ and $C$ values in Section 
H 
of the online supplement. 

The figure indicates that SRDQN performs well in all cases and finds policies whose costs are close to those of {\tt BS-BS}. After convergence (i.e., after 60000 training episodes), the average gap between the cost of {\tt BS-SRDQN} and {\tt BS-BS}, over all four agents, is 2.31\%.

\begin{figure}
	\FIGURE
	{\includegraphics[scale=0.46]{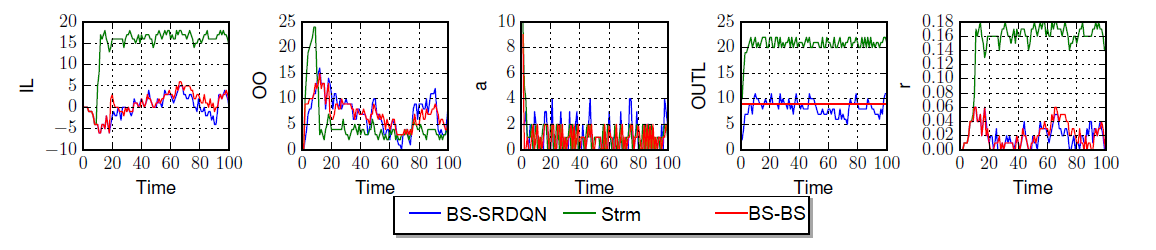}}
	{$IL_t$, $OO_t$, $a_t$, $r_t$, and OUTL when SRDQN plays retailer with $\beta_1=50$ ({\tt BS-SRDQN}), when all players play Sterman ({\tt Strm}), and when all players play base-stock ({\tt BS-BS}). \label{fig:4Agent:vs_optm:play_DNN_Retailer} }
	{}
\end{figure}

Figure \ref{fig:4Agent:vs_optm:play_DNN_Retailer} shows the trajectories of the retailer's inventory level ($IL$), on-order quantity ($OO$), order quantity ($a$), reward ($r$), and order up to level (OUTL) for a single game, when the retailer is played by the SRDQN ({\tt BS-SRDQN}), as well as when it is played by a base-stock policy ({\tt BS-BS}), and when all agents follow the Sterman formula ({\tt Strm}). 
The base-stock policy and SRDQN have similar $IL$ and $OO$ trends, and as a result their rewards are also very close: {\tt BS-BS} has a cost of $[1.42, 0.00, 0.02, 0.05]$ (total 1.49) and {\tt BS-SRDQN} has $[1.43, 0.01, 0.02, 0.08]$ (total 1.54, or 3.4\% larger). 
(Note that {\tt BS-BS} has a slightly different cost here than reported on page~\pageref{opt_cost} because those costs are the average costs of 50 samples while this cost is from a single sample.) 
Similar trends are observed when the SRDQN plays the other three roles; see Section 
G 
of the online supplement. 
This suggests that the SRDQN can successfully learn to achieve costs close to {\tt BS-BS} when the other agents also follow a base-stock policy. (The OUTL plot shows that the SRDQN does not quite {\em follow} a base-stock policy, even though its costs are similar.) Also, see \href{https://youtu.be/gQa6iWGcGWY}{\url{https://youtu.be/ZfrCrmGsDJE}} for a video animation of the policy that the SRDQN learns in this case.

\subsubsection{SRDQN Plus Sterman Formula}\label{sec:results:dnn_vs_formula}

\begin{figure}	
	\centering
	\caption{Total cost (upper figure) and normalized cost (lower figure) for {\tt Strm-SRDQN} and {\tt Strm-BS} cases. Caption indicates the role played by SRDQN.}
	\label{fig:dqn_vs_three_formula}
	\begin{subfigure}{0.24\textwidth}
		\centerline{ \includegraphics[scale=0.16]{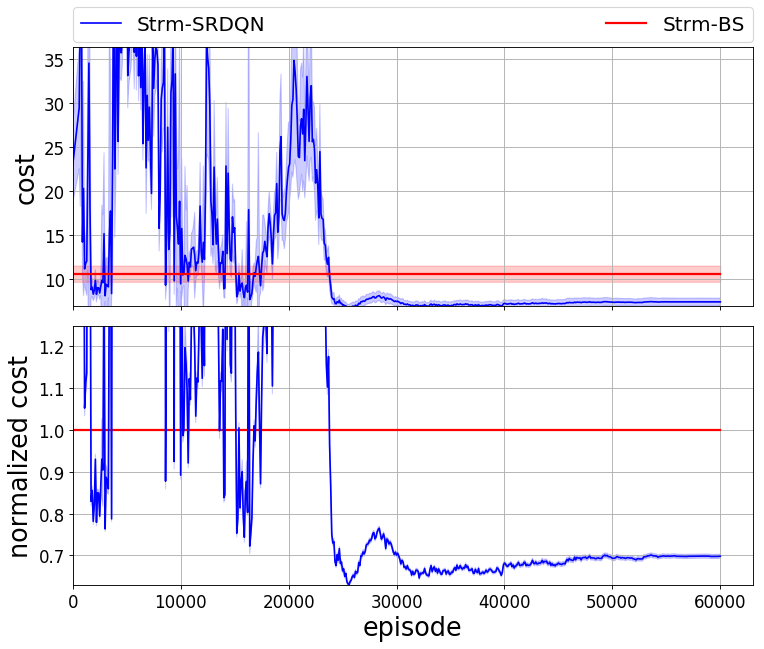}}
		\vspace{-10pt}		
		\caption{retailer}
		 \label{fig:4Agent:vs_frmu:A_DNN_Retailer}
	\end{subfigure}	
	\begin{subfigure}{0.24\textwidth}
		\centerline{ \includegraphics[scale=0.16]{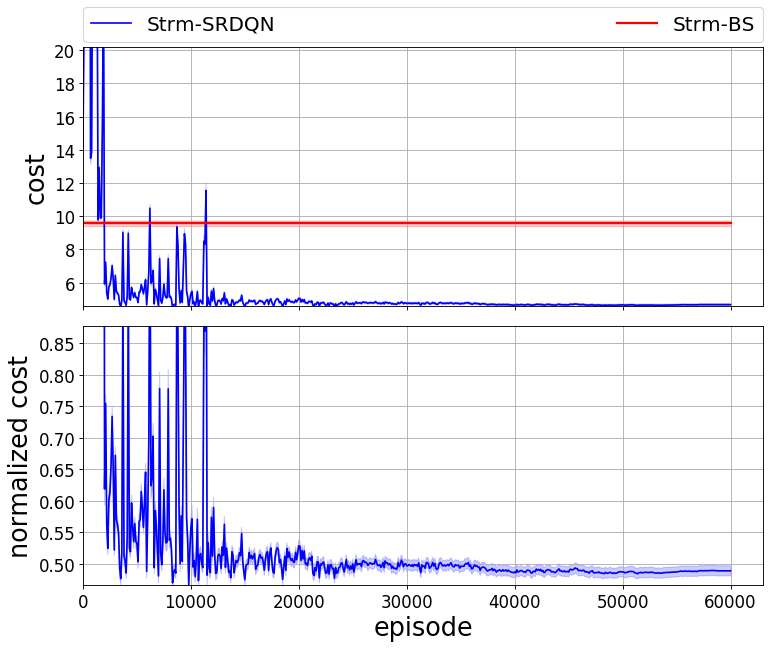}}
		\vspace{-10pt}		
		\caption{warehouse}
		\label{fig:4Agent:vs_frmu:A_DNN_Warehouse}
	\end{subfigure}
	\begin{subfigure}{0.24\textwidth}
		\centerline{ \includegraphics[scale=0.16]{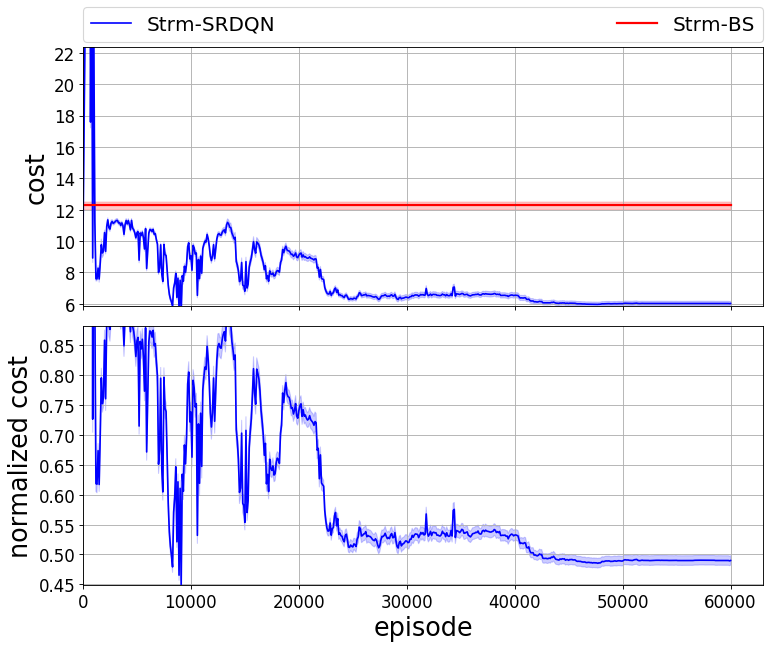}}
		\vspace{-10pt}		
		\caption{distributor}	   
		\label{fig:4Agent:vs_frmu:A_DNN_Distributer} 
	\end{subfigure}	
	\begin{subfigure}{0.24\textwidth}
		\centerline{ \includegraphics[scale=0.16]{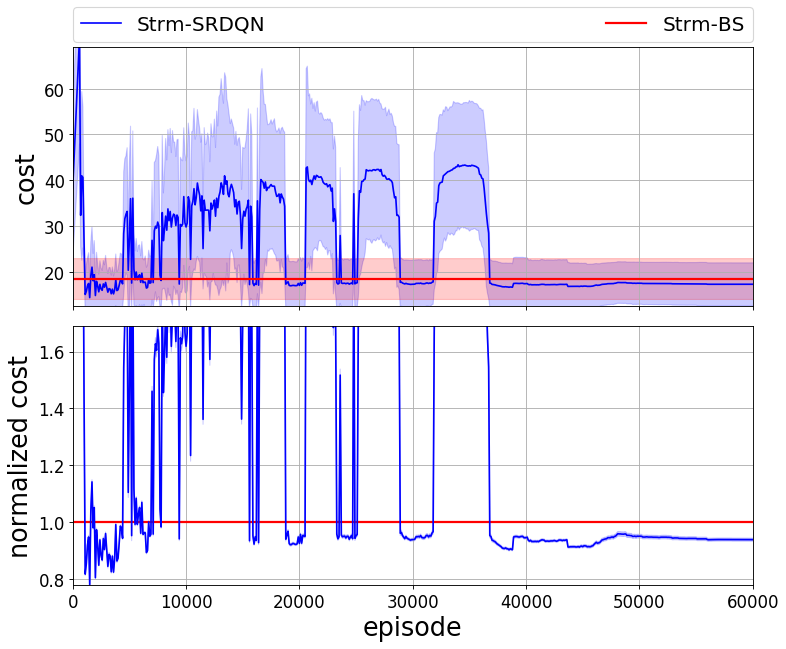}}
		\vspace{-10pt}
		\caption{manufacturer}
		\label{fig:4Agent:vs_frmu:A_DNN_Manufacturer}
	\end{subfigure}
\vspace{-20pt}
\end{figure}

Figure \ref{fig:dqn_vs_three_formula} shows the results of the case in which the three non-SRDQN agents use the formula proposed by \cite{sterman1989modeling}. 
(See Section 
B 
of online supplement for the formula and its parameters.) The single agent either uses SRDQN ({\tt Strm-SRDQN}) or optimized base-stock levels ({\tt Strm-BS}). 
From the figure, it is evident that the SRDQN plays much {\em better} than the optimized base-stock policy. This is because if the other three agents do not follow a base-stock policy, it is no longer optimal for the fourth agent to follow a base-stock policy. In general, the optimal inventory policy when other agents do not follow a base-stock policy is an open question. This figure suggests that SRDQN is able to learn to play effectively in this setting. 

Table \ref{tb:4Agent:vs_frmu:dqn_result} gives the cost of all four agents when a given agent plays using either SRDQN ({\tt Strm-SRDQN}) or a base-stock policy ({\tt Strm-BS}) and the other agents play using the Sterman formula. 
From the table, we can see that SRDQN learns how to play to decrease the costs of the other agents, and not just its own costs---for example, the warehouse's and manufacturer's costs are significantly lower when the distributor uses SRDQN than they are when the distributor uses a base-stock policy. Similar conclusions can be drawn from Figure~\ref{fig:dqn_vs_three_formula}. This shows the power of SRDQN when it plays with co-player agents that do not play rationally, i.e., do not follow a base-stock policy, which is common in real-world supply chains. Also, we note that when all agents follow the Sterman formula, the average cost of the agents is [2.10, 4.85, 11.22, 13.41], for a total of 31.58, much higher than when any one agent uses SRDQN.
Finally, for details on $IL,OO,a,r,$ and OUTL in this case, see Section 
G 
of the online supplement.

{\SingleSpacedXI
\begin{table}
	\centering
	\caption{Average cost under different choices of which agent uses SRDQN or base-stock, with Sterman co-players. \label{tb:4Agent:vs_frmu:dqn_result}}
	\begin{tabular}{l|ccccc}
		& \multicolumn{5}{c}{Cost ({\tt Strm-SRDQN}, {\tt Strm-BS})} \\
		\cline{2-6} 
		SRDQN~Agent & Retailer & Warehouse & Distributor & Manufacturer & Total \\	\hline
		Retailer  & (1.75, 3.57) & (0.89, 1.04) & (1.75, 2.35) & (3.02, 3.60) & (7.41, 10.56) \\ 
		Warehouse  & (1.67, 2.19) & (0.26, 0.48) & (1.02, 2.62) & (1.73, 4.26) & (4.68, 9.56) \\
		Distributor  & (3.24, 1.80) & (0.74, 3.66) & (0.00, 2.01) & (2.03, 4.78) & (6.01, 12.25) \\
		Manufacturer  & (1.89, 1.75) & (3.89, 3.53) & (9.45, 8.45) & (2.03, 4.67) & (17.26, 18.40) \\ \hline 
	\end{tabular}
\end{table}	
}

\subsection{Literature Benchmarks}\label{sec:result:literature_cases}

We next test SRDQN on beer game settings from the literature. These have larger demand-distribution domains, and therefore larger plausible action spaces, so represent harder instances for SRDQN.
In all instances in this section, $l^{in}=[2,2,2,2]$ and $l^{tr}=[2,2,2,1]$. Cost coefficients and the base-stock levels for each instance are presented in Table \ref{tb:experimental_parameters}. 

Note that the Clark--Scarf algorithm assumes that stage 1 is the only stage with non-zero stockout costs, whereas the $\mathbb{U}[0,8]$ instance has non-zero costs at every stage. Therefore, we used a heuristic approach based on a two-moment approximation, similar to that proposed by \cite{Gr85}, to choose the base-stock levels; see \cite{snyder2018meio}. In addition, the $C(4,8)$ demand process is non-stationary---4, then 8---but we allow only stationary base-stock levels. Therefore, we chose to set the base-stock levels equal to the values that would be optimal if the demand were 8 in every period. 
Finally, in the experiments in this section, we assume that agent $i$ observes $AS_{t}^i$ {\em after} choosing $a_{t}^i$, whereas in Section~\ref{sec:uniform_0_2_result} we assumed the opposite. Therefore, the agents in these experiments have one fewer piece of information when choosing actions, and are therefore more difficult to train. 

{\SingleSpacedXI
\begin{table}
	\centering
	\caption{Cost parameters and base-stock levels for instances with uniform, normal, and classic demand distributions.}
	\label{tb:experimental_parameters}
	\begin{tabular}{l|ccc}
		demand & $c_p$ & $c_h$ & base-stock level \\ \hline
		$\mathbb{U}[0,8]$ & [1,1,1,1] & [0.50,0.50,0.50,0.50] & [19,20,20,14] \\
		$\mathbb{N}(10,2^2)$ & [10,0,0,0] & [1.00,0.75,0.50,0.25] & [48,43,41,30] \\		
		$C(4,8)$ & [1,1,1,1] & [0.50,0.50,0.50,0.50] & [32,32,32,24] 
	\end{tabular}
\end{table}
}

\begin{figure}
	\caption{Confidence intervals for Normal and Uniform distributions.}
	\label{fig:confidence_interval_normal_uniform}
\begin{subfigure}{0.24\textwidth}
	\includegraphics[scale=0.3]{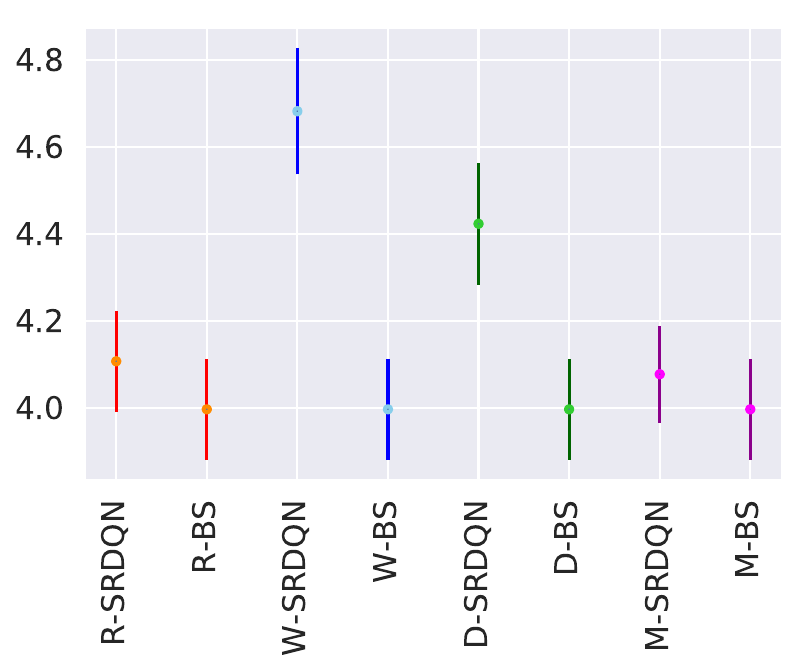}
	\caption{$U[0,8]$, {\tt BS}}
	\label{fig:confidence_interval-literature_uniform_BS}
\end{subfigure}
\begin{subfigure}{0.24\textwidth}
	\includegraphics[scale=0.3]{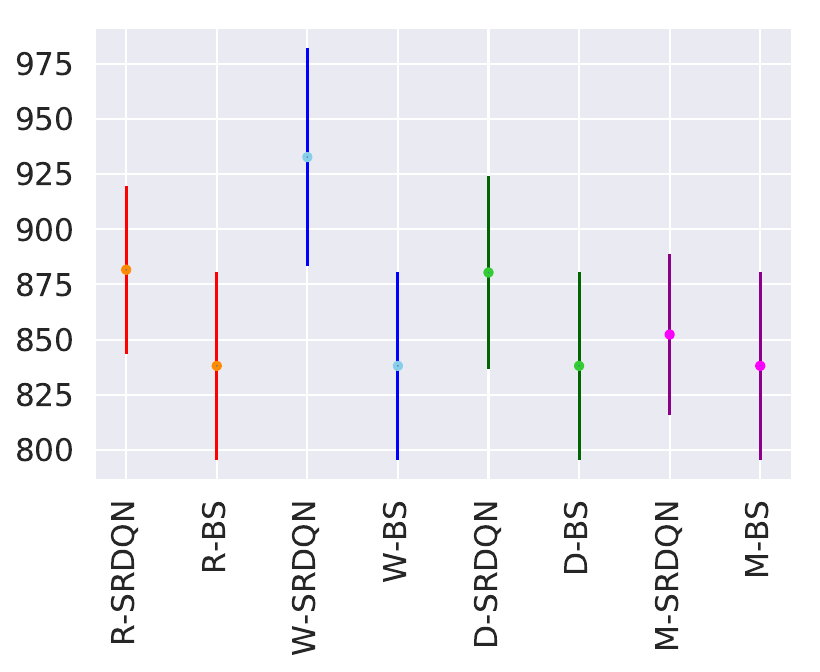}
	\caption{$\mathbb{N}(10,2^2)$, {\tt BS}}
	\label{fig:confidence_interval-literature_normal_BS}
\end{subfigure}
\begin{subfigure}{0.24\textwidth}
	\includegraphics[scale=0.3]{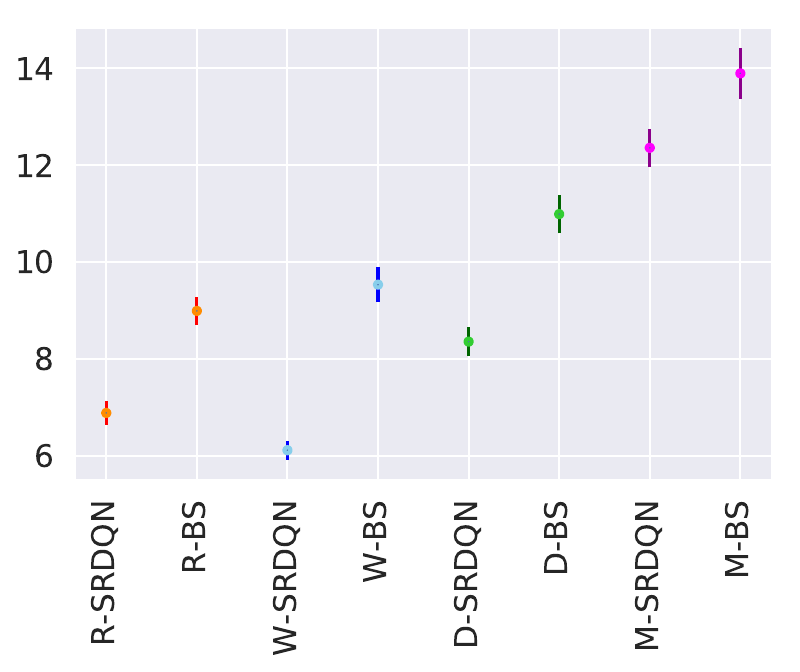}
	\caption{$U[0,8]$, {\tt Strm}}
	\label{fig:confidence_interval-literature_uniform_Strm}
\end{subfigure}
\begin{subfigure}{0.24\textwidth}
	\includegraphics[scale=0.3]{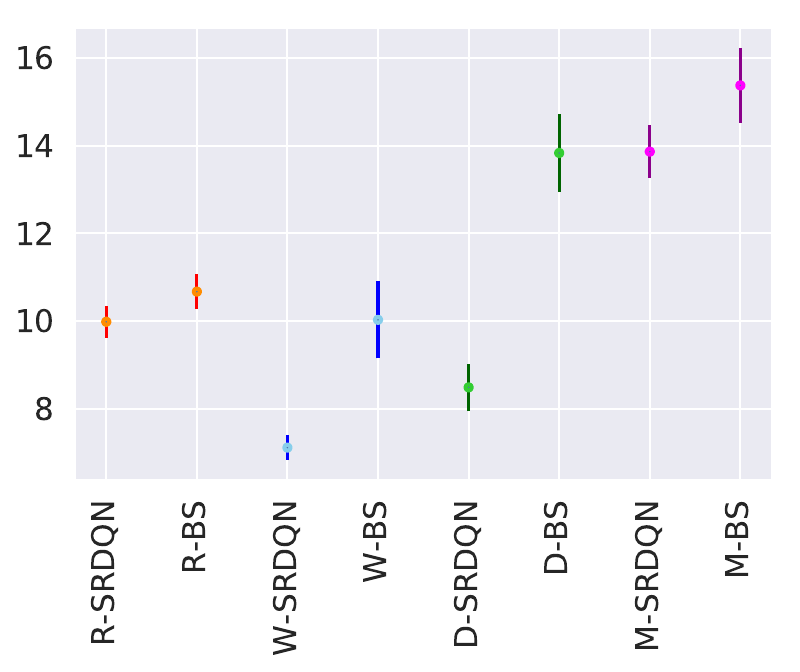}
	\caption{$\mathbb{N}(10,2^2)$, {\tt Strm}}
	\label{fig:confidence_interval-literature_normal_Strm}
\end{subfigure}

\begin{subfigure}{0.24\textwidth}
	\includegraphics[scale=0.3]{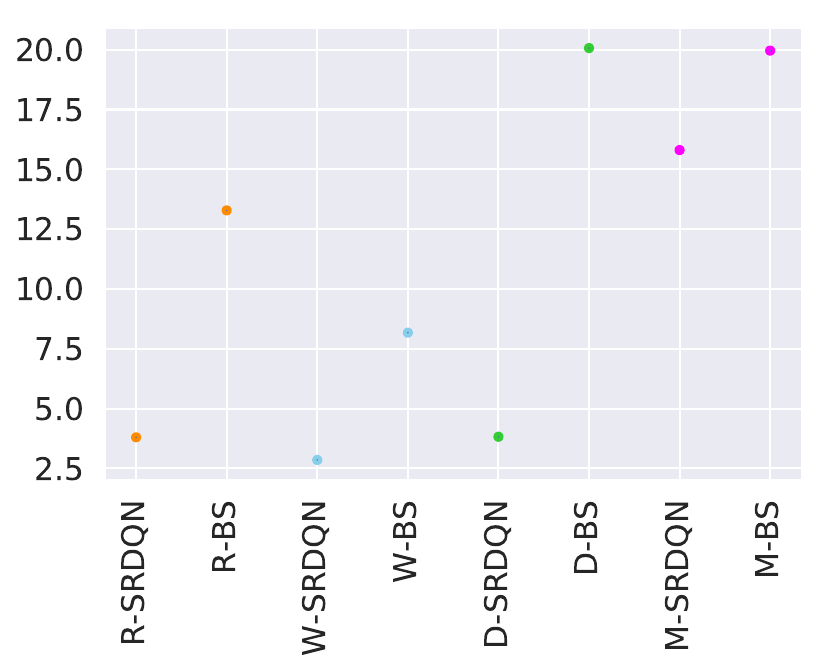}
	\caption{$\mathbb{C}(4, 8)$, {\tt Strm}}
	\label{fig:confidence_interval-literature_classic_Strm}
\end{subfigure}
\begin{subfigure}{0.24\textwidth}
	\includegraphics[scale=0.3]{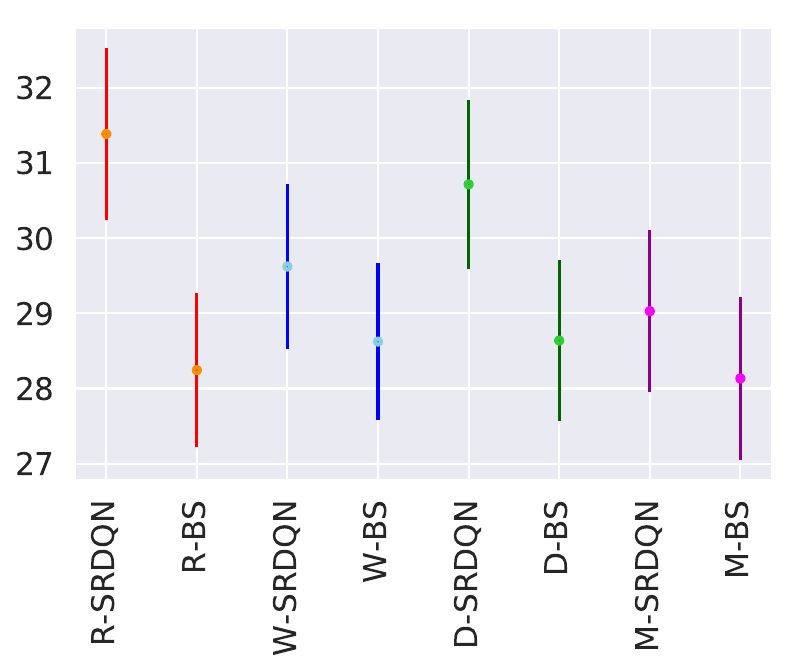}
	\caption{$U[0,8]$, {\tt Rand}}
	\label{fig:confidence_interval-literature_uniform_Rand}
\end{subfigure}
\begin{subfigure}{0.24\textwidth}
	\includegraphics[scale=0.3]{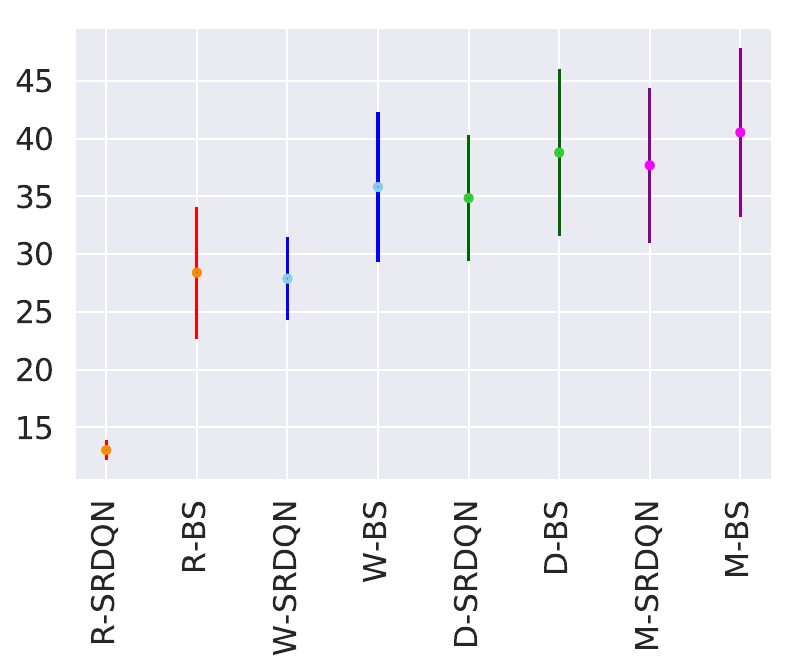}
	\caption{$\mathbb{N}(10,2^2)$, {\tt Rand}}
	\label{fig:confidence_interval-literature_normal_Rand}
\end{subfigure}
\begin{subfigure}{0.24\textwidth}
	\includegraphics[scale=0.3]{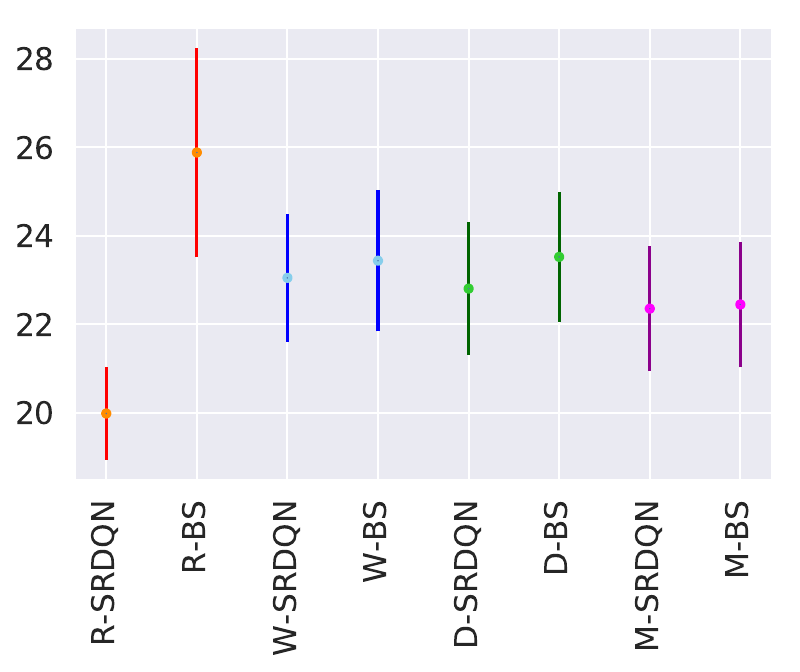}
	\caption{$\mathbb{C}(4, 8)$, {\tt Rand}}
	\label{fig:confidence_interval-literature_classic_Rand}
\end{subfigure}
\end{figure}

Table \ref{tb:literature_results} shows the results of the cases in which the SRDQN agent plays with co-players who follow base-stock, Sterman, and random policies (as listed in the ``Co-Player Type'' column). In each group of columns, the first column (``{\tt -SRDQN}'') gives the average cost (over 50 instances) when one agent (indicated by the ``Agent'' column) is played by the SRDQN and co-players are played by the player type indicated in the ``Co-Player Type'' column. 
The second column in each group (``{\tt -BS}'' gives the corresponding cost when the SRDQN agent is replaced by a base-stock agent (using the optimized base stock levels) and the co-players remain as the row-set determines. The third column (``Gap'') gives the percentage difference between these two costs. We use R, W, D, W to represent retailer, wholesaler, distributor, and manufacturer, respectively. 
(For example, when the co-players are Sterman agents, and the SRDQN agent plays the role of the distributor, the average total cost is 8.35, whereas if the distributor is played by a base-stock agent, the average total cost is 10.10.) In addition, Figure~\ref{fig:confidence_interval_normal_uniform} presents 90\% confidence intervals for all cases, except the classic distribution with {\tt BS} co-players, which has 0 variance.

As the first row-set in Table \ref{tb:literature_results} shows, when the SRDQN plays with base-stock co-players under uniform or normal demand distributions, it obtains costs that are reasonably close to the case when all players use a base-stock policy, with average gaps of 8.15\% and 5.80\%, respectively. These gaps are not quite as small as those in Section~\ref{sec:uniform_0_2_result}, due to the larger action spaces in the instances in this section. Nevertheless, these gaps do not necessarily mean that SRDQN performs significantly 
worse than {\tt BS}. Although the gap is positive for all cases with {\tt BS} co-players, Figures \ref{fig:confidence_interval-literature_uniform_BS} and \ref{fig:confidence_interval-literature_normal_BS} show that in 5 out of 8 cases, the average costs of SRDQN and {\tt BS} are statistically equal. In addition, since a base-stock policy is optimal at every stage, the small gaps demonstrate that the SRDQN can learn to play the game well for these demand distributions. For the classic demand process, the percentage gaps are larger. To see why, note that if the demand were equal to 8 in every period, the optimal base-stock levels for the classic demand process will result in $IL_i^t=0, \forall i, t$. The four initial periods of demand equal to 4 disrupt this effect slightly, but the cost of the optimal base-stock policy for the classic demand process is asymptotically 0 as the time horizon goes to infinity. The absolute gap attained by the SRDQN is quite small---an average of 0.49 vs.~0.34 for the base-stock cost---but the percentage difference is large simply because the optimal cost is close to 0. Indeed, if we allow the game to run longer, the cost of both algorithms decreases, and so does the absolute gap. For example, when the SRDQN plays the retailer, after 500 periods the discounted costs are 0.0090 and 0.0062 for {\tt BS-SRDQN} and {\tt BS-BS}, respectively, and after 1000 periods, the costs are 0.0001 and 0.0000 (to four-digit precision).


Similar to the results of Section \ref{sec:results:dnn_vs_formula}, when the SRDQN plays with co-players who follow the Sterman formula, it performs far better than {\tt Strm-BS}. As Table \ref{tb:literature_results} shows, SRDQN performs 23\% better than {\tt Strm-BS} on average. In addition Figures \ref{fig:confidence_interval-literature_uniform_Strm}, \ref{fig:confidence_interval-literature_normal_Strm}, and \ref{fig:confidence_interval-literature_classic_Strm} show that in 11 out of 12 cases, the cost of {\tt Strm-SRDQN} is statistically smaller than that of {\tt Strm-BS}.

On the other hand, when SRDQN plays with co-players who use the random policy, for all demand distributions {\tt Rand-SRDQN} learns well to play so as to minimize the total cost of the system, although {\tt Rand-BS} on average obtains 12\% better solutions. Nonetheless, as Figures \ref{fig:confidence_interval-literature_uniform_Rand}, \ref{fig:confidence_interval-literature_normal_Rand}, and \ref{fig:confidence_interval-literature_classic_Rand} show, {\tt Rand-BS} obtains a statistically smaller cost in only in one case, while in two cases SRDQN obtains a statistically smaller cost, and in the rest of the nine cases, the means of the costs are statistically equal.
Our conjecture to explain why the results here are not as strong as those for {\tt Strm-BS} is that when the co-players take random actions, the demand process faced by the single agent is approximately an iid random process, for which a base-stock policy would be approximately optimal, similar to the {\tt BB-BS} case. (In contrast, Sterman players place orders that are highly correlated over time, rather than iid random.) 

{\SingleSpacedXI
	\begin{table}[]
		\centering
		\small
		\caption{Results of literature datasets.}
		\label{tb:literature_results}
		\begin{tabular}{ccccc|ccc|ccc}
		Co-Player & & \multicolumn{3}{c}{Uniform $\mathbb{U}[0,8]$} & \multicolumn{3}{c}{Normal $\mathbb{N}(10,2^2)$} & \multicolumn{3}{c}{Classic $C(4,8)$} \\
		Type & Agent     & {\tt -SRDQN} & {\tt -BS}     & Gap (\%) & {\tt -SRDQN} & {\tt -BS}     & Gap (\%) & {\tt -SRDQN} & {\tt -BS}    & Gap (\%)  \\ \hline
		\multirow{5}{*}{{\tt BS-}} & \multicolumn{1}{c|}{R}  & 4.11  & 4.00 & 2.76    & 4.41 & 4.19 & 5.19     & 0.50   & 0.34   & 45.86     \\
		& \multicolumn{1}{c|}{W} & 4.69  & 4.00 & 17.15    & 4.66 & 4.19 & 11.28    & 0.47   & 0.34   & 36.92     \\
		& \multicolumn{1}{c|}{D} & 4.42  & 4.00 & 3.67    & 4.40 & 4.19 & 5.04     & 0.67   & 0.34   & 96.36     \\
		& \multicolumn{1}{c|}{M} & 4.08  & 4.00 & 2.02     & 4.26 & 4.19 & 1.69     & 0.30   & 0.34   & -13.13    \\ 
		& \multicolumn{3}{l}{Average}               & 8.15    &        &        & 5.80     &        &        & 41.50    \\
		\hline
		\multirow{5}{*}{{\tt Strm-}} & \multicolumn{1}{c|}{R} & 6.88  & 8.74  & -21.29 & 9.98  & 10.54 & -5.27  & 3.80  & 5.31 & -28.54 \\
		& \multicolumn{1}{c|}{W} & 5.90  & 8.58  & -31.22 & 7.11  & 8.72 & -18.40 & 2.85  & 3.84  & -25.76 \\
		& \multicolumn{1}{c|}{D} & 8.35  & 10.10 & -17.32 & 8.49  & 11.52 & -26.35 & 3.82  & 17.70 & -78.41 \\
		& \multicolumn{1}{c|}{M} & 12.36 & 12.75 & -3.10 & 13.86 & 15.05 & -7.87  & 15.80 & 17.02 & -7.13 \\ 
		& \multicolumn{3}{l}{Average} & -18.23 &       &       & -14.47 &       &       & -34.96 \\
		\hline
		\multirow{5}{*}{{\tt Rand-}} 
		& \multicolumn{1}{c|}{R} & 31.39 & 27.69 & 13.34 & 13.03 & 15.93 & -18.19 & 19.99 & 20.30 & -1.55 \\
		& \multicolumn{1}{c|}{W} & 29.62 & 27.59 & 7.37  & 27.87 & 22.33 &  24.80 & 23.05 & 20.03 &  15.08  \\
		& \multicolumn{1}{c|}{D} & 30.72 & 27.97 & 9.82  & 34.85 & 26.03 &  33.90 & 22.81 & 21.06 &  8.31 \\
		& \multicolumn{1}{c|}{M} & 29.03 & 27.09 & 7.16  & 37.68 & 29.36 &  28.36 & 22.36 & 20.17 &  10.85  \\ 
		& \multicolumn{3}{l}{Average} & 9.42  &       &       & 17.22 &       &       & 8.17 
	\end{tabular}
\end{table}
} 

\subsection{Real-World Data Sets}\label{sec:results:real_world_dataset}
In order to test the performance of SRDQN on more challenging cases, we used two real-world datasets. The main question we wish to address in this section is {\em whether our method still performs well even if the historical data are scant}. The basic idea is to take the historical data, fit a demand distribution to those data, and then train the SRDQN on many samples generated randomly from that distribution; in other words, we use a small data set to generate a much larger one, which is used for training. A secondary question in this section is {\em whether the method performs well even when the demand distribution has larger mean and variance}, in which case the $d+x$ rule is less effective.

In each dataset, we randomly selected three categories/items for our experiment. Using the empirical demand distribution, we generated 60000 episodes and used the same training procedure as in previous sections. In particular, we assumed the same shortage and holding costs as in the setting with  ${\mathbb N}(10,2)$ demand in Section \ref{sec:result:literature_cases}, obtained the optimal base-stock levels, and trained models for the three co-player types. We note that in these experiments we report averages over 200 episodes (i.e. 20000 periods), instead of 50 as in previous sections, because the demands are noisier in this experiment. 

In both datasets, the demands are badly scaled, meaning the demand values can be quite large (e.g., up to 140). This is a difference from the cases in Section \ref{sec:result:literature_cases} and it makes the training much more difficult. Therefore, we scaled the demands by dividing each demand observation by the maximum demand across all items, times 10, so that all demands are within $[0,10]$. 
The results below use the resulting, scaled datasets. The results are promising, and from them we can conclude that the SRDQN performs well for properly scaled datasets. For datasets with larger demand values, our preliminary results showed that the SRDQN attains larger costs than the optimized base-stock policy---usually within 10\%, and always within 20\%, on average. It is possible that one could train a network on a scaled dataset and then ``unscale'' the results to apply to the original dataset; such an approach is a topic for future research, as are improved methods for training directly on non-scaled datasets.


Note that in the subsections below, we do not report results for random co-players. We did not test this case in order to conserve computational resources and also since these results tend to be similar to the cases with base-stock co-players.

\subsubsection{Real-World Dataset I}\label{sec:results:basket_dataset}
The first dataset is a ``basket'' dataset \citep{FoodMart}, which consists of data from a retailer (Foodmart) in 1997 and 1998. This dataset provides demands for 24 item categories, from which we randomly chose three. Histograms of the demands for the selected categories are shown in Figures \ref{subfig:basket_demand6}--\ref{subfig:basket_demand22}. Note that in each category there are fewer than 500 historical samples, which makes the problem hard to solve. 

Table \ref{tb:basket_results_scaled} presents the results. As shown, in all cases {\tt BS-SRDQN} obtains policies whose costs are close to those of {\tt BS-BS}, with an average gap of 6.22\%. Moreover, when it plays with Sterman co-players, in all but two cases {\tt Strm-SRDQN} performs better than {\tt Strm-BS}, with an average cost that is 3.27\% lower. 

{
\small
\begin{figure}
\caption{Demand histograms for the randomly selected categories/items for datasets I (subfigures (a)--(c)) and II (subfigures (d)--(e)).}
\label{fig:basket_demands}			
\centering
\begin{subfigure}{0.30\textwidth}
	\centering
	\includegraphics[scale=0.14]{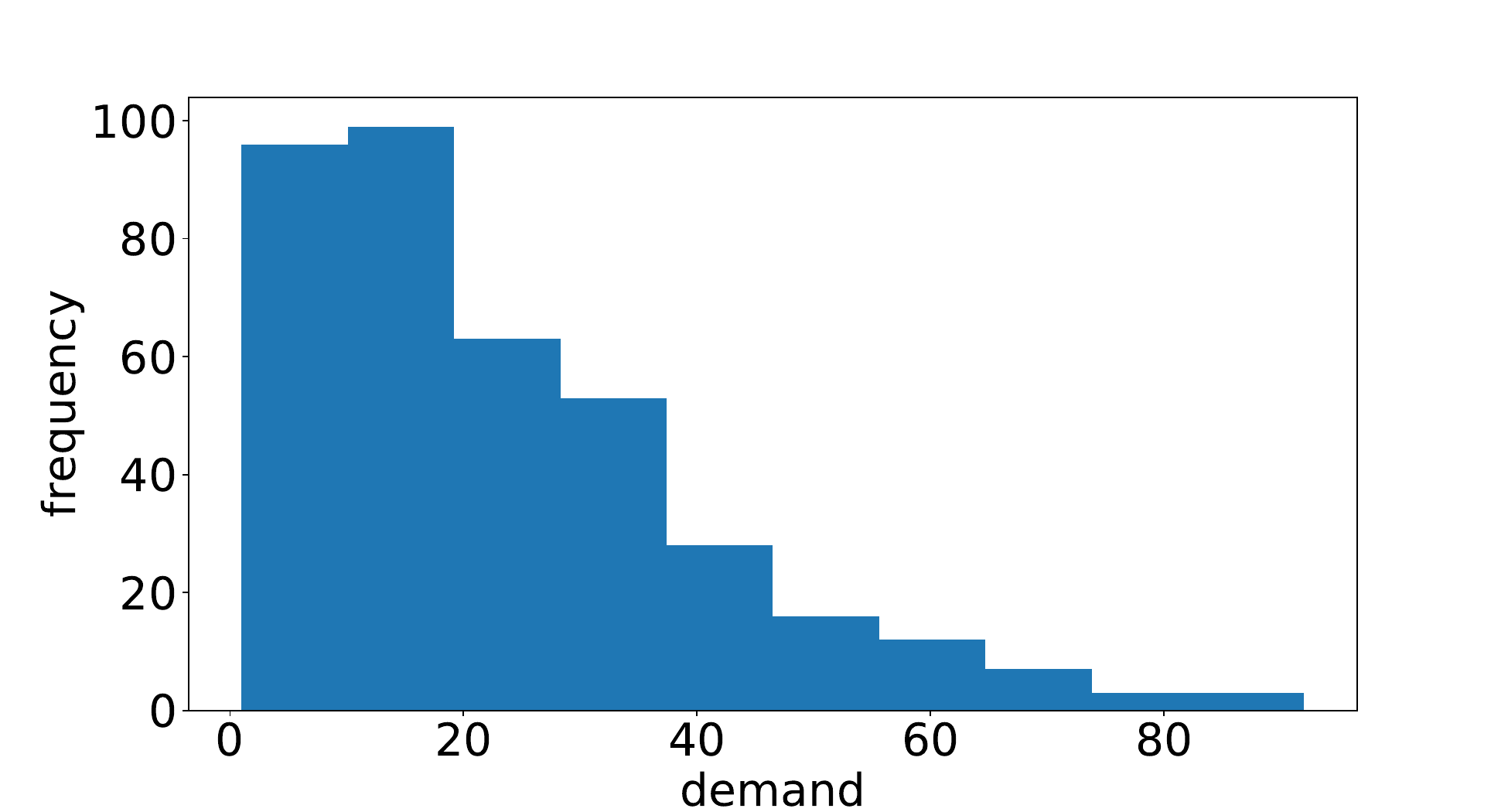}
	\vspace{-13pt}			
	\caption{{\small Category 6 (475 records)}}
	\label{subfig:basket_demand6}		
\end{subfigure}
\begin{subfigure}{0.30\textwidth}
	\centering
	\includegraphics[scale=0.14]{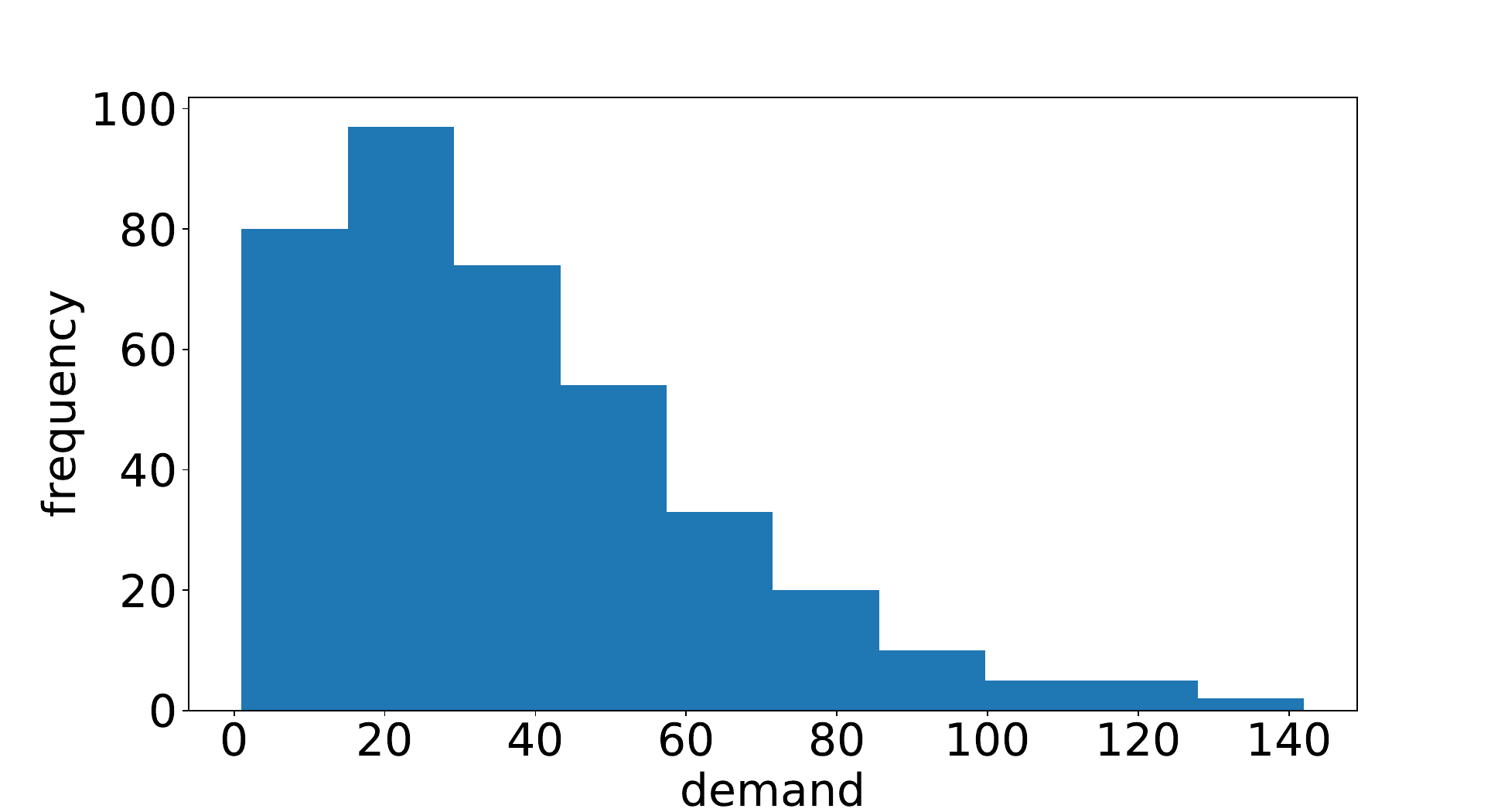}
	\vspace{-13pt}						
	\caption{{\small Category 13 (475 records)}}
	\label{subfig:basket_demand13}	
\end{subfigure}	
\begin{subfigure}{0.30\textwidth}
	\centering
	\includegraphics[scale=0.14]{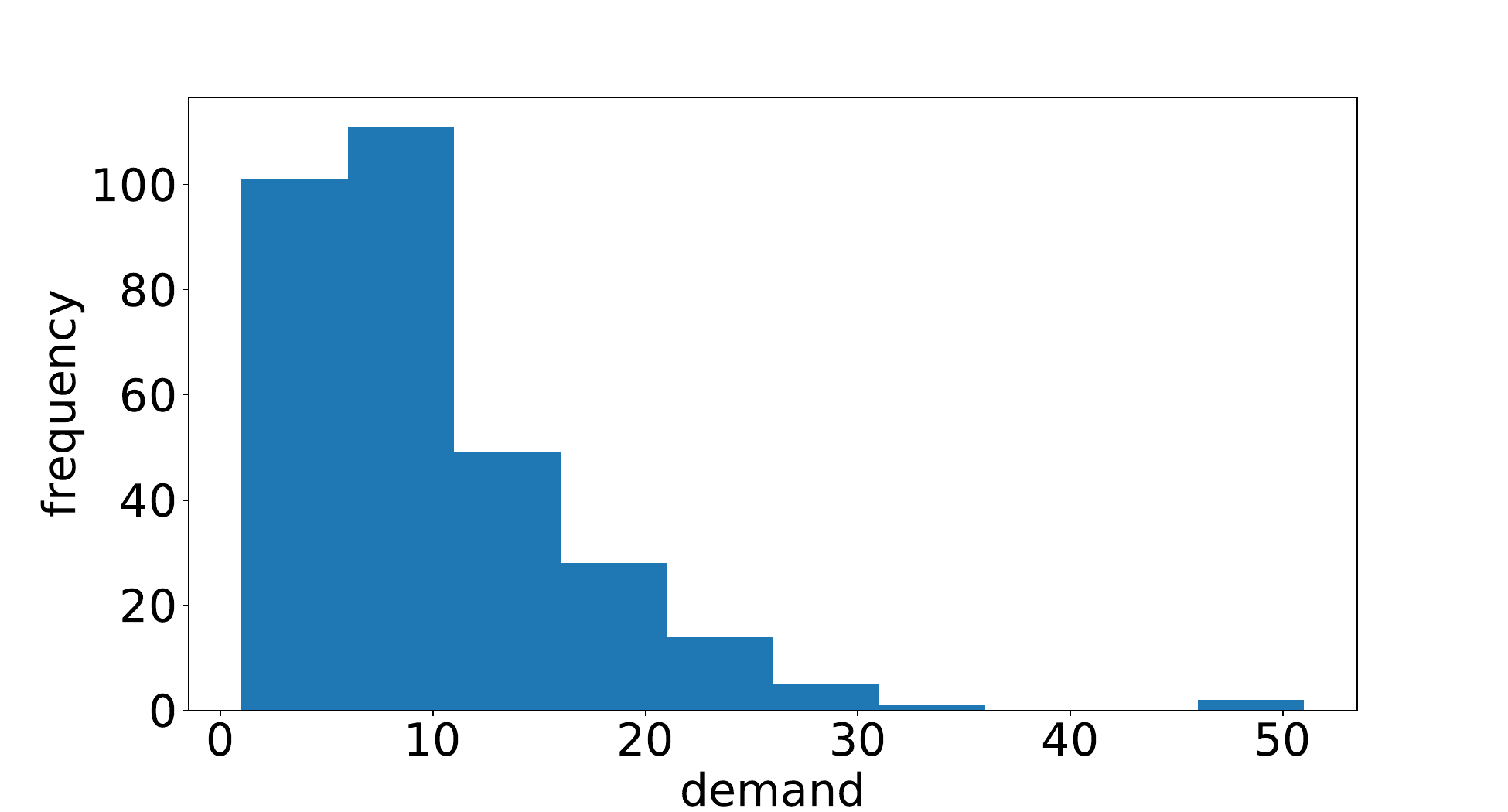}
	\vspace{-13pt}						
	\caption{{\small Category 22 (389 records)}}
	\label{subfig:basket_demand22}	
\end{subfigure}	

\begin{subfigure}{0.30\textwidth}
	\centering
	\includegraphics[scale=0.14]{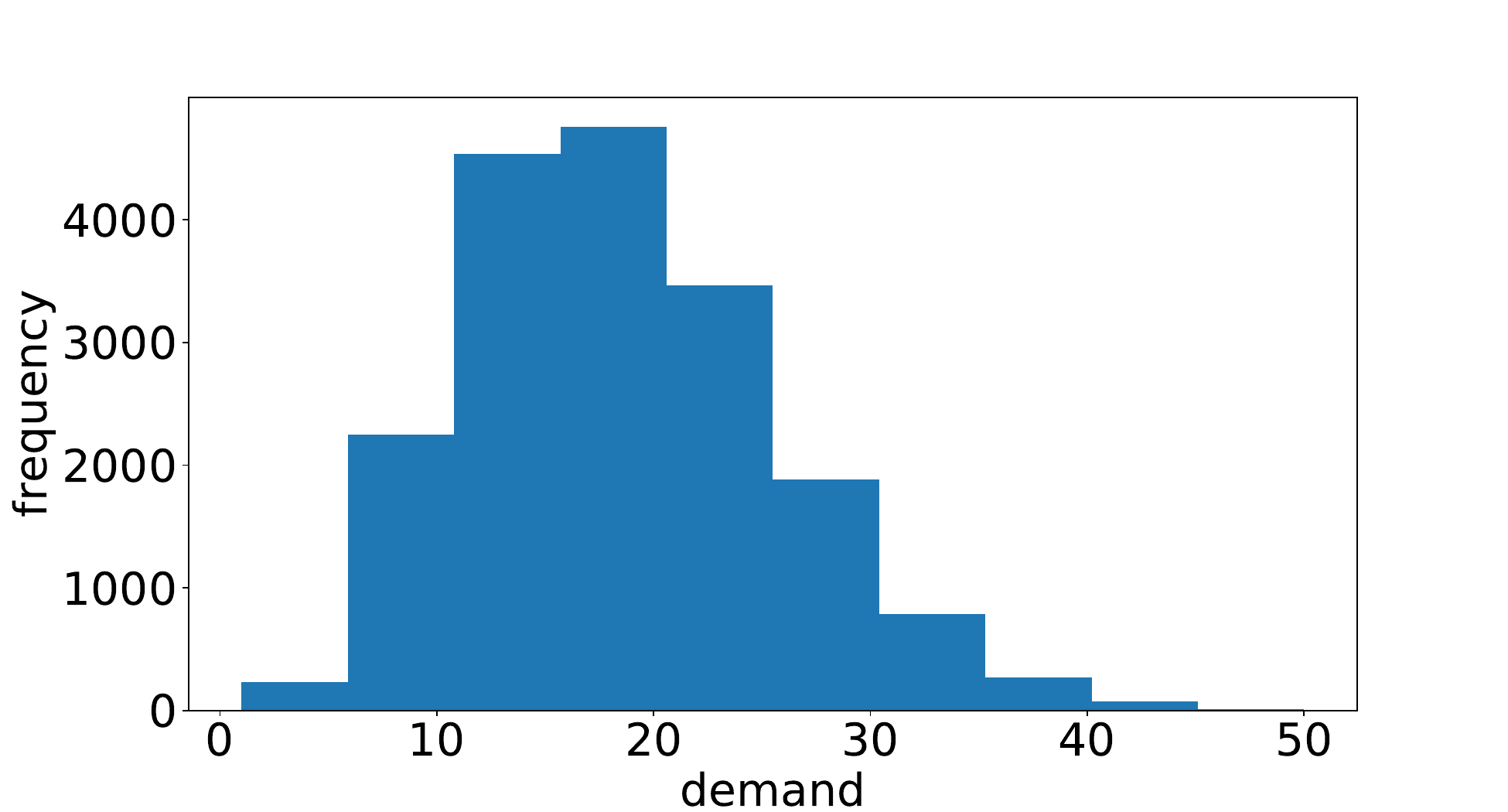}
	\vspace{-13pt}			
	\caption{{\small Item 5 (18260 records)}}
	\label{subfig:forecasting_demand5}		
\end{subfigure}
\begin{subfigure}{0.30\textwidth}
	\centering
	\includegraphics[scale=0.14]{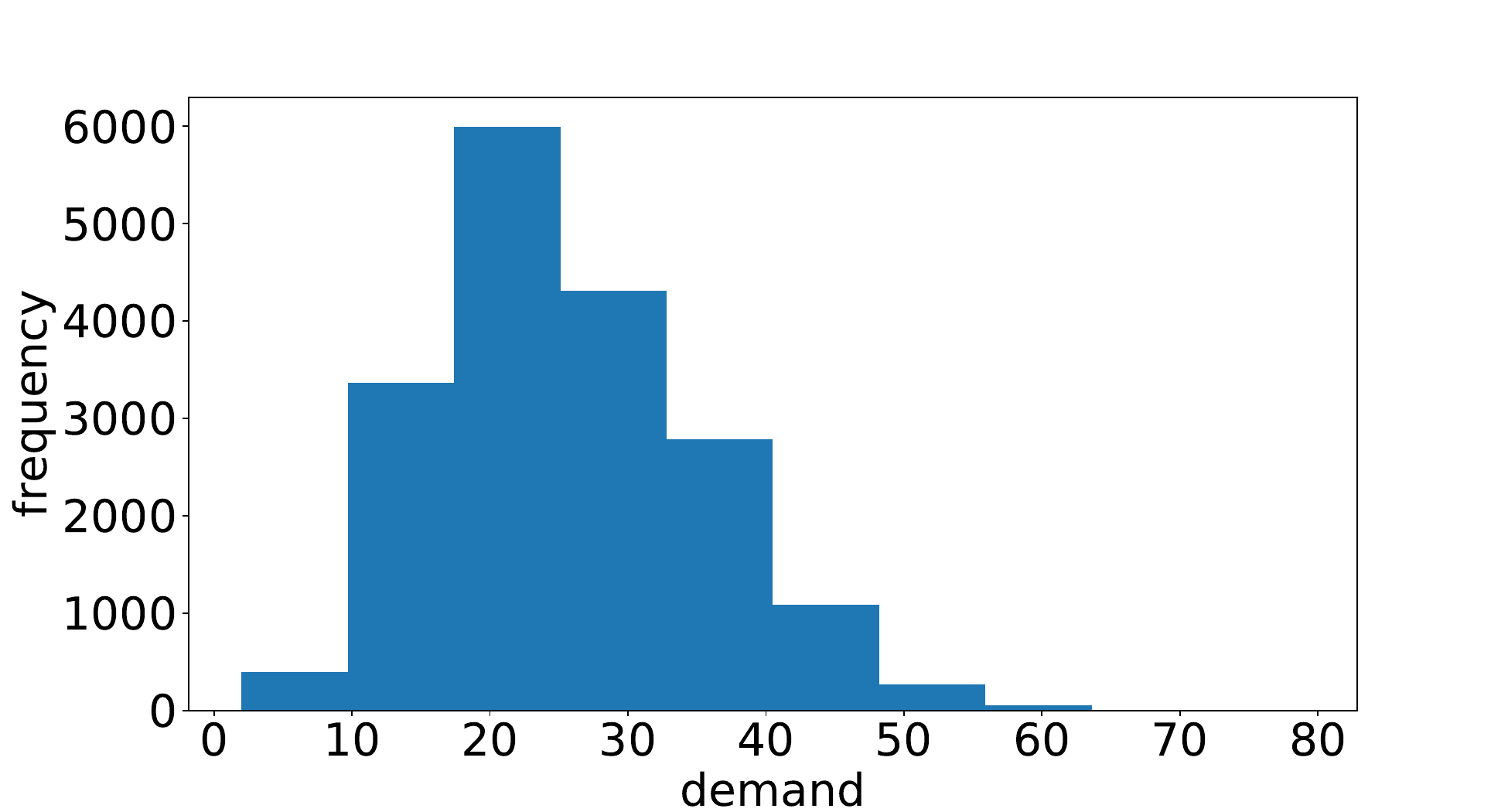}
	\vspace{-13pt}						
	\caption{{\small Item 34 (18260 records)}}
	\label{subfig:forecasting_demand34}	
\end{subfigure}	
\begin{subfigure}{0.30\textwidth}
	\centering
	\includegraphics[scale=0.14]{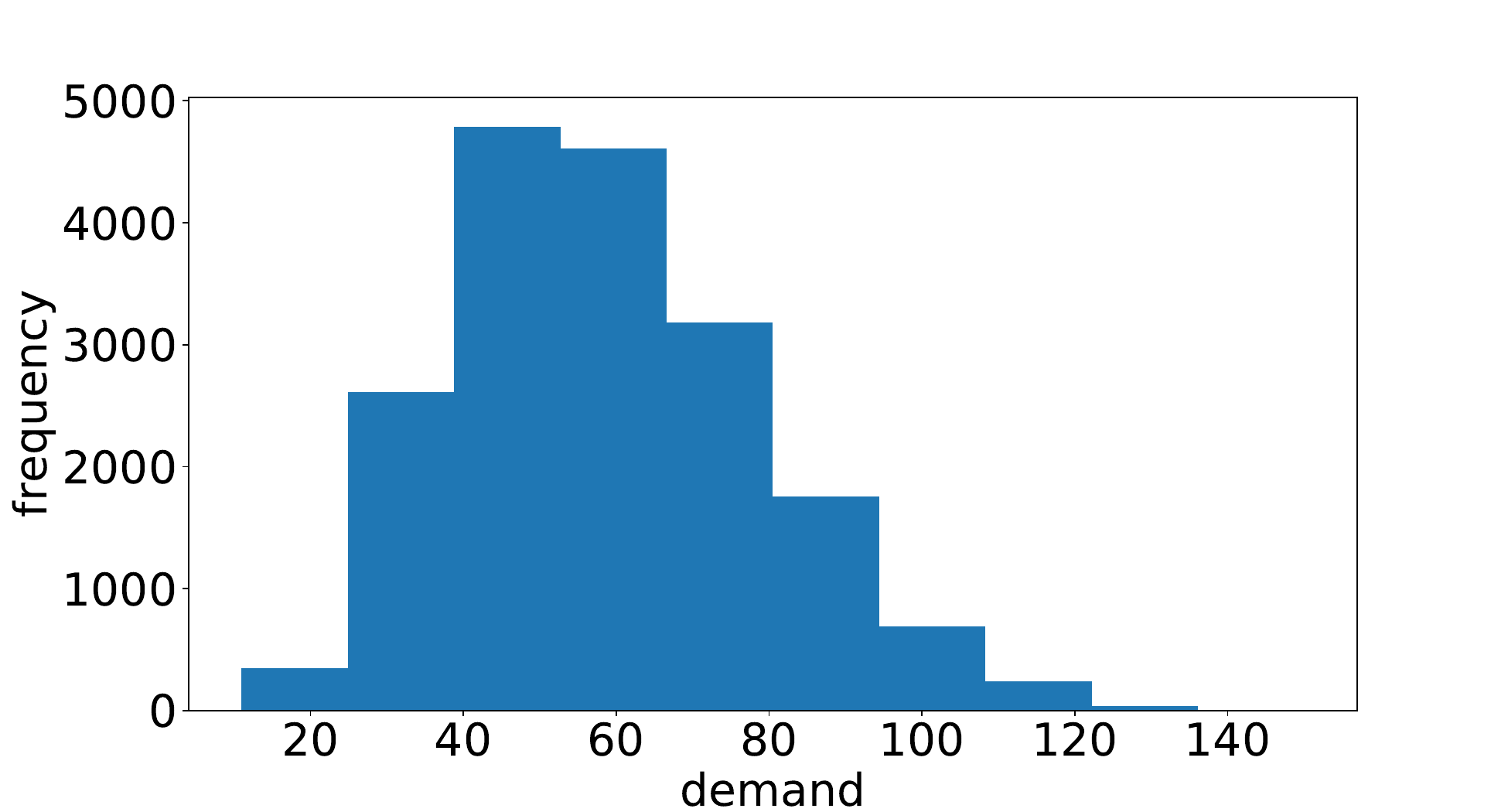}
	\vspace{-13pt}						
	\caption{{\small Item 46 (18260 records)}}
	\label{subfig:forecasting_demand46}	
\end{subfigure}	
\end{figure}
}

{\SingleSpacedXI
	\begin{table}[]
		\centering
		\caption{Results for real-world dataset I with scaled demands.}
		\label{tb:basket_results_scaled}
		\begin{tabular}{ccccc|ccc|ccc}
			Co-Player & & \multicolumn{3}{c}{Category 6} & \multicolumn{3}{c}{Category 13} & \multicolumn{3}{c}{Category 22} \\
			Type & Agent & {\tt -SRDQN}     & {\tt -BS}   & Gap (\%) & {\tt -SRDQN}     & {\tt -BS}  & Gap (\%) & {\tt -SRDQN}     & {\tt -BS}     & Gap (\%)  \\ \hline
			\multirow{5}{*}{{\tt BS-}} & \multicolumn{1}{c|}{R} & 4.25  & 4.08   & 4.05     & 4.16  & 4.07   & 2.38     & 3.10 & 3.01   & 3.00 \\
			& \multicolumn{1}{c|}{W} & 4.32  & 4.08   & 5.83    & 4.36  & 4.07   & 7.29    & 3.14 & 3.01   & 4.33 \\
			& \multicolumn{1}{c|}{D} & 4.64  & 4.08   & 13.68    & 4.24  & 4.07   & 4.19     & 3.17 & 3.01   & 5.11   \\
			& \multicolumn{1}{c|}{M} & 4.32  & 4.08   & 5.85    & 4.66  & 4.07   & 14.64     & 3.14 & 3.01   & 4.33    \\ 
			& \multicolumn{3}{l}{Average} & 7.35    &   &   & 7.12    &  &  & 4.20  \\
			\hline
			\multirow{5}{*}{{\tt Strm-}} & \multicolumn{1}{c|}{R} & 5.70  & 5.96   & -4.28     & 5.69 & 6.26  & -9.04   & 4.28 & 4.40   & -2.79 \\
			& \multicolumn{1}{c|}{W} & 6.56 & 7.01  & -6.47   & 6.89 & 7.04  & -2.12   & 4.93 & 5.26   & -6.14 \\
			& \multicolumn{1}{c|}{D} & 6.95 & 7.81  & -11.04   & 7.57 & 7.70  & -1.74   & 6.24 & 6.11   & 2.16  \\
			& \multicolumn{1}{c|}{M} & 8.83 & 9.16  & -3.57    & 9.74 & 8.99  & 8.31    & 7.10 & 7.29   & -2.56 \\ 
			& \multicolumn{3}{l}{Average} & -6.34    &   &   & -1.15   & &    & -2.33   \\
		\end{tabular}
	\end{table}			
} 

\subsection{Real-World Dataset II}\label{sec:results:forecasting_dataset}
The second real-world dataset includes five years of sales data for 50 items \citep{kaggle}, with 18260 records per item. Figures \ref{subfig:forecasting_demand5}--\ref{subfig:forecasting_demand46} show the demand histograms of the three randomly selected categories. We trained SRDQN for similar cases as in Section \ref{sec:results:basket_dataset}.

Table \ref{tb:forecast_results_scaled} presents the results for the three co-player types. With base-stock co-players, {\tt BS-SRDQN} always learns policies whose costs are close to those of {\tt BS-BS}, with an average gap of 7.44\%. In the case of Sterman co-players, in most cases (11 out of 12) {\tt Strm-SRDQN} attains a smaller cost than {\tt Strm-BS}, with an average gap of 10.12\%. 

{\SingleSpacedXI
	\begin{table}[]
		\centering
		\caption{Results of real-world dataset II with scaled demands.}
		\label{tb:forecast_results_scaled}
		\begin{tabular}{ccccc|ccc|ccc}
			Co-Player & & \multicolumn{3}{c}{Category-5} & \multicolumn{3}{c}{Category-34} & \multicolumn{3}{c}{Category-46} \\
			Type & Agent & {\tt -SRDQN}     & {\tt -BS}   & Gap (\%) & {\tt -SRDQN}     & {\tt -BS} & Gap (\%) & {\tt -SRDQN}     & {\tt -BS}   & Gap (\%)  \\ \hline
			\multirow{5}{*}{{\tt BS-}} 
			& \multicolumn{1}{c|}{R} & 3.23  & 3.08   & 4.88     & 2.87  & 2.70   & 6.21     & 3.28 & 3.13   & 4.77 \\
			& \multicolumn{1}{c|}{W} & 3.38  & 3.08   & 9.91    & 2.96  & 2.70   & 9.65    & 3.28 & 3.13   & 4.77 \\
			& \multicolumn{1}{c|}{D} & 3.40  & 3.08   & 10.58    & 2.95  & 2.70   & 9.28     & 3.29 & 3.13   & 5.07   \\
			& \multicolumn{1}{c|}{M} & 3.30  & 3.08   & 7.25    & 2.95  & 2.70   & 9.28     & 3.37 & 3.13   & 7.73    \\ 
			& \multicolumn{3}{l}{Average} & 8.15    &   &   & 8.60    &  &  & 5.57  \\
			\hline
			\multirow{5}{*}{{\tt Strm-}} 
			& \multicolumn{1}{c|}{R} & 5.70  & 5.35 & -5.26   & 4.84 & 4.78  & 1.33   & 5.26 & 5.49   & -4.26 \\
			& \multicolumn{1}{c|}{W} & 4.85 & 5.49  & -11.61   & 4.44 & 4.74  & -6.25   & 4.42 & 5.58   & -20.80 \\
			& \multicolumn{1}{c|}{D} & 5.29 & 6.35  & -16.68   & 4.48 & 5.24  & -14.46   & 5.53 & 6.87   & -19.60  \\
			& \multicolumn{1}{c|}{M} & 7.18 & 7.74  & -7.16    & 5.73 & 6.25  & -8.30    & 7.85 & 8.56   & -8.37 \\ 
			& \multicolumn{3}{l}{Average} & -10.18    &   &   & -6.92   & &    & -13.26   \\
		\end{tabular}
	\end{table}			
} 

Finally, given the 90\% confidence intervals, Table~\ref{tb:summary_win_numbers} summarizes the experiments of this section by counting the number of times that {\tt SRDQN} or {\tt BS} obtains a statistically  smaller cost, or that they are statistically equal. As shown, with {\tt Strm} co-players, {\tt SRDQN} obtains a statistically smaller cost in 75\% of cases (27 out of 36), equal costs in 22\% (8 cases out 36), and a larger cost in 2.7\% (1 case out 36). 
This is due to the fact that  the {\tt BS} player uses a fixed base-stock level for the entire episode, which may not be optimal for playing with {\tt Strm} agents who play irrationally.
On the other hand, {\tt SRDQN} learns the expected cost of the episode---the Q-value---for each state--action pair. So, at each time-step of the game, {\tt SRDQN} looks at its current state and, based on what is has learned from the behavior of the other agents in the same or similar states, it suggests the action that results in the smallest approximate sum of costs. 

In contrast, with {\tt BS} co-players, the three co-players follow rational policies and {\tt BS-BS} is optimal. Similarly, for each state--action pair, {\tt SRDQN} learns the sum of the costs for the entire episode. The learning is quite effective; it achieves statistically equal costs in 47.2\% of cases (19 out of 36), slightly larger costs in 50\%, and a smaller cost in one case (see the results for C(4,8) in Section \ref{sec:result:literature_cases}). 
Moreover, for {\tt Rand} co-players, {\tt SRDQN} performs almost as well as {\tt BS}; in 75\% of cases they are in a statistical tie, in 16.7\% {\tt SRDQN} outperforms {\tt BS}, and in 8.3\% {\tt BS} obtains smaller costs.

{\SingleSpacedXI
	\begin{table}
		\centering
		\small
		\caption{Summary of the performance for each algorithm with literature cases.}
		\label{tb:summary_win_numbers}
		\begin{tabular}{lccc|ccc|ccc}
			& \multicolumn{3}{c}{Literature Cases} &  \multicolumn{3}{c}{Real-World dataset I} & \multicolumn{3}{c}{Real-World dataset II} \\
			co-players & {\tt SRDQN} & {\tt BS} & tie  & {\tt SRDQN} & {\tt BS} & tie  & {\tt SRDQN} & {\tt BS} & tie \\ \hline
			{\tt BS-} & 1 & 6 & 5 & 0 & 5 & 7 & 0 & 7 & 5 \\
			{\tt Strm-} & 11 & 0 & 1 & 5 & 1 & 6 & 11 & 0 & 1\\
			{\tt Rand-} & 2 & 1 & 9 &  &  &  &  &  &  \\ \hline
		\end{tabular}
	\end{table}
}

To summarize, SRDQN performs well on well-scaled real-world datasets, regardless of the way the other agents play. 
The SRDQN agent learns to perform nearly as well as a base-stock policy when its co-players follow a base-stock policy (in which case it is optimal for the remaining agents to follow a base-stock policy). When playing with irrational co-players, it achieves a much smaller cost than a base-stock policy does.

\subsection{Sensitivity Analysis}\label{sec:results:sensitivity_analysis}
In this section, we explore the ability of a trained model to perform well even if it plays the game using settings that are different from those used during training. 
In particular, we analyze the sensitivity of a trained model to changes in the cost coefficients (in  Section \ref{sec:results:sensitivity_analysis:cpch}) and in the size of the historical data set (in Section \ref{sec:results:sensitivity_analysis:data_size}). 
 
\subsubsection{Parameters}\label{sec:results:sensitivity_analysis:cpch}

In order to analyze the sensitivity of the trained model to changes in the shortage and holding cost coefficients, we multiply $c_p$ and $c_h$ (independently) by $100+dev_{c_p}$ and $100+dev_{c_h}$ percent, where 
$\{dev_{c_p}, dev_{c_h}\} \in \{-90,-80,\ldots,-10,10,20,\ldots,100,125,150,175,200,250,300,400\}$.
For each of these new sets of cost coefficients, we use the models that were trained under the original cost coefficients for the demand distributions in Section~\ref{sec:result:literature_cases} and use these models to play the game for 50 episodes under the new costs. To generate the {\tt BS-BS} and {\tt Strm-BS} benchmarks, we obtained the optimal base-stock levels for each new set of costs by the Clark--Scarf algorithm, and had the agent play the game for 50 episodes using a base-stock policy with these base-stock levels. Note that for Sterman co-players, the Clark--Scarf algorithm does not provide the optimal base-stock level for the remaining player. However, searching for the best possible base-stock levels in {\tt Strm-BS} is quite expensive and there are 8112 cases to run. Therefore, due to a lack of computational resources, we simply used the Clark--Scarf algorithm to find base-stock levels for each case. This choice is justified by the fact that our goal in this section is to study the robustness of SRDQN rather than its performance compared to a base-stock policy.

\begin{figure}
	\caption{Sensitivity analysis with respect to cost, perturbing $c_p$ and $c_h$ with the ${\mathbb N}(10,2)$ (sub-figures \ref{fig:sensitivity_normal_bs} and \ref{fig:sensitivity_normal_strm}), the $\mathbb{U}[0,8]$ (sub-figures \ref{fig:sensitivity_uniform_bs} and \ref{fig:sensitivity_uniform_strm}), and the $C(4,8)$ (sub-figures \ref{fig:sensitivity_classic_bs} and \ref{fig:sensitivity_classic_strm}) demand distributions. Each sub-figure shows the result for one co-player type.
	}
	\vspace{-8pt}			
	\label{fig:sensitivity_analysis}			
	\centering
	\begin{subfigure}{0.47\textwidth}
		\centering
		\includegraphics[scale=0.32]{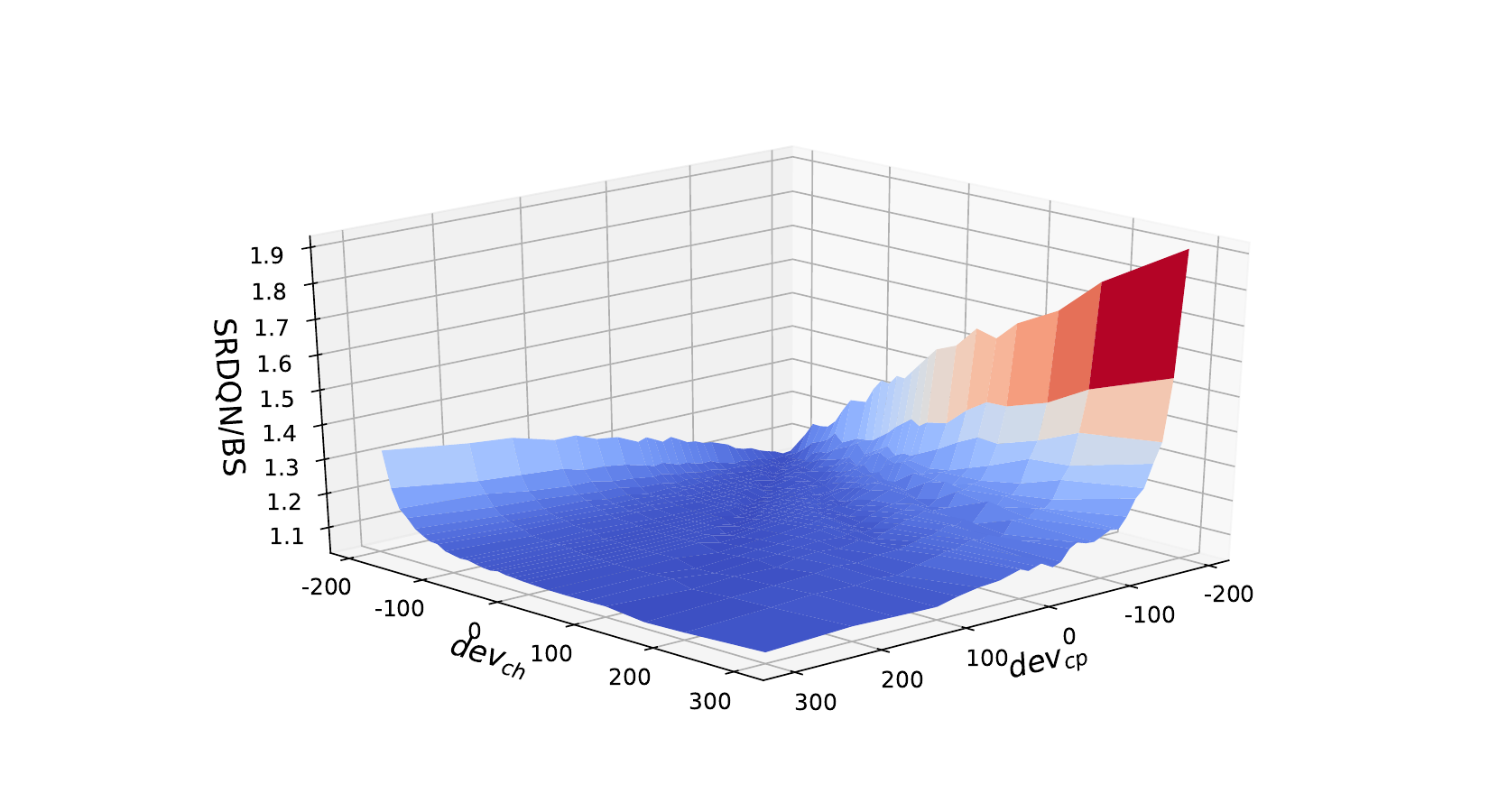}
		\vspace{-45pt}			
		\caption{Base-stock}
		\label{fig:sensitivity_normal_bs}		
	\end{subfigure}
	\hspace{9pt}
	\begin{subfigure}{0.47\textwidth}
		\centering
		\includegraphics[scale=0.32]{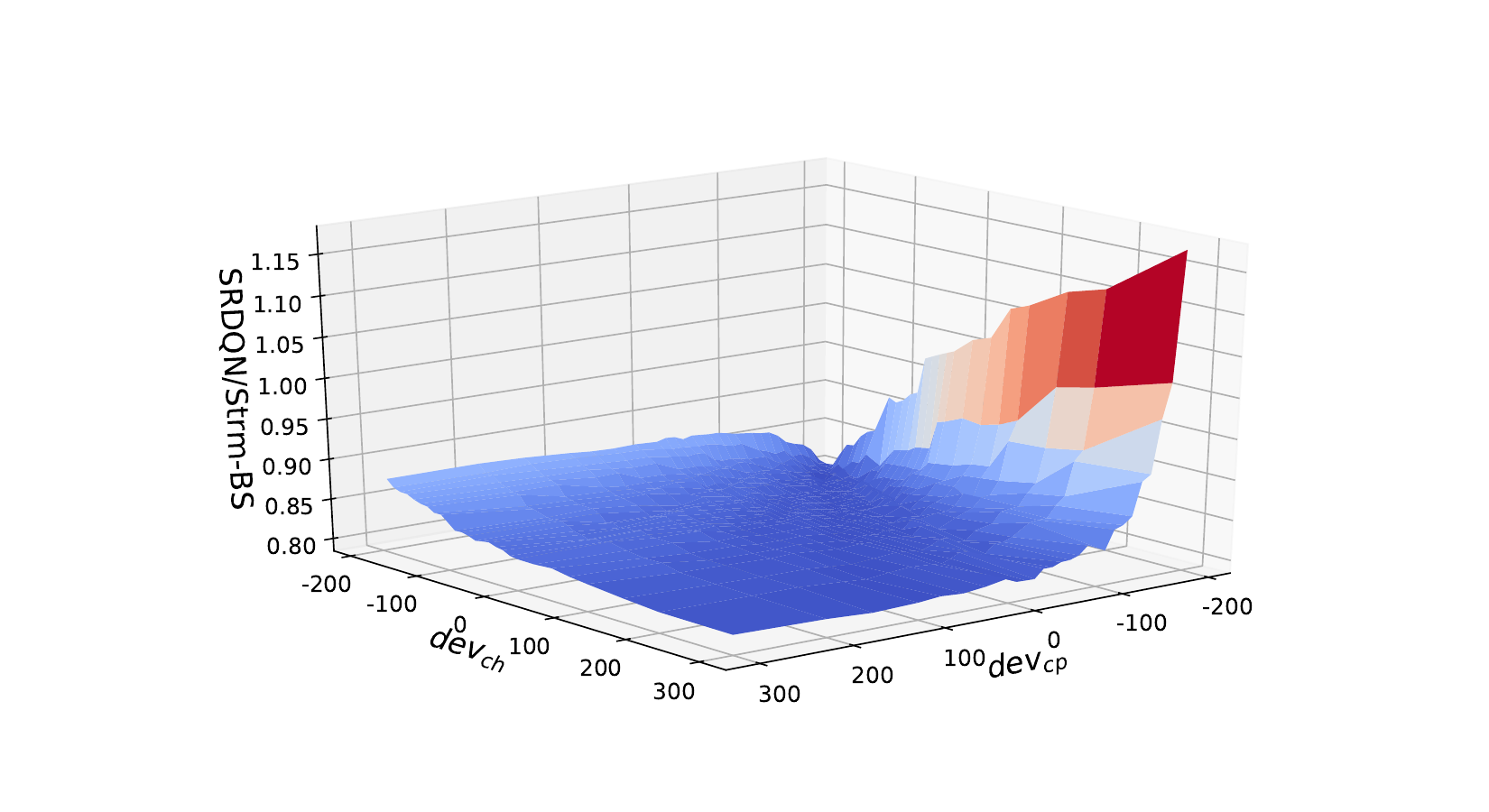}
		\vspace{-45pt}			
		\caption{Sterman}
		\label{fig:sensitivity_normal_strm}	
	\end{subfigure}	
	\vspace{-11pt}
	
	\begin{subfigure}{0.47\textwidth}
		\centering
		\includegraphics[scale=0.32]{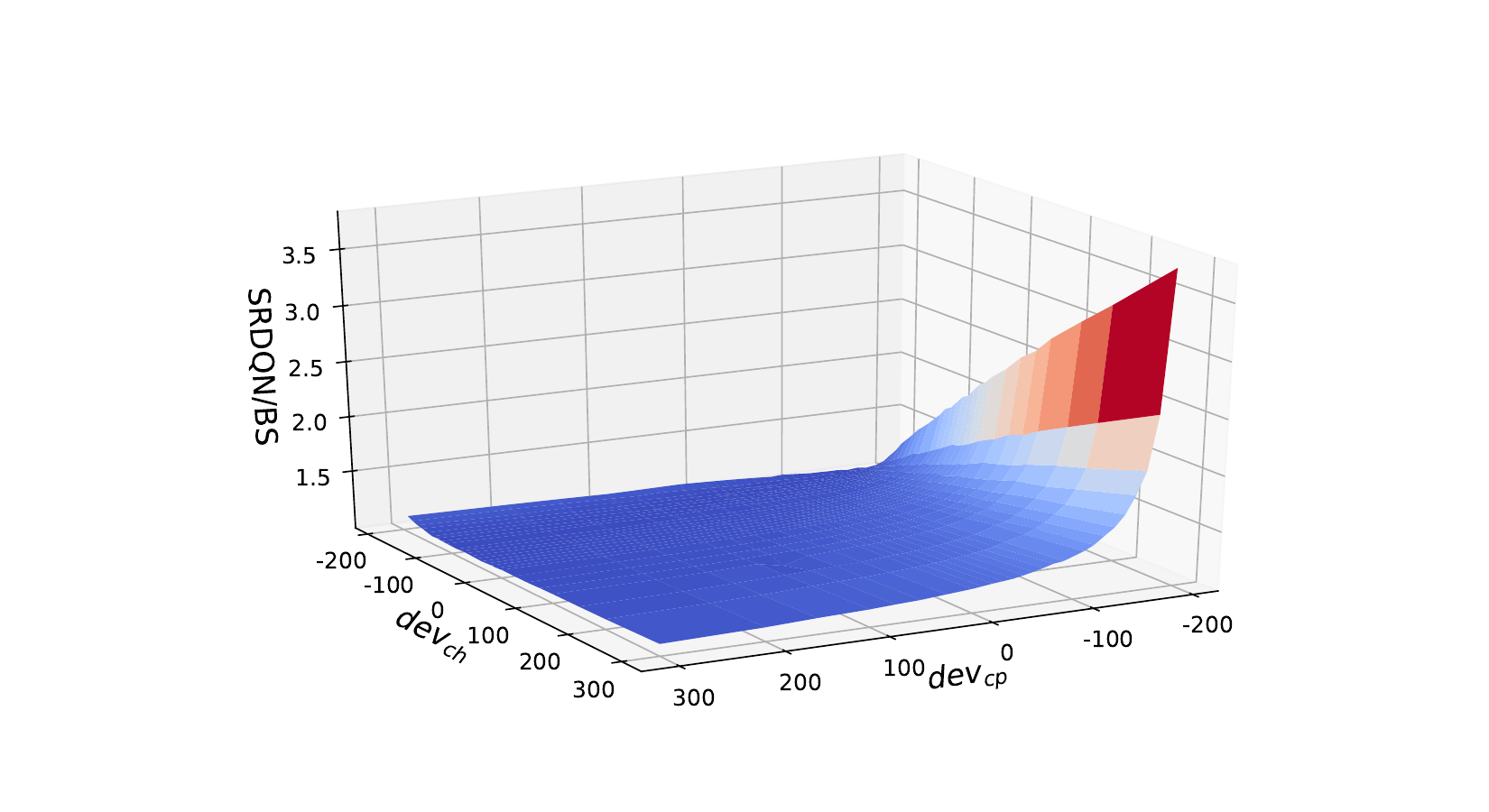}
		\vspace{-45pt}			
		\caption{Base-stock}
		\label{fig:sensitivity_uniform_bs}		
	\end{subfigure}
	\hspace{9pt}
	\begin{subfigure}{0.47\textwidth}
		\centering
		\includegraphics[scale=0.32]{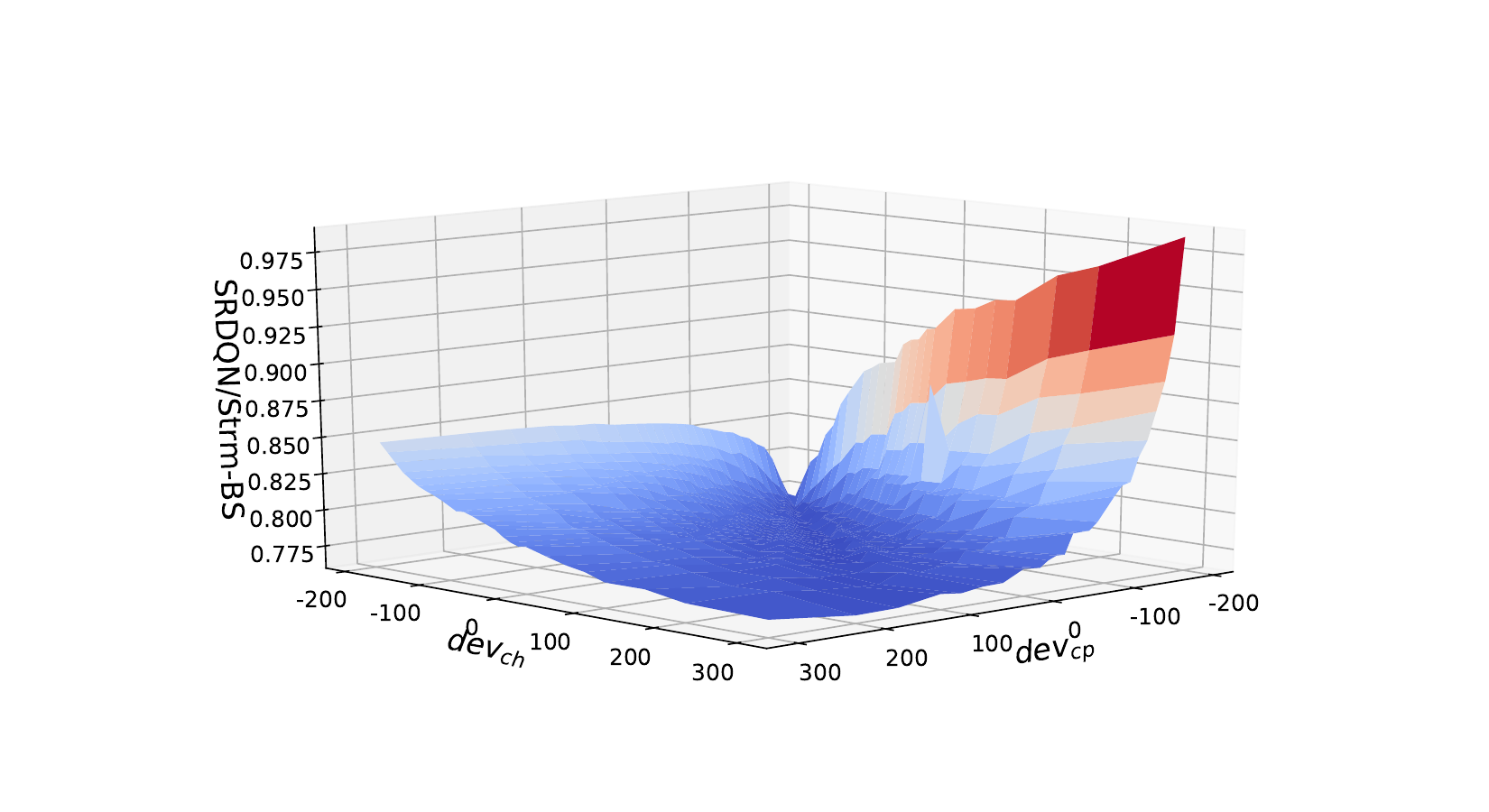}
		\vspace{-45pt}						
		\caption{Sterman}
		\label{fig:sensitivity_uniform_strm}	
	\end{subfigure}	
	\vspace{-11pt}

	\begin{subfigure}{0.47\textwidth}
		\centering
		\includegraphics[scale=0.32]{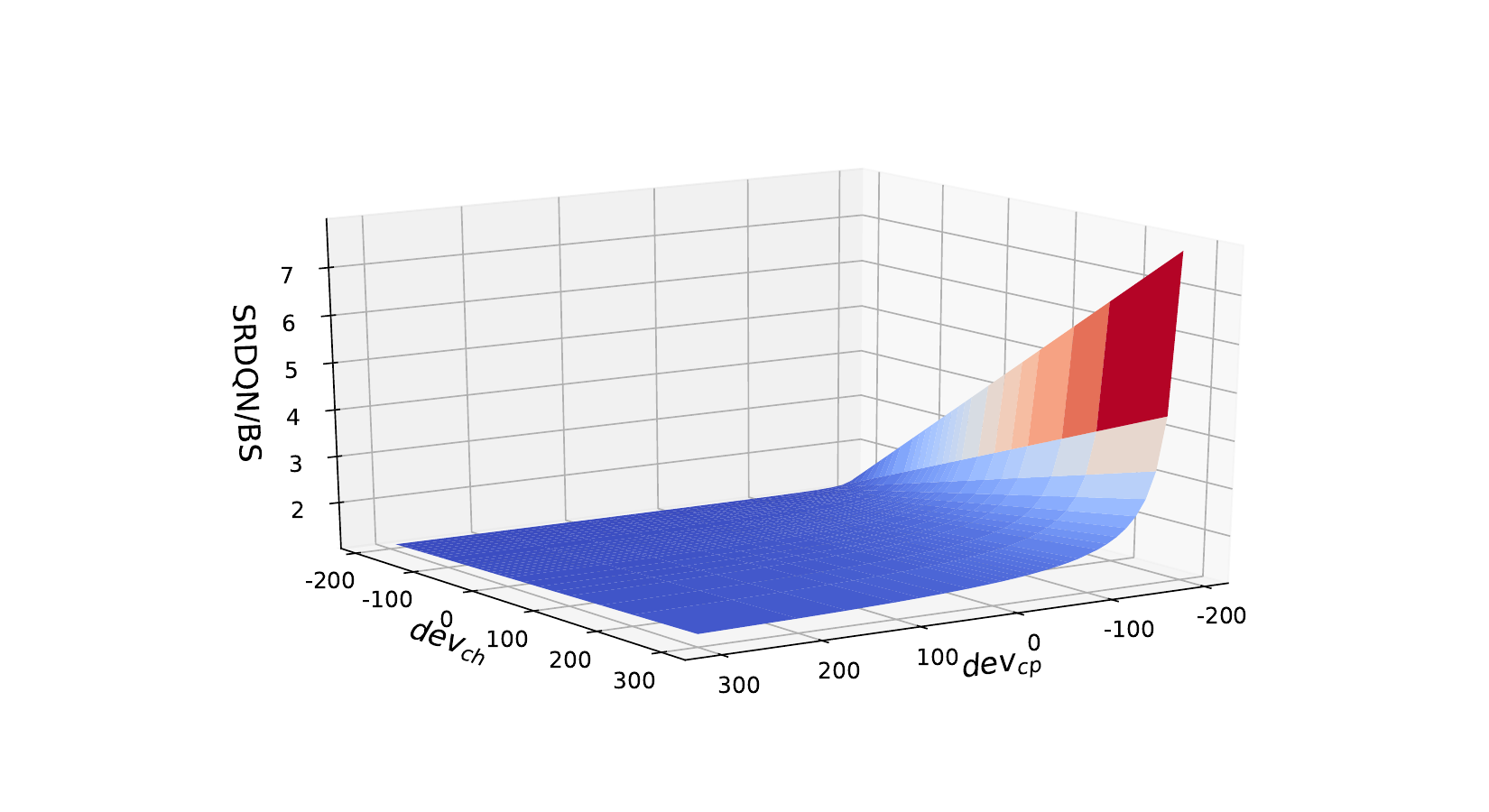}
		\vspace{-45pt}			
		\caption{Base-stock}
		\label{fig:sensitivity_classic_bs}		
	\end{subfigure}
	\hspace{8pt}
	\begin{subfigure}{0.47\textwidth}
		\centering
		\includegraphics[scale=0.32]{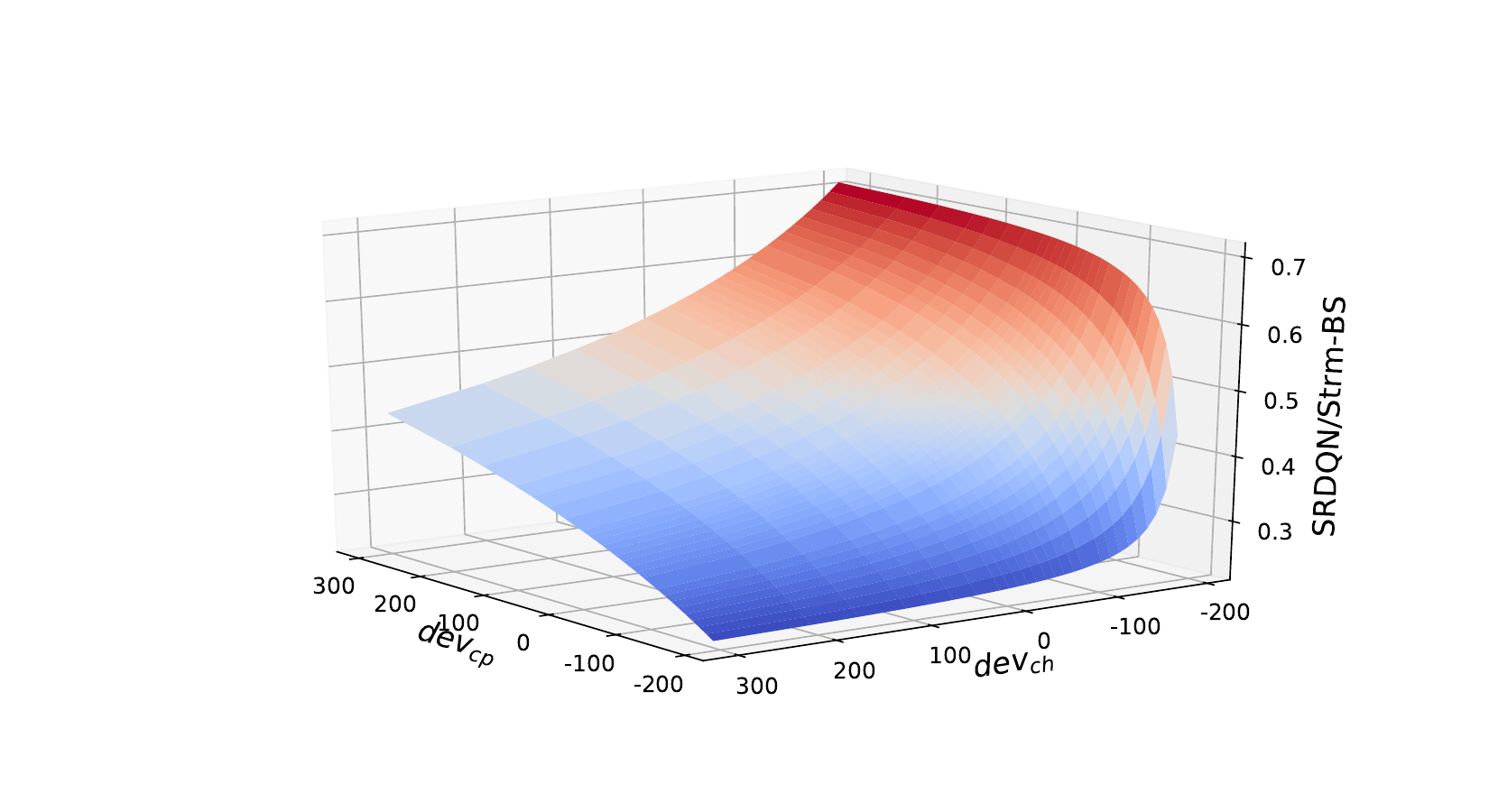}
		\vspace{-45pt}						
		\caption{Sterman}
		\label{fig:sensitivity_classic_strm}	
	\end{subfigure}	
\end{figure}

Figure \ref{fig:sensitivity_analysis} show the results for all demand distributions. For each pair of $dev_{c_p}$ and $dev_{c_h}$ values, the figure plots the ratio of the {\tt BS-SRDQN} or {\tt Strm-SRDQN} cost to the {\tt BS-BS} or {\tt Strm-BS} cost, averaged over the 50 episodes. 
From the figures, it is clear that our algorithm works well without any additional training for a wide range of cost values. When $dev_{c_p}/dev_{c_h} \approx 1$, i.e., the cost parameters scale roughly in sync, we get almost the same performance that we got from the fully trained model.  The reason for this is that $dev_{c_p}/dev_{c_h} \approx 1$ results in same ratio of Q-values for each action, and as a result the action-selection procedure is not affected by the cost coefficients' deviation. As $dev_{c_p}/dev_{c_h}$ moves away from 1, the cost ratio increases, relatively slowly. The sharpest increase in the SRDQN/BS ratio occurs when the stockout cost decreases but the holding cost increases. The reason for this behavior is that the model is trained assuming the stockout cost is greater than the holding cost; when the stockout cost decreases and the holding cost increases, the trained Q-values become less relevant and the trained model fails to provide good results.

\subsubsection{Size of Historical Data Set} \label{sec:results:sensitivity_analysis:data_size} 

In many real-world supply chain settings, the size of the available historical data set is much smaller than what is required to train a SRDQN model. In Section \ref{sec:results:basket_dataset}, we tested our method on a real-world data set and concluded that it still performed well if we train it on simulated data from an empirical distribution that is estimated from the available historical data. In this section, we test this approach more rigorously, explicitly comparing its performance with that of a model that has been trained on the full data set. 

To this end, we we generated 100 observations from the demand distributions used in Section \ref{sec:result:literature_cases}, i.e., ${\mathbb U}[0,8]$ and ${\mathbb N}(10,2)$. (Note that we excluded the ``classic'' demand distribution $C(4,8)$, since the demand pattern is fixed in that distribution). Then, for each demand distribution we obtained a histogram of those 100 data points and generated 60000 episodes using the empirical distribution implied by the histogram. We trained two models (one for each distribution) on these new data sets. We then tested the performance of these ``low-observation'' models under the same test data sets that we used in Section \ref{sec:result:literature_cases}). By comparing the performance of the low-observation models with that of the ``full-observation'' models (i.e., the models trained on the full data sets in Section \ref{sec:result:literature_cases}), we can explore whether a massive historical data set is truly required for training the model.

\begin{figure}
	\caption{Comparison of full-observation and low-observation training data for two demand distributions.}
	\label{fig:small_observation}			
	\centering
	\begin{subfigure}{0.3\textwidth}
		\centering
		\includegraphics[scale=0.17]{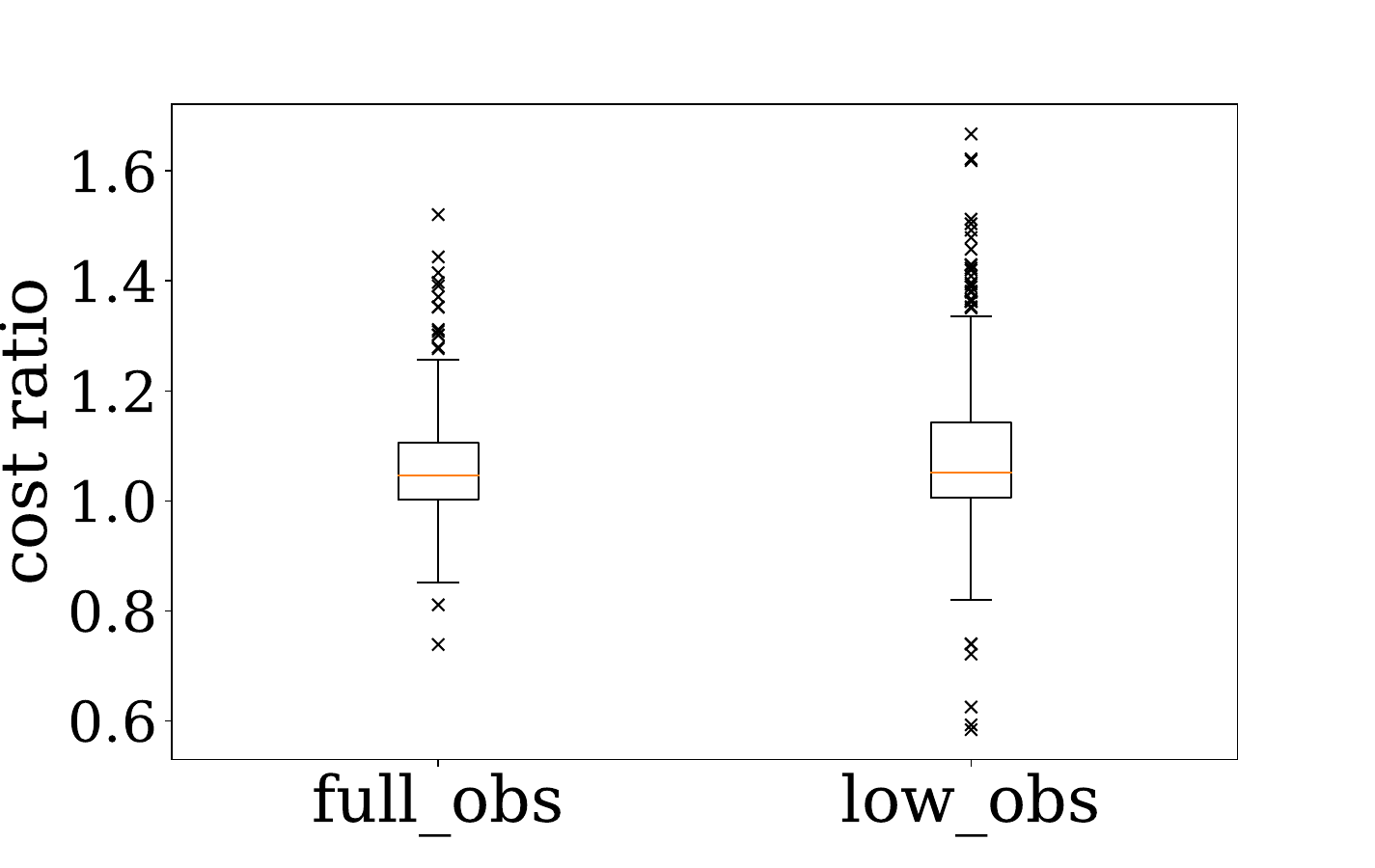}
		\vspace{-10pt}			
		\caption{Normal distribution}
		\label{fig:small_observation_normal}		
	\end{subfigure}
	\begin{subfigure}{0.3\textwidth}
		\centering
		\includegraphics[scale=0.17]{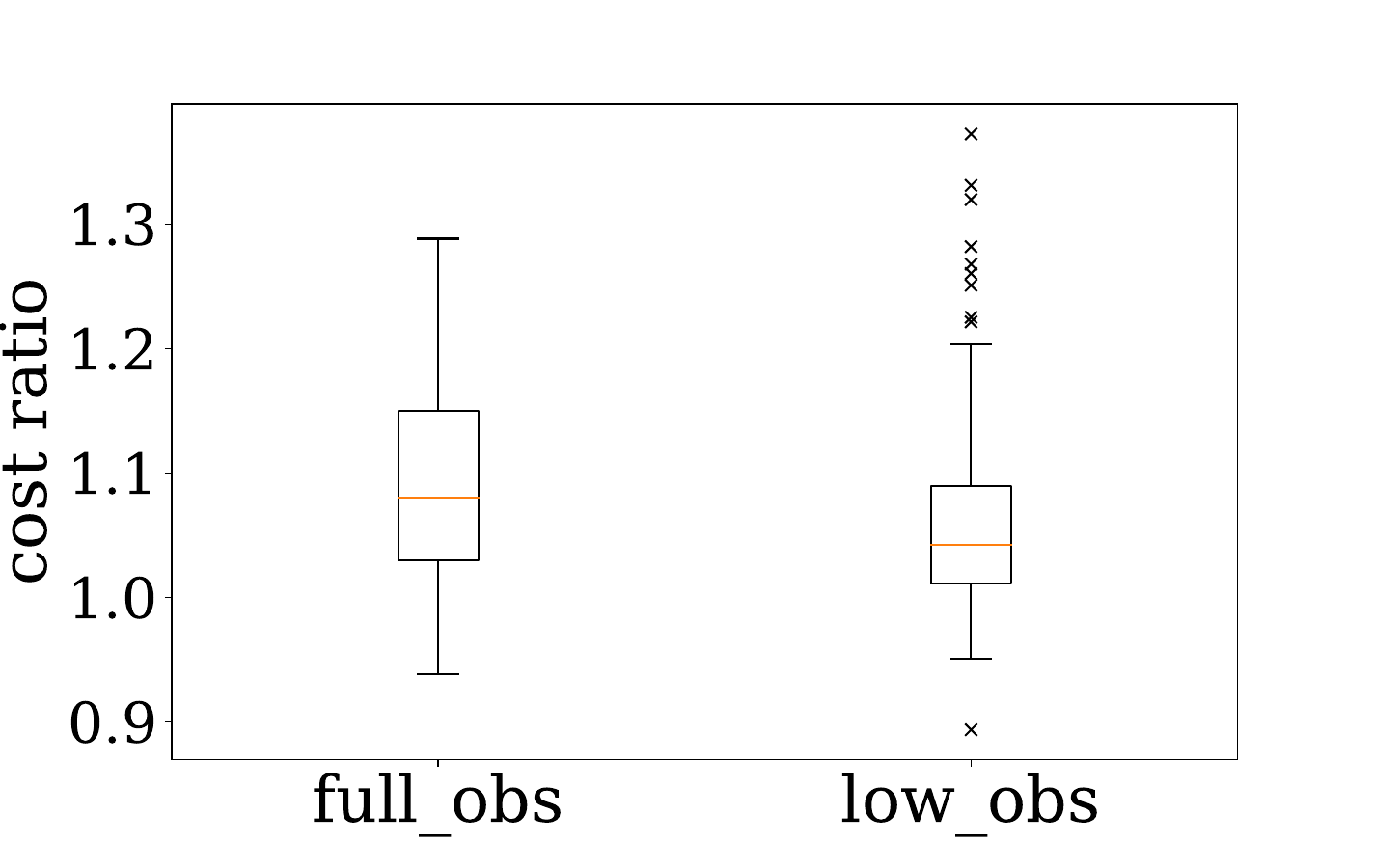}
		\vspace{-10pt}						
		\caption{Uniform distribution}
		\label{fig:small_observation_uniform}	
	\end{subfigure}	
\end{figure}

Our null hypothesis is that the low-observation and full-observation models have equal performance.  In order to test this null hypothesis, Figure \ref{fig:small_observation} provides box plots for both the full-observation and low-observation models, when the SRDQN plays with base-stock co-players, for both normal and uniform demands. 

Although the low-observation model has more outliers, the performance of two models is quite similar; the confidence intervals overlap, and we cannot reject the null hypothesis that the two models have equal performance. So, even with only 100 historical observations, our model works well, with performance that is statistically similar to that in Section \ref{sec:result:literature_cases}.

\subsection{Faster Training through Transfer Learning}\label{sec:results:transfer_learning}

We trained an SRDQN network with shape	$[50, 180, 130, 61, 5]$, $m=10$, $\beta=20$, and $C=10000$ for each agent, with the same holding and stockout costs and action spaces as in Section \ref{sec:uniform_0_2_result}, using 60000 training episodes, and used these as the base networks for our transfer learning experiment. (In transfer learning, all agents should have the same network structure to share the learned network among different agents.) 
The remaining agents use base-stock levels obtained by the Clark--Scarf algorithm. 
		
Table \ref{tb:transfer_learning_result} shows a summary of the results of the six cases discussed in Section \ref{sec:transfer_learning}. 
The first set of columns indicates the holding and shortage cost coefficients, the size of the action space, as well as the demand distribution and the co-players' policy for the base agent (first row) and the target agent (remaining rows).
The ``Gap'' column indicates the average gap between the cost of {\tt BS-SRDQN} and the cost of {\tt BS-BS}; in the first row, it is analogous to the 2.31\% average gap reported in Section~\ref{sec:results:dnn_vs_base_stock}. The average gap is relatively small in all cases, which shows the effectiveness of the transfer learning approach. 
Moreover, this approach is more efficient than training the agents from scratch, as demonstrated in the last column, which reports the average CPU times for all agent. In order to get the base agents, we did hyper-parameter tuning and trained 140 instances to get the best possible set of hyper-parameters, which resulted in a total of 28,390,987 seconds of training. However, using the transfer learning approach, we do not need any hyper-parameter tuning; we only need to check which source agent and which $k$ provides the best results. This requires only 12 instances to train and resulted in an average training time (across cases 1--4) of 1,613,711 seconds---17.6 times faster than training the base agent. 
Additionally, in case 5, in which a normal distribution is used, full hyper-parameter tuning took 20,396,459 seconds, with an average gap of 4.76\%, which means transfer learning was 16.6 times faster on average. We did not run the full hyper-parameter tuning for the instances of case 6, but it is similar to that of case 5 and should take similar training time, and as a result a similar improvement from transfer learning.  Thus, once we have a trained agent $i$ with a given set $P_1^i$ of parameters, demand $D_1$ and co-players' policy $\pi_1$, we can efficiently train a new agent $j$ with parameters $P_2^j$, demand $D_2$ and co-players' policy $\pi_2$. 

	{\SingleSpacedXI
	\begin{table}[]
		\centering
		\caption{Results of transfer learning when $\pi_1$ is {\tt BS-SRDQN} and $D_1$ is $\mathbb{U}[0,2]$.}
		\label{tb:transfer_learning_result}
		\begin{tabular}{lccccccc|cc}
			& \multicolumn{4}{c}{(Holding, Shortage) Cost Coefficients} & \multirow{2}{*}{\begin{tabular}[c]{@{}c@{}} $|\mathcal{A}|$ \end{tabular}} & \multirow{2}{*}{$D_2$} & \multirow{2}{*}{$\pi_2$} & Gap & CPU Time \\ \cline{2-5}			
			& \multicolumn{1}{c}{R} & \multicolumn{1}{c}{W} & \multicolumn{1}{c}{D} & \multicolumn{1}{c}{M} & & & & (\%)  & (sec) \\ \hline
			\multicolumn{1}{l|}{Base agent} & (2,2)  & (2,0) & (2,0)  & (2,0)  & 5 & ${\mathbb U}[0,2]$ & {\tt BS-SRDQN} & 2.31  & 28,390,987 \\ \hline
			\multicolumn{1}{l|}{Case 1} & (2,2)  & (2,0) & (2,0)  & (2,0)  & 5 & ${\mathbb U}[0,2]$ & {\tt BS-SRDQN} &  6.06 & 1,593,455 \\
			\multicolumn{1}{l|}{Case 2}     & (5,1)   & (5,0)    & (5,0)      & (5,0)       & 5   & ${\mathbb U}[0,2]$ & {\tt BS-SRDQN} & 6.16 & 1,757,103   \\
			\multicolumn{1}{l|}{Case 3}    & (2,2)   & (2,0)    & (2,0)      & (2,0) & 11  & ${\mathbb U}[0,2]$ & {\tt BS-SRDQN} & 10.66  & 1,663,857  \\
			\multicolumn{1}{l|}{Case 4}   & (10,1)   & (10,0)    & (10,0)      & (10,0)       & 11  & ${\mathbb U}[0,2]$ & {\tt BS-SRDQN} & 12.58 &  1,593,455 \\ 
			\multicolumn{1}{l|}{Case 5}   & (1,10)   & (0.75,0)    & (0.5,0)      & (0.25,0)       & 11  & ${\mathbb N}(10,2^2)$ & {\tt BS-SRDQN} & 17.41 &  1,234,461 \\ 
			\multicolumn{1}{l|}{Case 6}   & (1,10)   & (0.75,0)    & (0.5,0)      & (0.25,0)       & 11  & ${\mathbb N}(10,2^2)$ & {\tt Strm-SRDQN} & -38.20 &  1,153,571 \\ 			
			\multicolumn{1}{l|}{Case 6}   & (1,10)   & (0.75,0)    & (0.5,0)      & (0.25,0)       & 11  & ${\mathbb N}(10,2^2)$ & {\tt Rand-SRDQN} & -0.25 &  1,292,295 \\ \hline
		\end{tabular}
	\end{table}	
	} 
	
\begin{figure}
		\caption{Results of transfer learning for case 4 (different agent, cost coefficients, and action space). In the four subfigures, the base agent $b$ and number of layers $k$ are: (a) $b=3,k=1$, (b) $b=1,k=1$, (c) $b=3,k=2$, and (d) $b=4,k=2$. }
		\label{fig:agents_cp10ch1_action5}			
		\centering
		\begin{subfigure}{0.24\textwidth}
			\centering
			\includegraphics[scale=0.16]{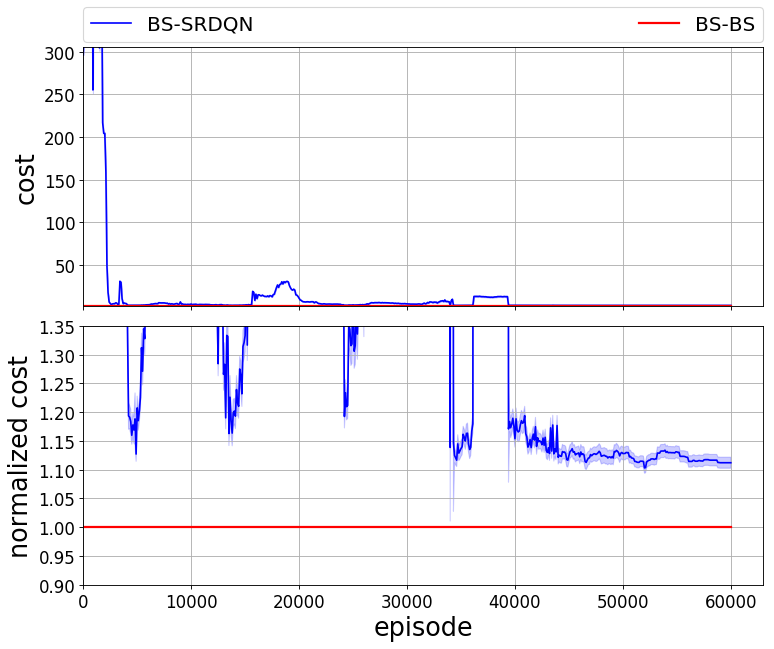}
			\vspace{-30pt}			
			\caption{Retailer}
			 \label{fig:agent_cp10ch1_action5_3-6-1}		
		\end{subfigure}
		\begin{subfigure}{0.24\textwidth}
			\centering
			\includegraphics[scale=0.16]{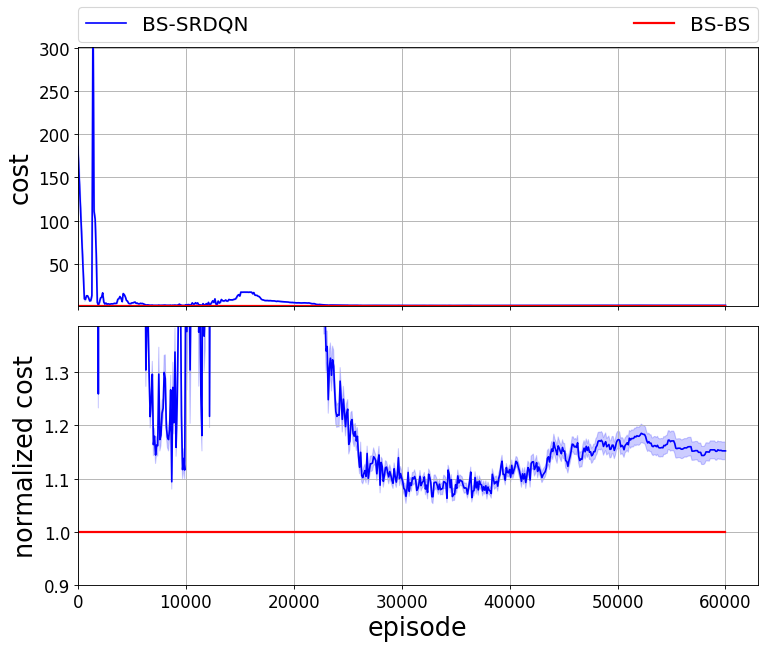}
			\vspace{-30pt}						
			\caption{Wholesaler}
			 \label{fig:agent_cp10ch1_action5_4-6-1}	
		\end{subfigure}	
		\begin{subfigure}{0.24\textwidth}
			\centering
			\includegraphics[scale=0.16]{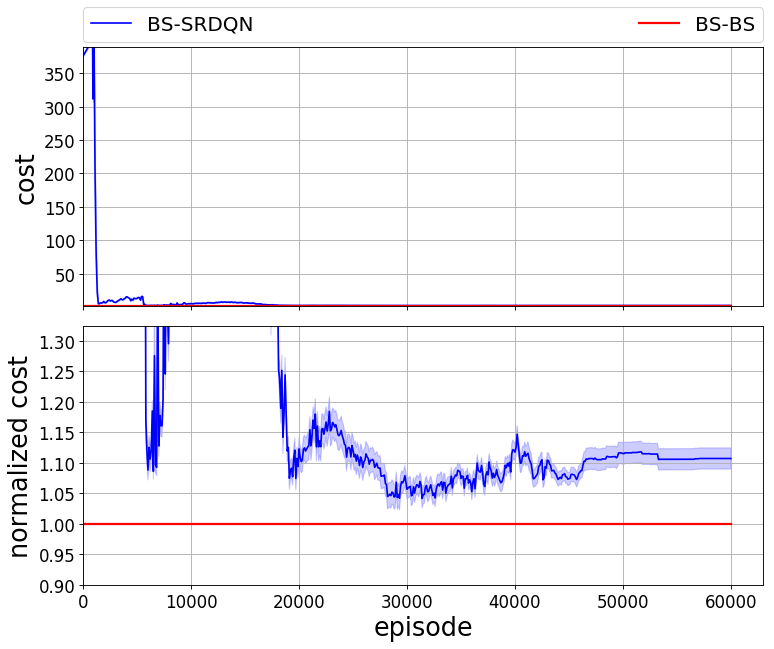}
			\vspace{-30pt}						
			\caption{Distributor}
			 \label{fig:agent_cp10ch1_action5_5-6-2}		
		\end{subfigure}		
		\begin{subfigure}{0.24\textwidth}
			\centering
			\includegraphics[scale=0.16]{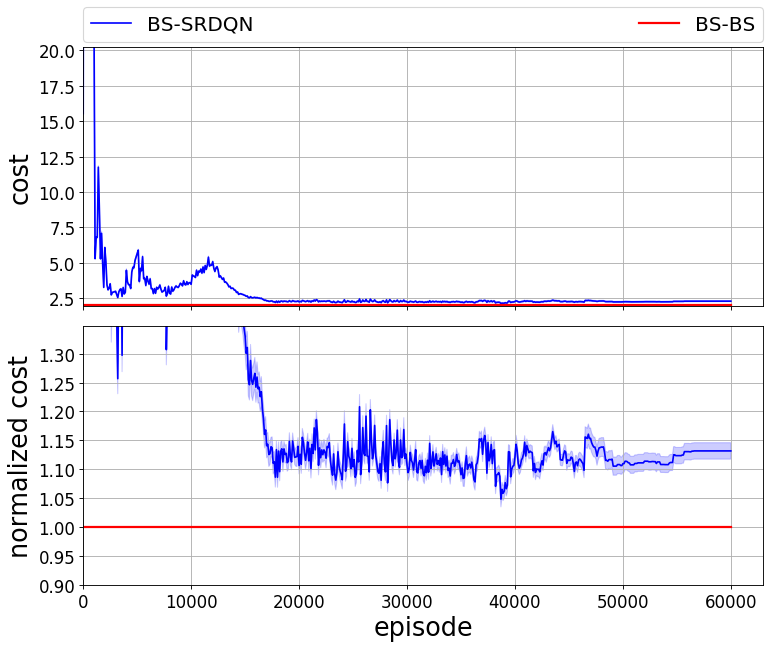}
			\vspace{-30pt}						
			\caption{Manufacturer}
			 \label{fig:agent_cp10ch1_action5_6-6-2}		
		\end{subfigure}		
	\end{figure}

	In order to get more insights about the transfer learning process, Figure \ref{fig:agents_cp10ch1_action5} shows the results of case 4, which is a quite complex transfer learning case that we test for the beer game.
	The target agents have holding and shortage costs (10,1), (10,0), (10,0), and (10,0) for agents 1 to 4, respectively; and each agent can select any action in $\{-5,\dots,5\}$. Each caption reports the base agent (shown by {\tt b}) and the value of $k$ used. 
	Compared to the original procedure (see Figure \ref{fig:dqn_vs_three_optimal}), i.e., $k=0$, the training is less noisy and after a few thousand non-fluctuating training episodes, it converges into the final solution.
	The resulting agents obtain costs that are close to those of {\tt BS-BS}, with a $12.58\%$ average gap compared to the {\tt BS-BS} cost. (The details of the other cases are provided in Sections 
	I.1---I.5 
 of the online supplement.)

Finally, Table~\ref{tb:transfer_learning_comparisons} explores the effect of $k$ on the tradeoff between training speed and solution accuracy. As $k$ increases, the number of trainable variables decreases and, not surprisingly, the CPU times are smaller but the costs are larger. For example, when $k=3$, the training time is 46.89\% smaller than the training time when $k=0$, but the solution cost is 17.66\% and 0.34\% greater than the {\tt BS-BS} policy, compared to 4.22\% and -11.65\% for $k=2$.

	{\SingleSpacedXI
	\begin{table}[]
		\centering
		\caption{Savings in computation time due to transfer learning. First row provides average training time among all instances. Third row provides average of the best obtained gap in cases for which an optimal solution exists. Fourth row provides average gap among all transfer learning instances, i.e., cases 1--6. \label{tb:transfer_learning_comparisons}}
		{}
		\begin{tabular}{l|cccc}
			& $k=0$ & $k=1$ &  $k=2$ & $k=3$ \\ \hline
			Training time  & 185,679   & 126,524 &  118,308 &  107,711  \\
			Decrease in time compared to $k=0$  & --- & 37.61\% & 41.66\%   & 46.89\%  \\ 
			Average gap in cases 1--4  & 2.31\% & 4.39\% &  4.22\% & 17.66\%  \\ 
			Average gap in cases 1--6  & --- & -15.95\% &  -11.65\% & 0.34\%  \\ 
			\hline
		\end{tabular}	
	\end{table}
	}
	
	To summarize, transferring the acquired knowledge between agents is much more efficient than training the agents from scratch. 
	The target agents achieve costs that are close to those of {\tt BS-BS}, and they achieve smaller costs than {\tt Strm-BS} and {\tt Rand-BS}, regardless of the dissimilarities between the source and the target agents. 
The training of the target agents starts from relatively small cost values, the training trajectories are stable and fairly non-noisy, and they quickly converge to a cost value close to that of {\tt BS-BS} or smaller than {\tt Strm-BS} and {\tt Rand-BS}. Even when the action space and costs for the source and target agents are different, transfer learning is still quite effective, resulting in a 12.58\% gap compared to {\tt BS-BS}. 
This is an important result, since it means that if the settings change---either within the beer game or in real supply chain settings---we can train new SRDQN agents much more quickly than we could if we had to begin each training from scratch. 	

\section{Conclusion and Future Work}\label{sec:beer_conclusion}
In this paper, we consider the beer game, a decentralized, multi-agent, cooperative supply chain problem. 
A base-stock inventory policy is known to be optimal for special cases, but once some of the agents do not follow a base-stock policy (as is common in real-world supply chains), the optimal policy of the remaining players is unknown. 
To address this issue, we propose an algorithm based on deep Q-networks. Since we use non-linear approximator, there is no theoretical guarantees to attain the global optimal, although it works well in practice.
It obtains near-optimal solutions when playing alongside agents who follow a base-stock policy and performs much better than a base-stock policy when the other agents use a more realistic model of ordering behavior. The training of the model might be slow, although the execution is very fast.
Furthermore, the algorithm does not require knowledge of the demand probability distribution and uses only historical data. 

To reduce the computation time required to train new agents with different cost coefficients or action spaces, we propose a transfer learning method. Training new agents with this approach takes less time since it avoids the need to tune hyper-parameters and has a smaller number of trainable variables. Moreover, it is quite powerful, resulting in beer game costs that are close to those of fully-trained agents while reducing the training time by an order of magnitude.

A natural extension of this paper is to apply our algorithm to supply chain networks with other topologies, e.g., distribution networks. 
Performing a theoretical analysis of the algorithm would be an interesting research direction, to determine the rate of convergence or worst-case error bounds. A more comprehensive numerical study would also shed additional light on the performance of our approach, especially if the study can be performed with real-world data. Another important extension is having multiple learnable agents. Finally, developing algorithms capable of handling continuous action spaces will improve the accuracy of our algorithm.

\ACKNOWLEDGMENT{%
This research was supported in part by NSF grant \#CMMI-1663256, \#CCF-1618717, \#CCF-1740796, XSEDE-DDM180004, XSEDE-IRI180020, and Intel AI DevCloud. This support is gratefully acknowledged.	
}

\bibliographystyle{abbrvnat}
\bibliography{beer_game}


\clearpage

{
\centering{\Large Online Supplements for A Deep Q-Network for the Beer Game: Reinforcement Learning for Inventory Optimization \\}
}
\vspace{30pt}

\begin{APPENDICES}	

	\section{$\epsilon$-Greedy Algorithm}\label{sec:appdx:epsilon_greedy}
	The $\epsilon$-greedy algorithm chooses a random actions with probability $\epsilon$, and with probability $1-\epsilon_t$ chooses the action greedily by evaluating a given function. In the context of the Q-Learning algorithm, with probability $\epsilon_t$ in time $t$, the algorithm chooses an action randomly, and with probability $1-\epsilon_t$, it chooses the action with the highest cumulative action value, i.e., $a_{t+1} = \text{argmax}_{a} Q(s_{t+1},a)$.
	The random selection of actions, called exploration, allows the agent to explore the solution space and gives an optimality guarantee to the Q-Learning algorithm if $\epsilon_t \rightarrow 0$ when $t \rightarrow \infty$ \citep{sutton1998reinforcement}.

	\section{Sterman Formula Parameters}\label{sec:appdx:sterman parameters}
	The computational experiments that use {\tt Strm} agents calculate the order quantity using $ q^i_t = \max \{0, AO^{i-1}_{t+1} + \alpha^i (IL^i_t - a^i) + \beta^i (OO_t^i - b^i)\}$, adapted from \cite{sterman1989modeling}, 
	%
	where $\alpha^i$, $a^i$, $\beta^i$, and $b^i$ are the parameters corresponding to the inventory level and on-order quantity. The idea is that the agent sets the order quantity equal to the demand forecast plus two terms that represent adjustments that the agent makes based on the deviations between its current inventory level (resp., on-order quantity) and a target value $a^i$ (resp., $b^i$).  We set $a^i=\mu_d$, where $\mu_d$ is the average demand; $b^i=\mu_d(l_i^{fi} + l_i^{tr})$; $\alpha^i = -0.5$; and $\beta^i = -0.2$ for all agents $i=1,2,3,4$. The negative $\alpha$ and $\beta$ mean that the player over-orders when the inventory level or on-order quantity fall below the target value $a_i$ or $b_i$.

	\section{Determining the Value of {\bf \em m} in the State Definition}\label{sec:appnd:choosing_m}
	As noted above, the DNN maintains information from the most recent $m$ periods in order to keep the size of the state variable fixed and to address the issue with the delayed observation of the reward. In order to select an appropriate value for $m$, one has to consider the value of the lead times throughout the game.
	First, when agent $i$ takes action $a^i_t$ at time $t$, it does not observe its effect until at least $l^{tr}_i + l^{in}_i$ periods later, when the order may be received. Moreover, agent $i+1$ may not have enough stock to satisfy the order immediately, in which case the shipment is delayed and in the worst case agent $i$ might not observe the corresponding reward $r^i_t$ until $\sum_{j=i}^{4} (l^{tr}_j + l^{in}_j)$ periods later.
	However, one needs the reward $r_t^i$ to evaluate the action $a^i_t$ taken. Thus, ideally $m$ should be chosen at least as large as $\sum_{j=1}^{4} (l^{tr}_j + l^{in}_j)$. 
	On the other hand, this value can be large and selecting a large value for $m$ results in a large input size for the DNN, which increases the training time. Therefore, selecting $m$ is a trade-off between accuracy and computation time, and $m$ should be selected according to the required level of accuracy and the available computation power. In our numerical experiment, $\sum_{j=1}^{4} (l^{tr}_j + l^{in}_j) = 15$ or $16$, and we test $m\in\{5,10\}$.
	
	\section{Why Can't Standard DQN Solve the Beer Game?}\label{sec:appnd:why_dqn_does_not_work}
	
	One naive approach to extend the DQN algorithm to solve the beer game is to use multiple DQNs, one for each agent. However, using DQN as the decision maker for each agent results in a competitive game in which each DQN agent plays independently to minimize its own cost. 
	For example, consider a beer game in which players 2, 3, and 4 each have a stand-alone, well-trained DQN 
	and the retailer (stage 1) uses a base-stock policy to make decisions. If the holding costs are positive for all players and the stockout cost is positive only for the retailer (as is common in the beer game), then the DQN at agents 2, 3, and 4 will return an optimal order quantity  of 0 in every period, since on-hand inventory hurts the objective function for these players, but stockouts do not. 
	This is a byproduct of the independent DQN agents minimizing their own costs without considering the total cost, which is obviously not an optimal solution for the system as a whole.

	\section{Experience Replay}\label{sec:appnd:experience_replay}
	The DNN algorithm requires a mini-batch of input and the corresponding set of output values to learn the Q-values. 
	Since SRDQN is an off-policy algorithm, we can use the new state $s_{t+1}$, the current state $s_t$, the action $a_t$ taken, and the observed reward $r_t$, in each period $t$. This information can provide the required set of input and output for the DNN; however, the resulting sequence of observations from the RL results in a non-stationary dataset with a strong correlation among consecutive records. This makes the DNN and, as a result, the RL prone to over-fitting the previously observed records and may even result in a diverging approximator \citep{sutton1998reinforcement}. 
	To avoid this problem, we use {\em experience replay} $E^i$ \citep{lin1992self} for agent $i$: We add observation $e^i_t=(s^i_t,a^i_t,r^i_t,s^i_{t+1})$ iteratively into $E_i$, and in each training step, we take a random mini-batch of the experience to break correlations among the training data, thereby reducing the variance of the output.
	
	\section{Confidence Interval for Real-World Dataset I and II}\label{sec:appd:confidence_intervals_basket_forecast}
	This appendix presents the 90\% confidence intervals for all cases of the real-world dataset I and II. As it is shown in Figures \ref{fig:confidence_interval_basket} and \ref{fig:confidence_interval_forecast}, in most of {\tt Strm} co-players, {\tt SRDQN} outperform {\tt BS}, and with the {\tt BS} co-players, in 12 cases {\tt BS-BS} obtains statistically smaller cost and in 12 cases {\tt SRDQN} and {\tt BS} obtains statistically equal costs. 
	
	\captionsetup[figure]{font=tiny,labelfont=tiny}
	
	\begin{figure}
		\centering
		\caption{Confidence intervals for the real-world dataset I.}
		\label{fig:confidence_interval_basket}
		\begin{subfigure}{0.16\textwidth}
			\includegraphics[scale=0.19]{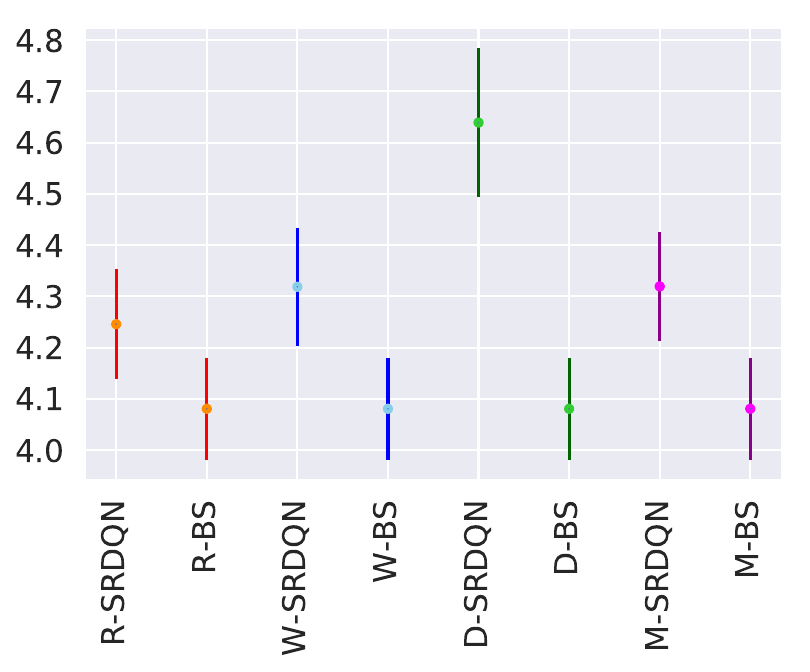}
			\caption{Category 6, {\tt BS}}
			\label{fig:confidence_interval-basket_6_BS}
		\end{subfigure}
		\begin{subfigure}{0.16\textwidth}
			\includegraphics[scale=0.19]{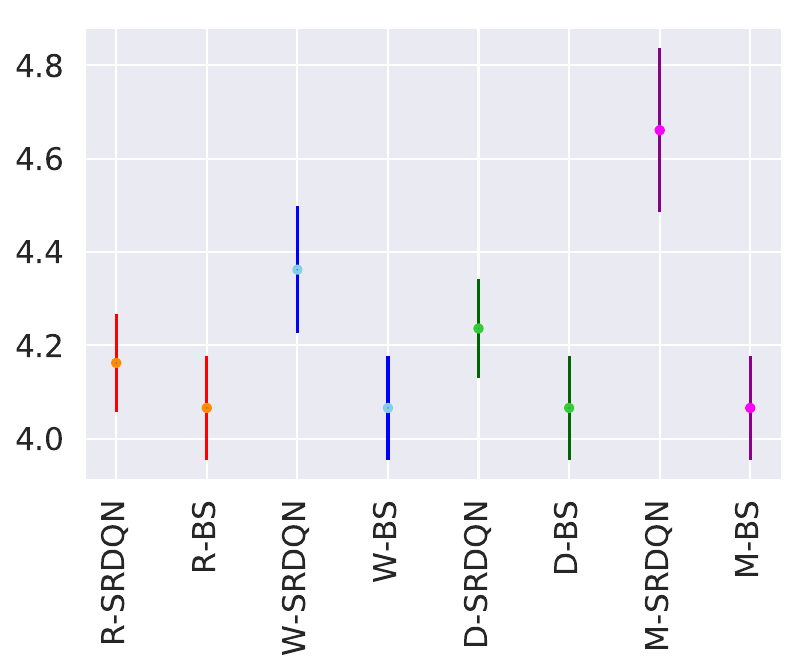}
			\caption{Category 13, {\tt BS}}
			\label{fig:confidence_interval-basket_13_BS}
		\end{subfigure}
		\begin{subfigure}{0.16\textwidth}
			\includegraphics[scale=0.19]{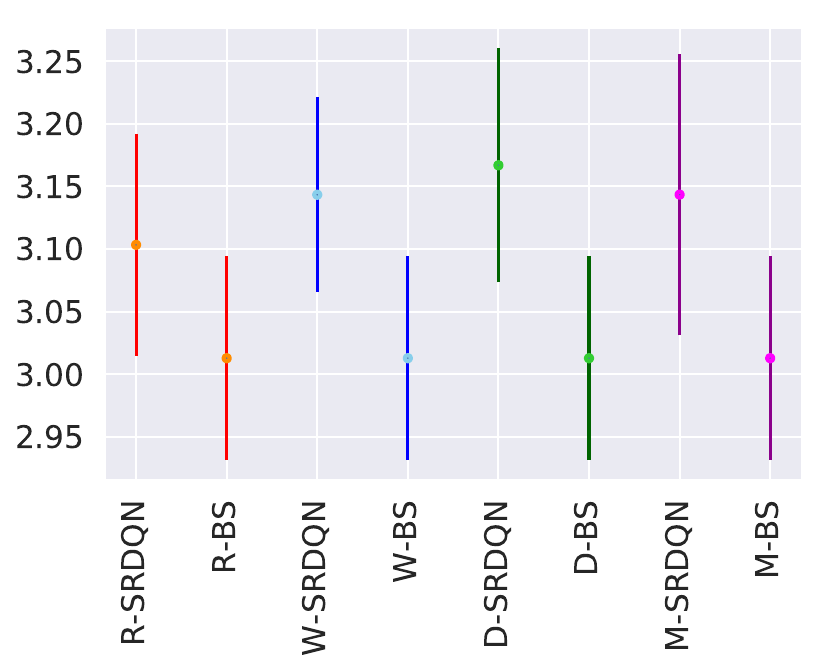}
			\caption{Category 22, {\tt BS}}
			\label{fig:confidence_interval-basket_22_BS}
		\end{subfigure}
		\begin{subfigure}{0.16\textwidth}
			\includegraphics[scale=0.19]{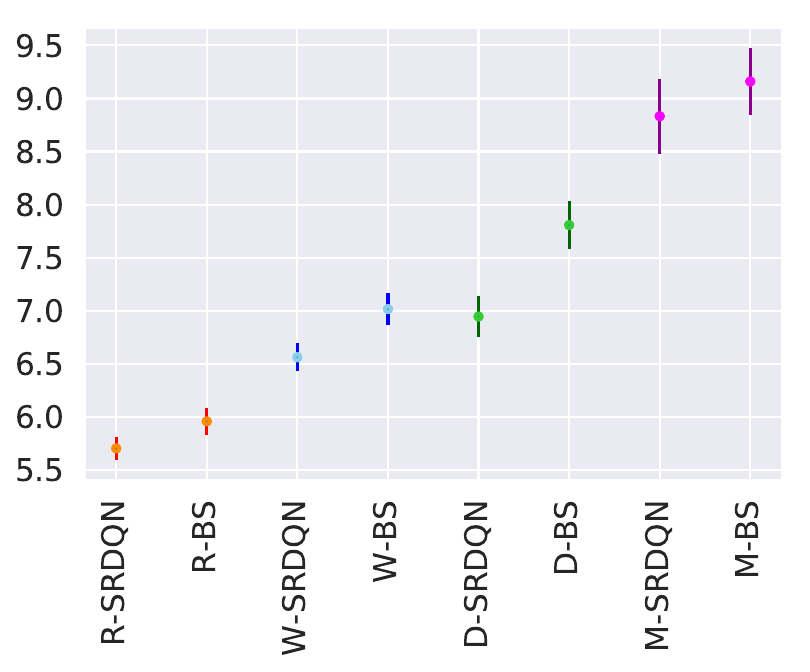}
			\caption{Category 6, {\tt Strm}}
			\label{fig:confidence_interval-basket_6_Strm}
		\end{subfigure}
		\begin{subfigure}{0.16\textwidth}
			\includegraphics[scale=0.19]{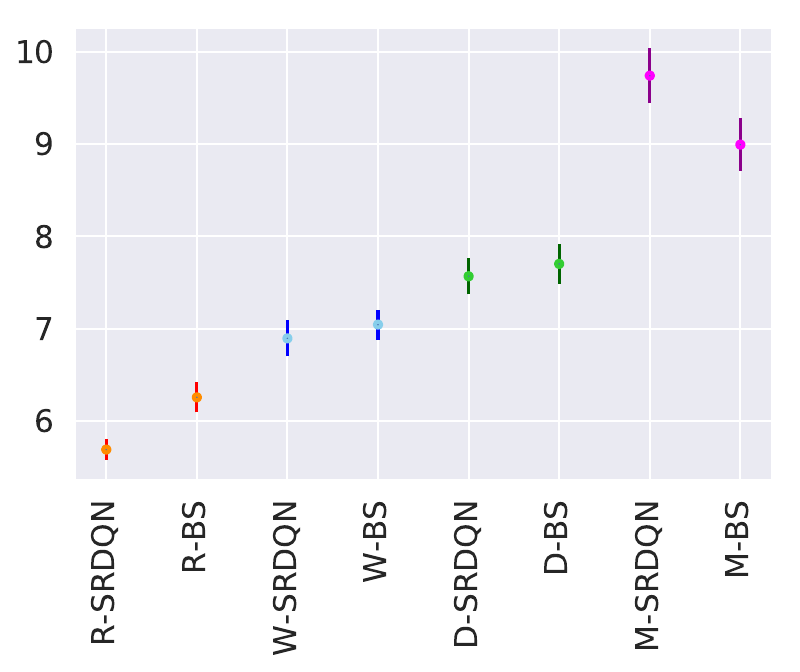}
			\caption{Category 13, {\tt Strm}}
			\label{fig:confidence_interval-basket_13_Strm}
		\end{subfigure}
		\begin{subfigure}{0.16\textwidth}
			\includegraphics[scale=0.19]{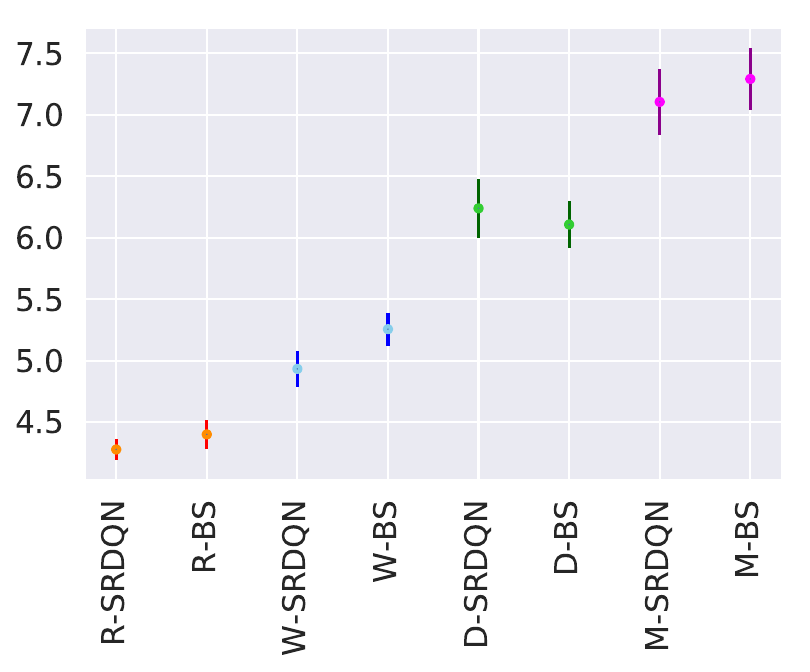}
			\caption{Category 22, {\tt Strm}}
			\label{fig:confidence_interval-basket_22_Strm}
		\end{subfigure}
	\end{figure}

		\begin{figure}
		\centering
		\caption{Confidence intervals for the real-world dataset II.}
		\label{fig:confidence_interval_forecast}
		\begin{subfigure}{0.16\textwidth}
			\includegraphics[scale=0.19]{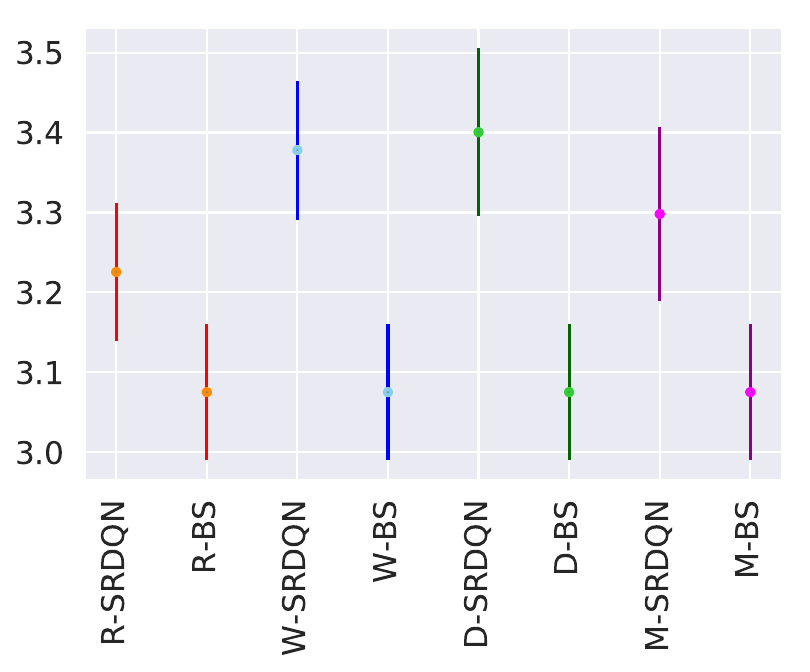}
			\caption{Category 5, {\tt BS}}
			\label{fig:confidence_interval-forecast_5_BS}
		\end{subfigure}
		\begin{subfigure}{0.16\textwidth}
			\includegraphics[scale=0.19]{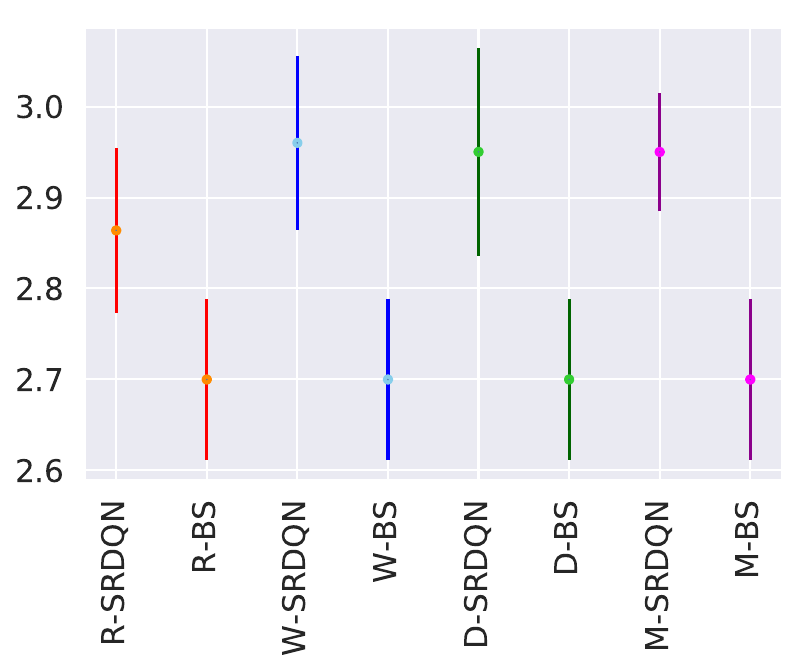}
			\caption{Category 34, {\tt BS}}
			\label{fig:confidence_interval-forecast_34_BS}
		\end{subfigure}
		\begin{subfigure}{0.16\textwidth}
			\includegraphics[scale=0.19]{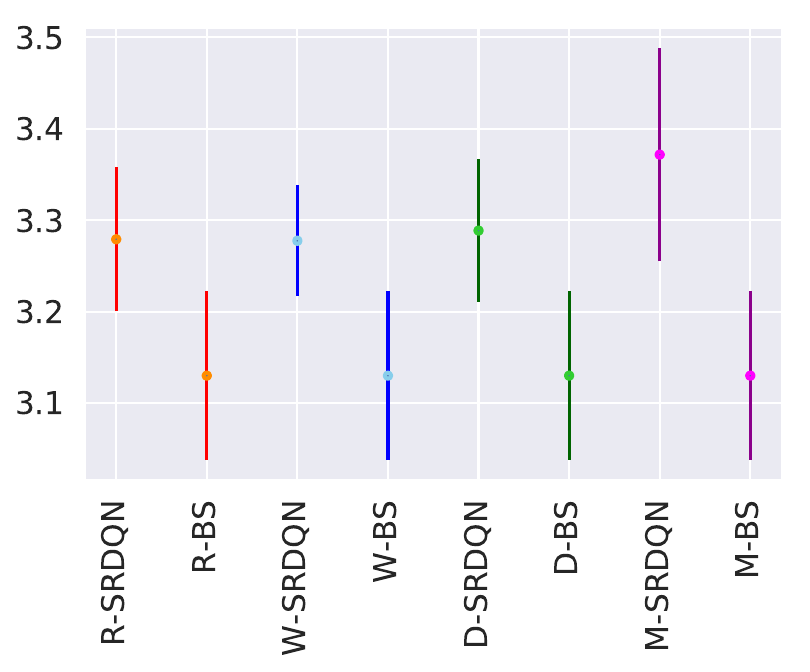}
			\caption{Category 46, {\tt BS}}
			\label{fig:confidence_interval-forecast_46_BS}
		\end{subfigure}
		\begin{subfigure}{0.16\textwidth}
			\includegraphics[scale=0.19]{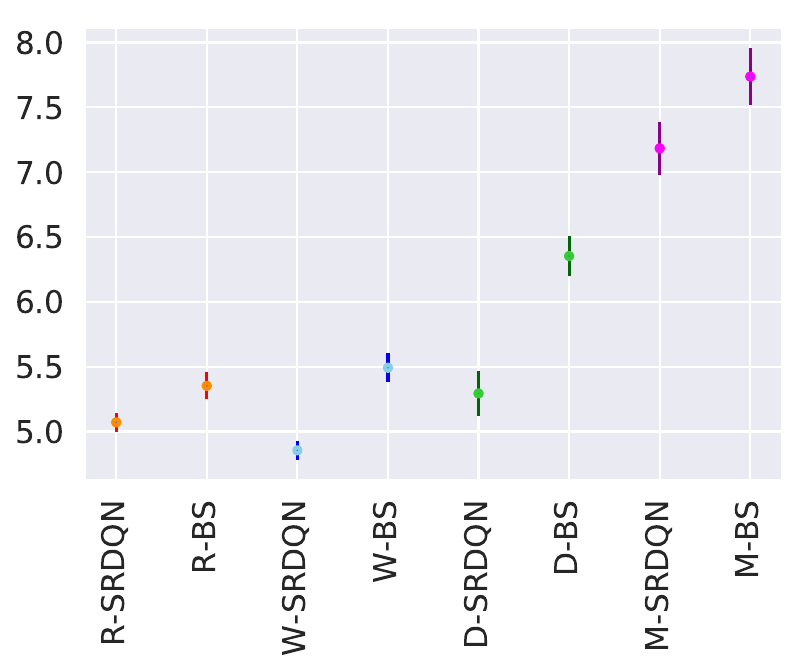}
			\caption{Category 5, {\tt Strm}}
			\label{fig:confidence_interval-forecast_5_Strm}
		\end{subfigure}
		\begin{subfigure}{0.16\textwidth}
			\includegraphics[scale=0.19]{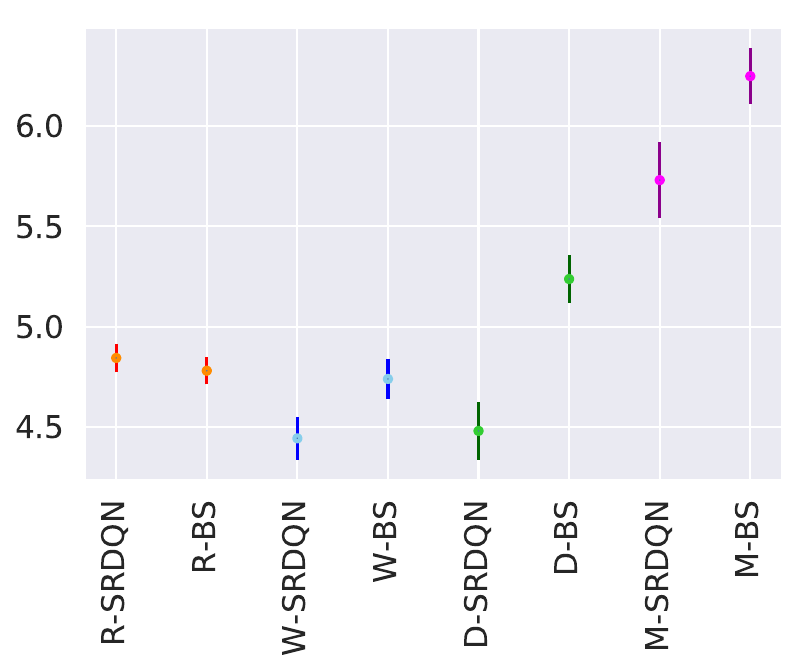}
			\caption{Category 34, {\tt Strm}}
			\label{fig:confidence_interval-forecast_34_Strm}
		\end{subfigure}
		\begin{subfigure}{0.16\textwidth}
			\includegraphics[scale=0.19]{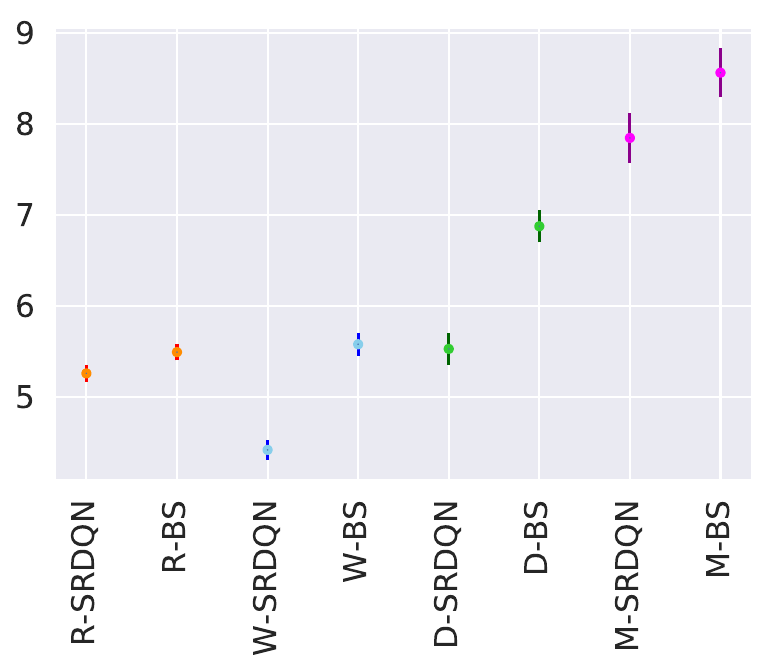}
			\caption{Category 46, {\tt Strm}}
			\label{fig:confidence_interval-forecast_46_Strm}
		\end{subfigure}
	\end{figure}

	\captionsetup[figure]{font=normalsize,labelfont=normalsize}
	
	\section{Extended Numerical Results} \label{sec:appd:more_play_results}
	
	This appendix shows additional results on the details of play of each agent. Figure \ref{fig:4Agent:vs_optm:b3:play_DNN_W_D_M} provides the details of $IL$, $OO$, $a$, $r$, and OUTL for each agent when the SRDQN retailer plays with co-players who use the base-stock levels obtained by the Clark--Scarf algorithm. Clearly, {\tt BS-SRDQN} attains a similar IL, OO, action, and reward to those of {\tt BS-BS}. 
	Figure \ref{fig:4Agent:vs_optm:b4:play_DNN_R_W_D_M} provides analogous results for the case in which the SRDQN manufacturer plays with three {\tt Strm} agents. The SRDQN agent learns that the shortage costs of the non-retailer agents are zero and exploits that fact to reduce the total cost.
	In each of the figures, the top set of charts provides the results of the retailer, followed by the warehouse, distributor, and manufacturer.

	\begin{figure}
		\centering
		\caption{$IL_t$, $OO_t$, $a_t$, and $r_t$ of all agents when SRDQN retailer plays with three base-stock co-players ({\tt BS-SRDQN}). \label{fig:4Agent:vs_optm:b3:play_DNN_W_D_M} }		
		\vspace{-8pt}
		{\includegraphics[scale=0.47]{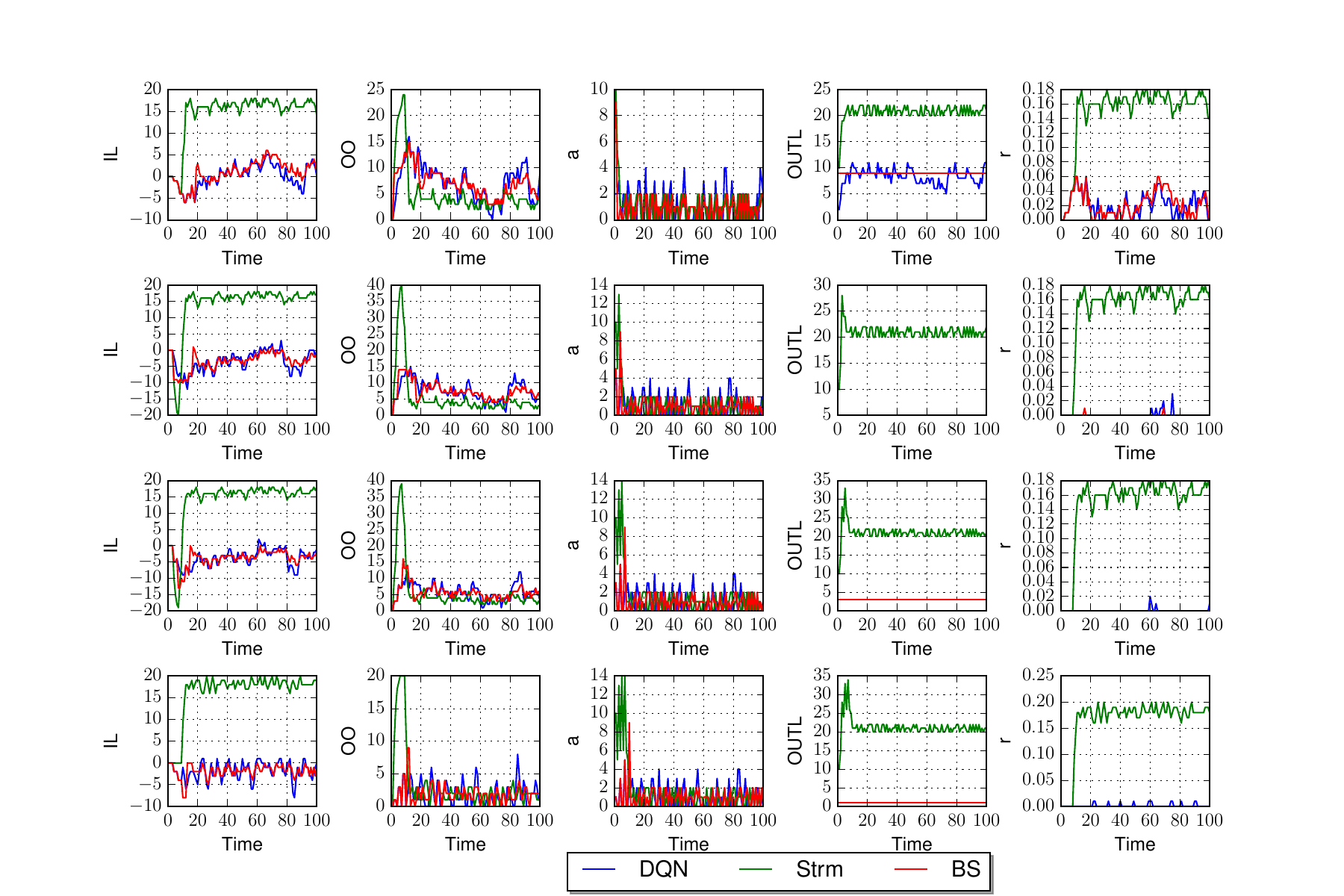}}
		{}
	\end{figure}

	\begin{figure}
		\centering
		\caption{$IL_t$, $OO_t$, $a_t$, and $r_t$ of all agents when SRDQN manufacturer plays with three Sterman co-players ({\tt Strm-SRDQN}).  \label{fig:4Agent:vs_optm:b4:play_DNN_R_W_D_M} }		
		\vspace{-8pt}			
		{\includegraphics[scale=0.47]{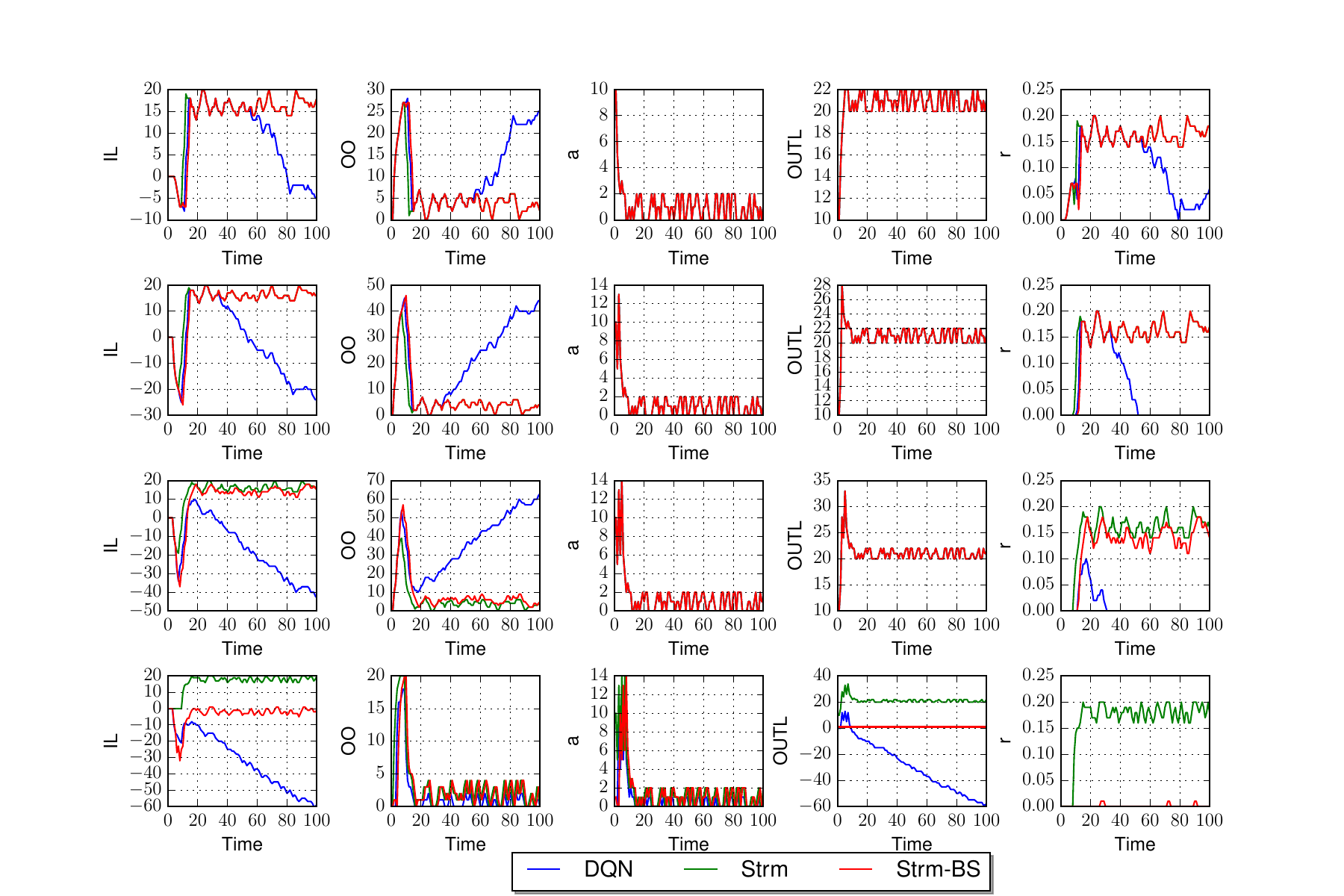}}
		{}
	\end{figure} 


	\section{The Effect of $\beta$ on the Performance of Each Agent}\label{sec:appdx:base_stock_beta}
	
	
	Figure \ref{fig:beta:dqn_vs_three_optimal} plots the training trajectories for SRDQN agents playing with three agents that use the base-stock levels obtained by the Clark--Scarf algorithm, using various values of $C$, $m$, and $\beta$. In each sub-figure, the blue line denotes the result when all players use a base-stock policy while the remaining curves each represent the agent using SRDQN with different values of $C$, $\beta$, and $m$, trained for 60000 episodes with a learning rate of $0.00025$.
	
	As shown in Figure \ref{fig:4Agent:vs_optm:beta:A_DNN_Retailer}, when the SRDQN plays the retailer, $\beta_1 \in \{20,40\}$ works well, and $\beta_1 = 40$ provides the best results. 
	As we move upstream in the supply chain (warehouse, then distributor, then manufacturer), smaller $\beta$ values become more effective (see Figures \ref{fig:4Agent:vs_optm:beta:A_DNN_Warehouse}--\ref{fig:4Agent:vs_optm:beta:A_DNN_Manufacturer}). Recall that the retailer bears the largest share of the optimal expected cost per period, and as a result it needs a larger $\beta$ than the other agents. Not surprisingly, larger $m$ values attain better costs since the SRDQN has more knowledge of the environment. Finally, larger $C$ works better and provides a stable SRDQN model. However, there are some combinations for which smaller $C$ and $m$ also work well, e.g., see Figure \ref{fig:4Agent:vs_optm:beta:A_DNN_Manufacturer}, trajectory $5000$-$20$-$5$.
	
	\begin{figure}
		\centering
		\caption{Total cost (upper figure) and normalized cost (lower figure) for {\tt BS-SRDQN.} 
		\label{fig:beta:dqn_vs_three_optimal}}		
		\vspace{-5pt}		
		\begin{subfigure}{0.28\textwidth}
			\centerline{ \includegraphics[width=1.4\textwidth]{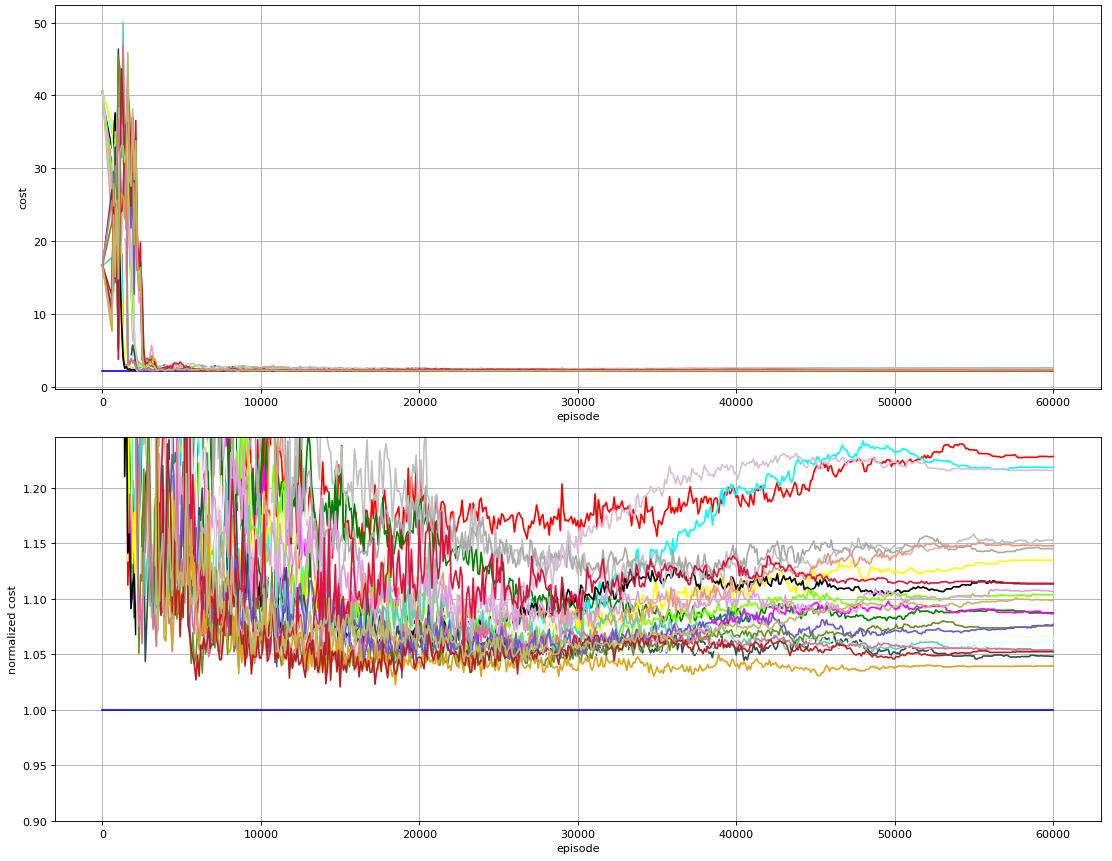}} 
			\vspace{-10pt}
			\caption{{\small SRDQN plays retailer} }
		 \label{fig:4Agent:vs_optm:beta:A_DNN_Retailer}  		
		\end{subfigure}
		\hspace{90pt}
		\begin{subfigure}{0.28\textwidth}
			\centerline{ \includegraphics[width=1.4\textwidth]{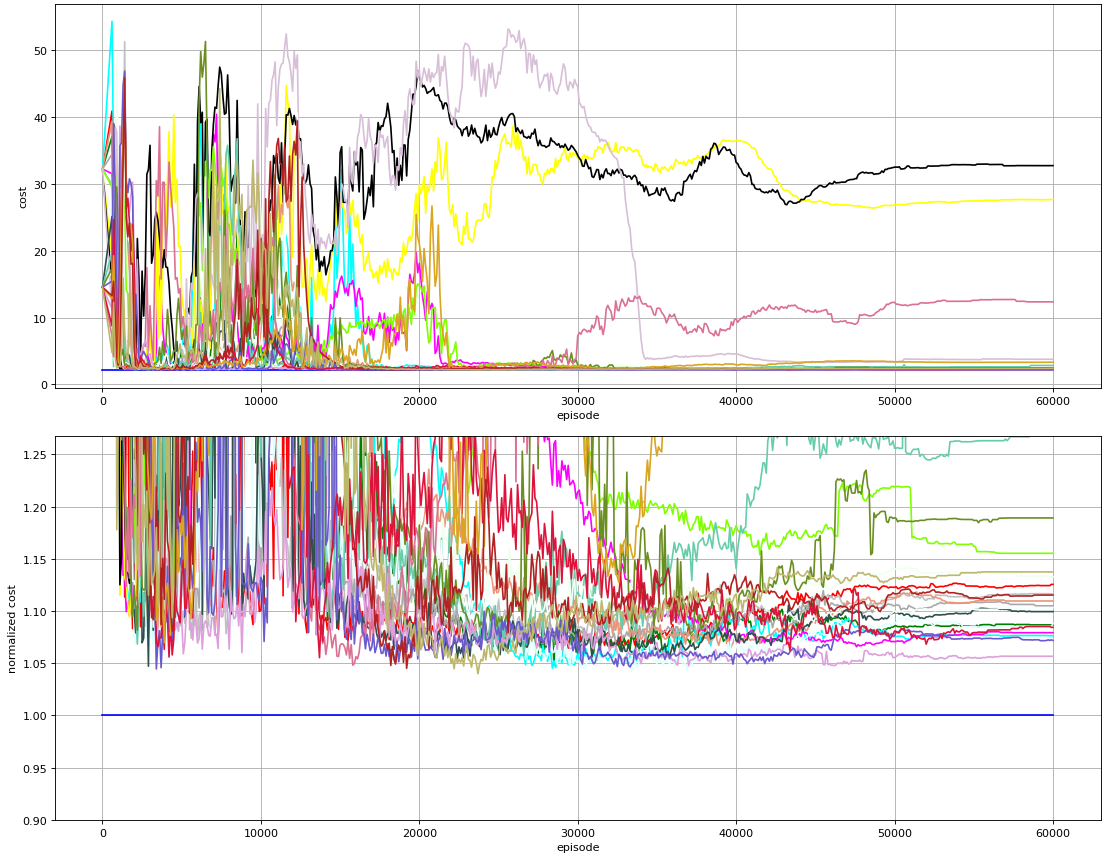}}
			\vspace{-10pt}			
			\caption{{\small SRDQN plays warehouse}}
		 \label{fig:4Agent:vs_optm:beta:A_DNN_Warehouse}  	
		\end{subfigure}		
		
		\begin{subfigure}{0.28\textwidth}
			\centerline{ \includegraphics[width=1.4\textwidth]{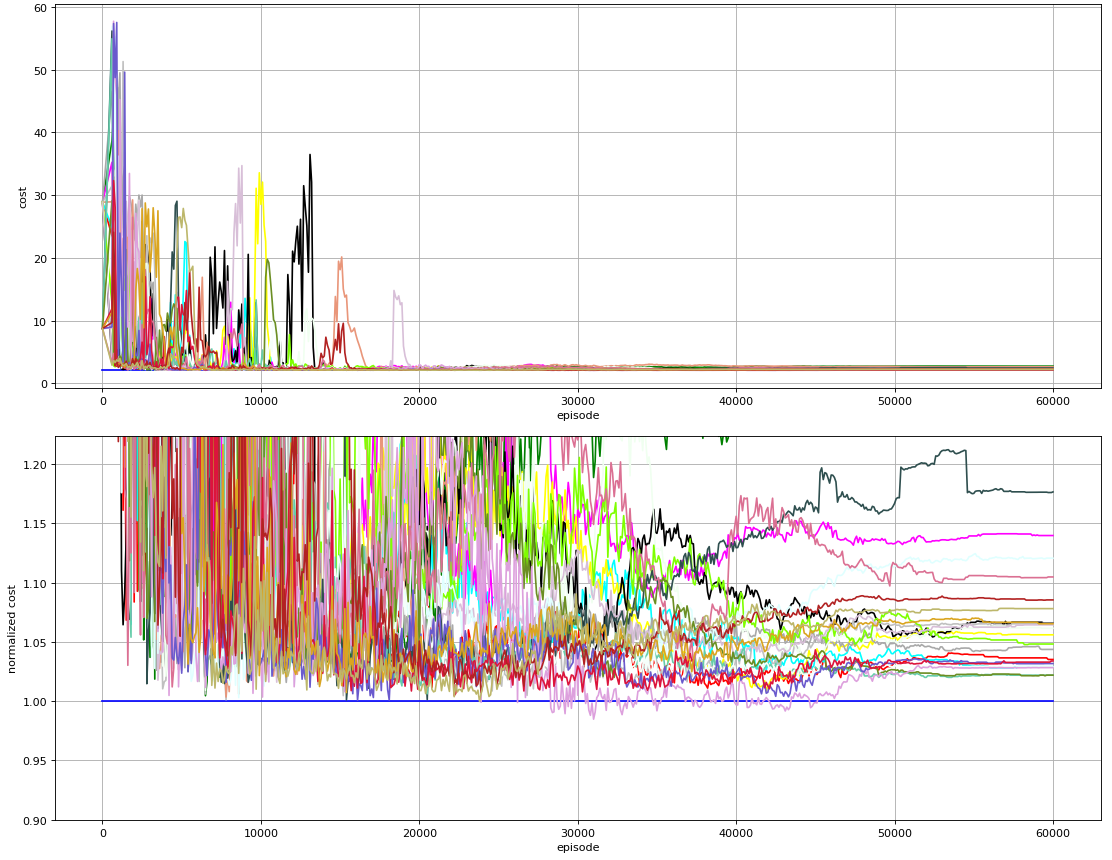}}
			\vspace{-10pt}			
			\caption{{\small SRDQN plays distributor} }			
		 \label{fig:4Agent:vs_optm:beta:A_DNN_Distributer}
		\end{subfigure}	
		\hspace{90pt}
		\begin{subfigure}{0.28\textwidth}
			\centerline{ \includegraphics[width=1.4\textwidth]{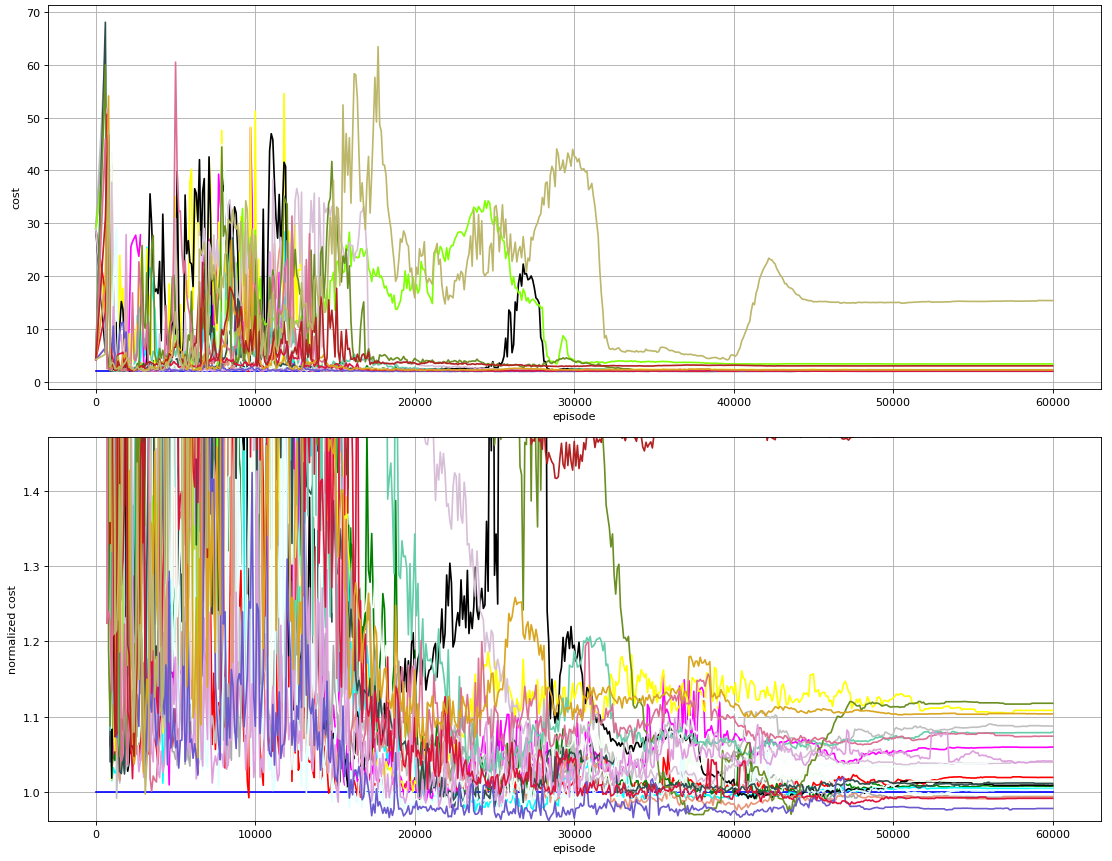}}
			\vspace{-10pt}			
			\caption{{\small SRDQN plays manufacturer}}
		 \label{fig:4Agent:vs_optm:beta:A_DNN_Manufacturer}
		\end{subfigure}	
		\vspace{-5pt}
		
		\begin{subfigure}{0.58\textwidth}
			\centerline{ \includegraphics[width=1.8\textwidth]{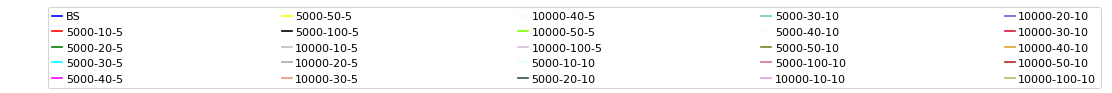}}
		\end{subfigure}						
		{}
	\end{figure}

\section{Extended Results on Transfer Learning}\label{sec:appdx:extended_tl}

\subsection{Transfer Knowledge Between Agents}\label{sec:results:transfer_learning:same_agents}

In this section, we present the results of the transfer learning method when the trained agent $i \in \{1,2,3,4\}$ transfers its first $k \in \{1,2,3\}$ layer(s) into co-player agent $j \in \{1,2,3,4\}$, $j \neq i$.
For each target-agent $j$, Figure \ref{fig:agents_all_same} shows the results for the best source-agent $i$ and the number of shared layers $k$, out of the 9 possible choices for $i$ and $k$. In the sub-figure captions, the notation $j$-$i$-$k$ indicates that source-agent $i$ shares weights of the first $k$ layers with target-agent $j$, so that those $k$ layers remain non-trainable. 

Except for agent 2, all agents obtain costs that are very close to those of the base-stock policy, with a $6.06\%$ gap, on average. (In Section \ref{sec:results:dnn_vs_base_stock}, the average gap is $2.31\%$.)
However, none of the agents was a good source for agent 2. It seems that the acquired knowledge of other agents is not enough to get a good solution for this agent, or the feature space that agent 2 explores is different from other agents, so that it cannot get a solution whose cost is close to the {\tt BS-BS} cost. 

\begin{figure}	
	\centering
	\caption{Results of transfer learning between agents with the same cost coefficients and action space.} \label{fig:agents_all_same}	
	\vspace{-5pt}
	\begin{subfigure}{0.24\textwidth}
		\centering
		\includegraphics[scale=0.16]{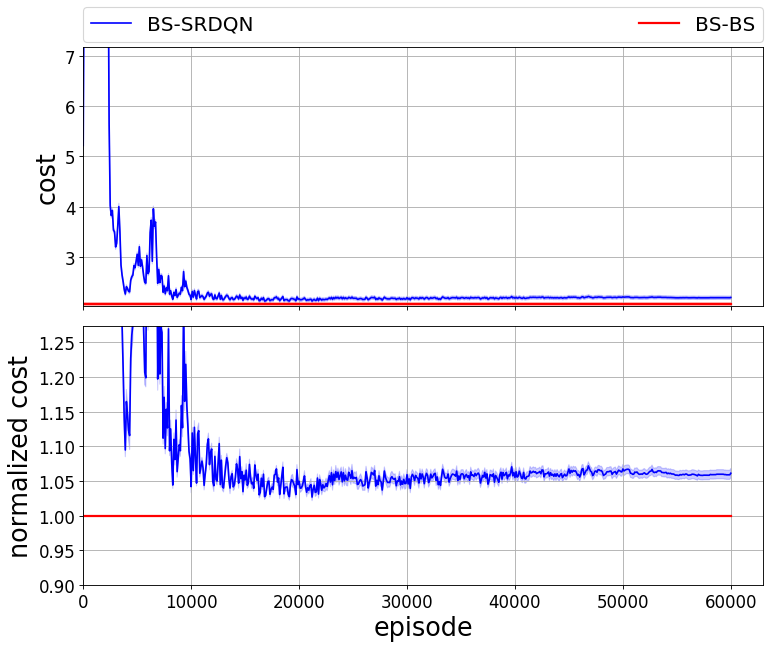}
		\caption{Case 1-4-1 \label{fig:agent_3-4-3}		}
		{}
	\end{subfigure}
	\begin{subfigure}{0.24\textwidth}
		\centering
		\includegraphics[scale=0.16]{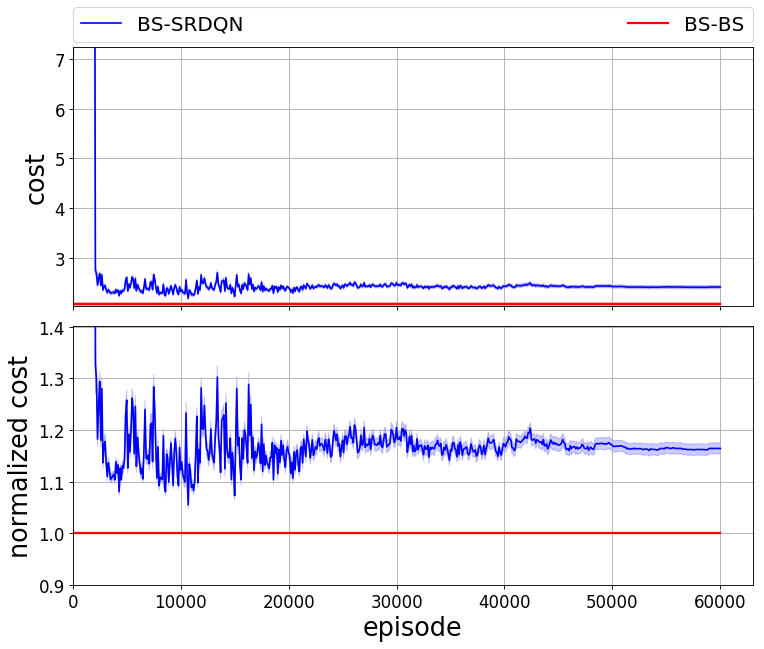}
		\caption{Case 2-4-1 \label{fig:agent_4-5-1}		}
		{}
	\end{subfigure}	
	\begin{subfigure}{0.24\textwidth}
		\centering
		\includegraphics[scale=0.16]{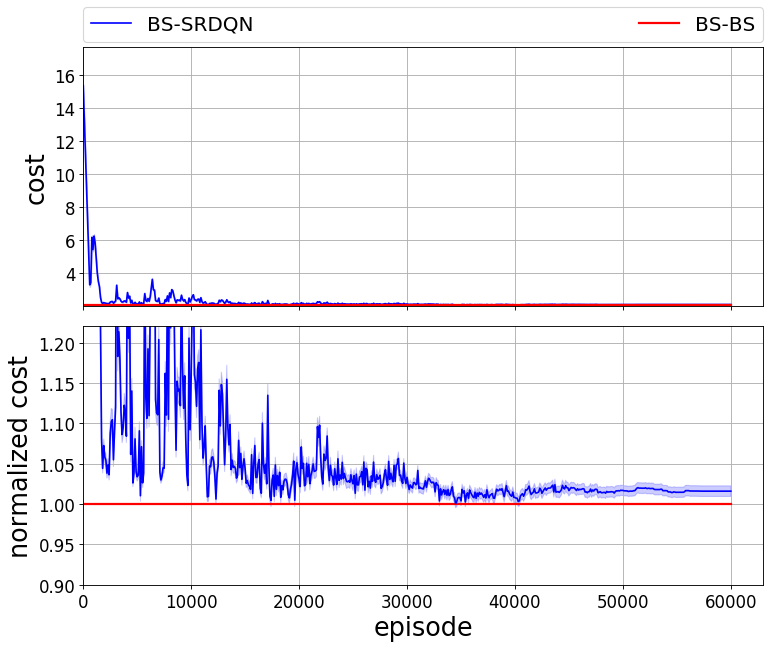}
		\caption{Case 3-1-1 \label{fig:agent_5-3-1}		}
		{}
	\end{subfigure}		
	\begin{subfigure}{0.24\textwidth}
		\centering
		\includegraphics[scale=0.16]{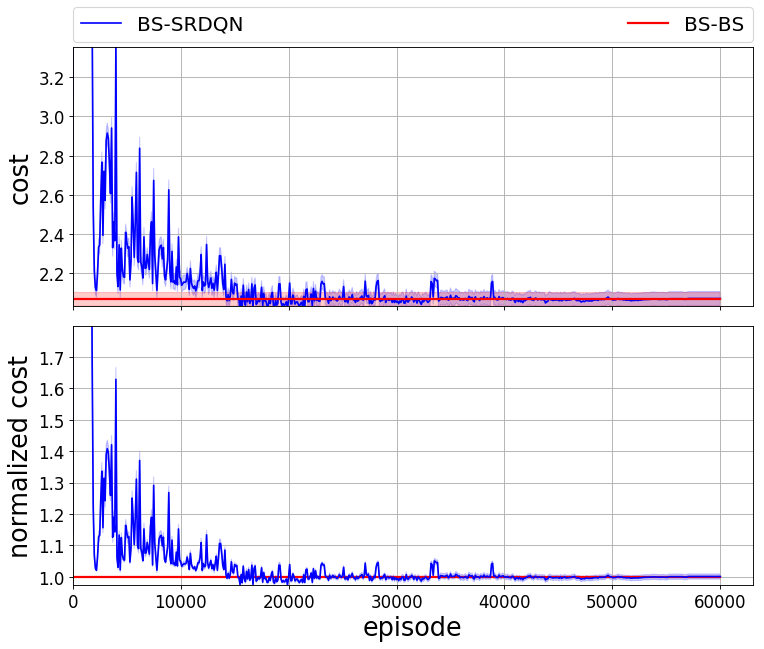}
		\caption{Case 4-2-1 \label{fig:agent_6-5-2}		}
		{}
	\end{subfigure}			
	{}
\end{figure}

In order to get more insight, consider Figure \ref{fig:dqn_vs_three_optimal}, which presents the best results obtained through hyper-parameter tuning for each agent. 
In that figure, all agents start the training with a large cost value, and after 25000 fluctuating iterations, each converges to a stable solution. 
In contrast, in Figure \ref{fig:agents_all_same}, each agent starts from a relatively small cost value, and after a few thousand training episodes converges to the final solution. Moreover, for agent 3, the final cost of the transfer learning solution is smaller than that obtained by training the network from scratch. And, the transfer learning method used one order of magnitude less CPU time than the approach in Section \ref{sec:results:dnn_vs_base_stock} to obtain very close results.

We also observe that agent $j$ can obtain good results when $k=1$ and $i$ is either $j-1$ or $j+1$. 
This shows that the learned weights of the first SRDQN network layer are general enough to transfer knowledge to the other agents, and also that the learned knowledge of neighboring agents is similar. 
Also, for any agent $j$, agent $i=1$ provides similar results to that of agent $i=j-1$ or $i=j+1$ does, and in some cases it provides slightly smaller costs, which shows that agent $1$ captures general feature values better than the others. 

	\subsection{Transfer Knowledge for Different Cost Coefficients}\label{sec:results:transfer_learning:different_cp_ch}

Figure \ref{fig:agents_cp5} shows the best results achieved for all agents, when agent $j$ has different cost coefficients, $(c_{p_2}, c_{h_2}) \neq (c_{p_1}, c_{h_1})$. We test target agents $j \in \{1,2,3,4\}$, such that the holding and shortage costs are (5,1), (5,0), (5,0), and (5,0) for agents 1 to 4, respectively. 
In all of these tests, the source and target agents have the same action spaces. All agents attain cost values close to the {\tt BS-BS} cost; in fact, the overall average cost is 6.16\% higher than the {\tt BS-BS} cost.

\begin{figure}	
	\centering
	\caption{Fesults of transfer learning between agents with different cost coefficients and same action space. \label{fig:agents_cp5}}	
	\vspace{-5pt}			
	\begin{subfigure}{0.24\textwidth}
		\centering
		\includegraphics[scale=0.16]{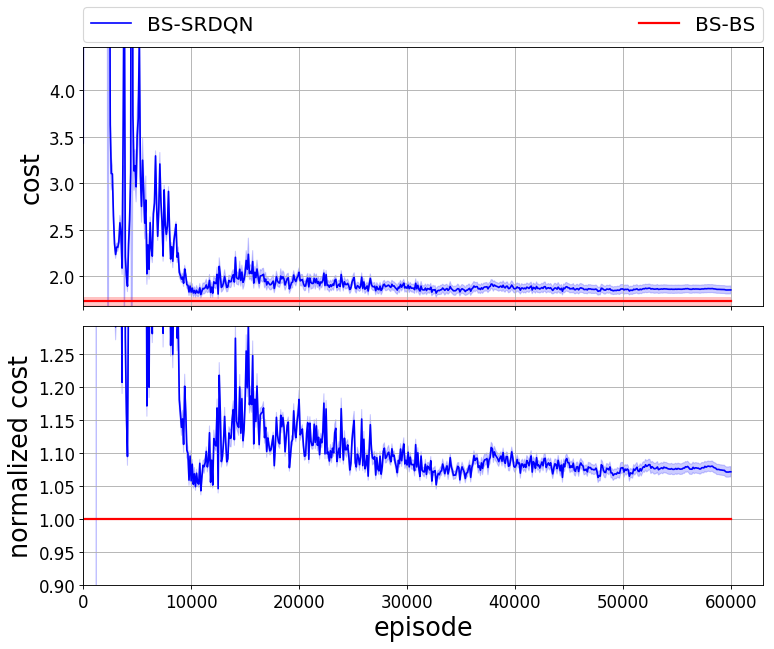}
		\caption{Case 1-4-1}
		\label{fig:agent_cp5_3-4-3}		
	\end{subfigure}
	\begin{subfigure}{0.24\textwidth}
		\centering
		\includegraphics[scale=0.16]{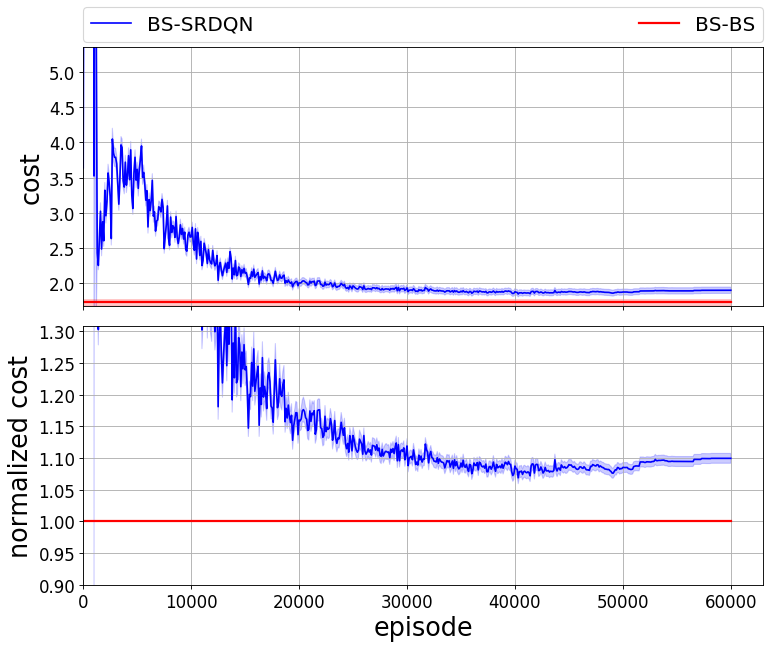}
		\caption{Case 2-3-3}
		\label{fig:agent_cp5_4-5-1}		
	\end{subfigure}	
	\begin{subfigure}{0.24\textwidth}
		\centering
		\includegraphics[scale=0.16]{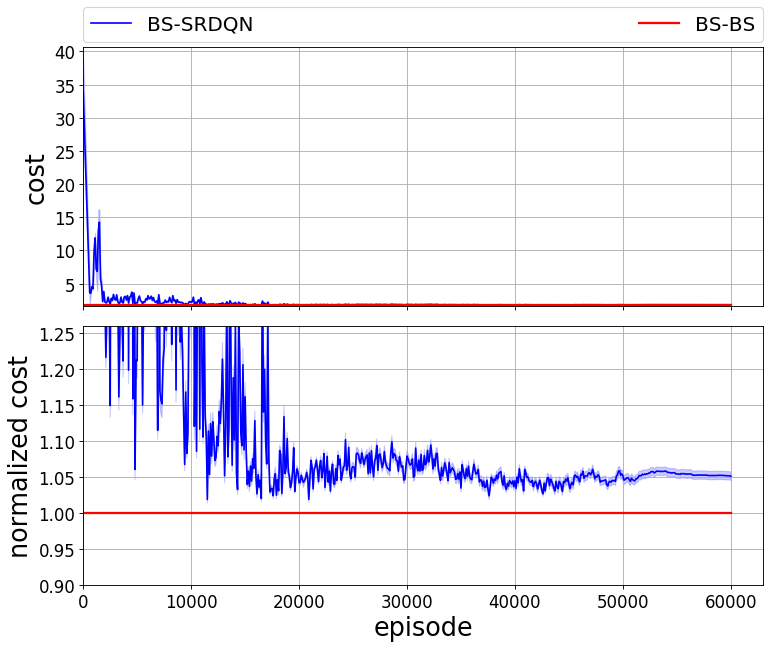}
		\caption{Case 3-1-1}
		\label{fig:agent_cp5_5-3-1}		
	\end{subfigure}		
	\begin{subfigure}{0.24\textwidth}
		\centering
		\includegraphics[scale=0.16]{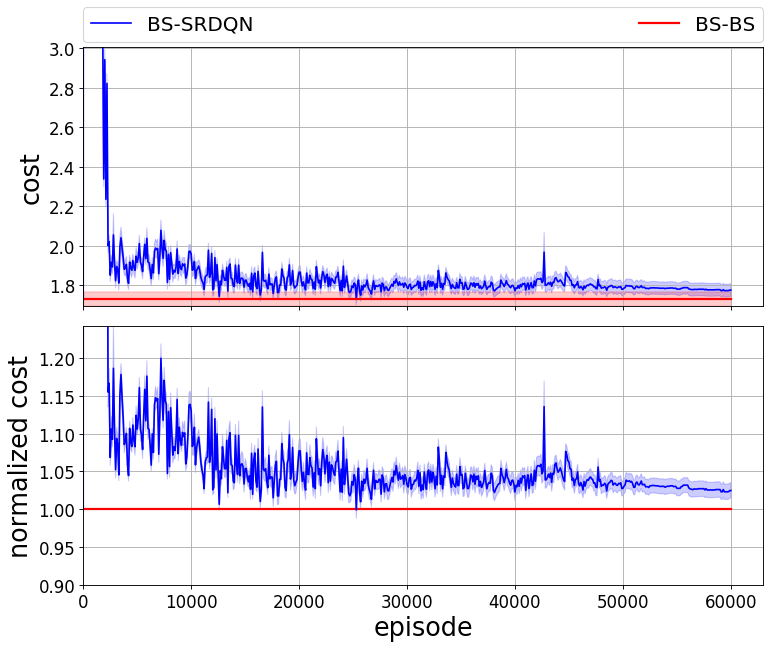}
		\caption{Case 4-4-2}
		\label{fig:agent_cp5_6-5-2}		
	\end{subfigure}		
	{}
\end{figure}	

In addition, similar to the results of Section \ref{sec:results:transfer_learning:same_agents}, base agent $i=1$ provides good results for all target agents. We also performed the same tests with shortage and holding costs (10,1), (1,0), (1,0), and (1,0) for agents 1 to 4, respectively, and obtained very similar results.

\subsection{Transfer Knowledge for Different Size of Action Space}\label{sec:results:transfer_learning:different_action}

Increasing the size of the action space should increase the accuracy of the $d+x$ approach. However, it makes the training process harder. It can be effective to train an agent with a small action space and then transfer the knowledge to an agent with a larger action space. 
To test this, we test target-agent $j \in \{1,2,3,4\}$ with action space $\{-5,\dots,5\}$, assuming that the source and target agents have the same cost coefficients. 

Figure \ref{fig:agents_action5} shows the best results achieved for all agents. All agents attained costs that are close to the {\tt BS-BS} cost, with an average gap of approximately 10.66\%.


\begin{figure}	
	\centering
	\caption{Results of transfer learning for agents with $|\mathcal{A}_1| \neq |\mathcal{A}_2|, (c^j_{p_1}, c^j_{h_1}) = (c^j_{p_2}, c^j_{h_2})$. \label{fig:agents_action5}}	
	\vspace{-5pt}
	\begin{subfigure}{0.24\textwidth}
		\centering
		\includegraphics[scale=0.16]{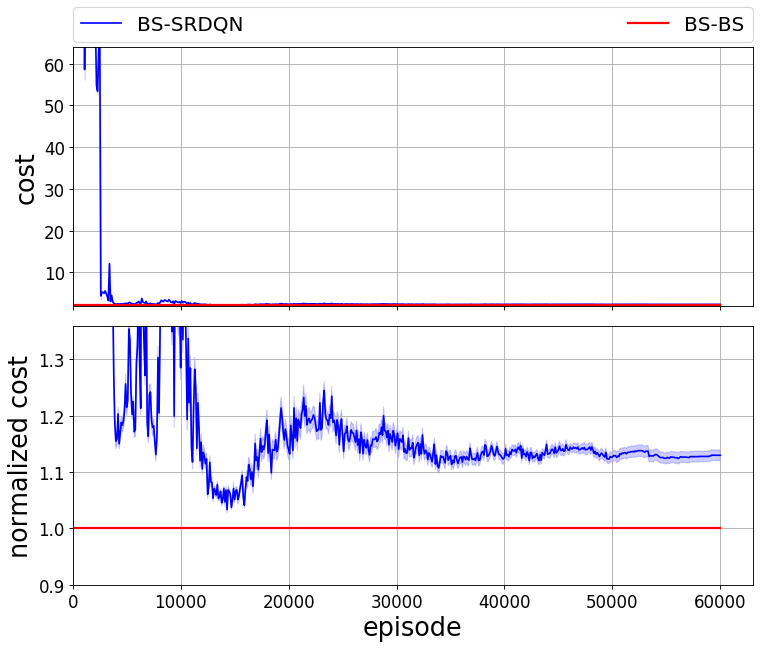}
		\caption{Case 1-3-1}
		\label{fig:agent_action5_3-3-2}		
	\end{subfigure}
	\begin{subfigure}{0.24\textwidth}
		\centering
		\includegraphics[scale=0.16]{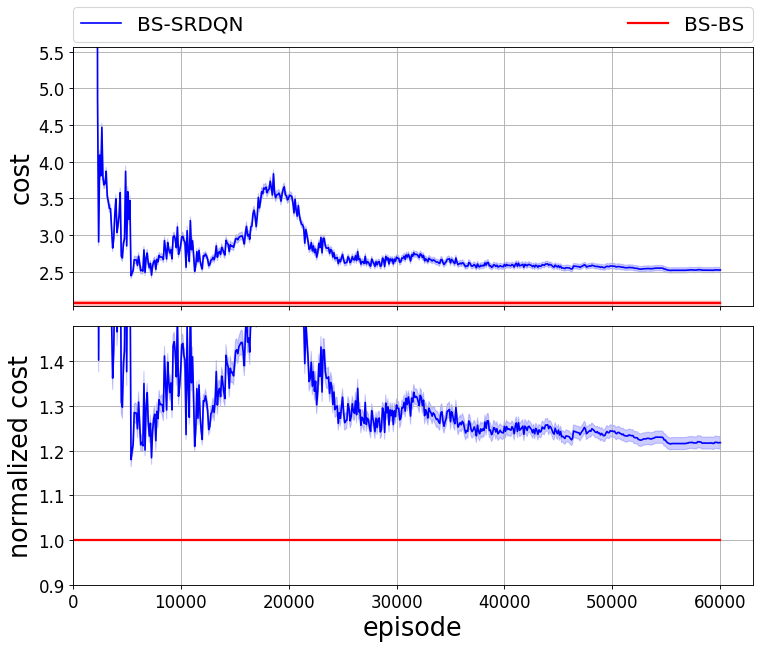}
		\caption{Case 2-3-2}
		\label{fig:agent_action5_4-4-1}	
	\end{subfigure}	
	\begin{subfigure}{0.24\textwidth}
		\centering
		\includegraphics[scale=0.16]{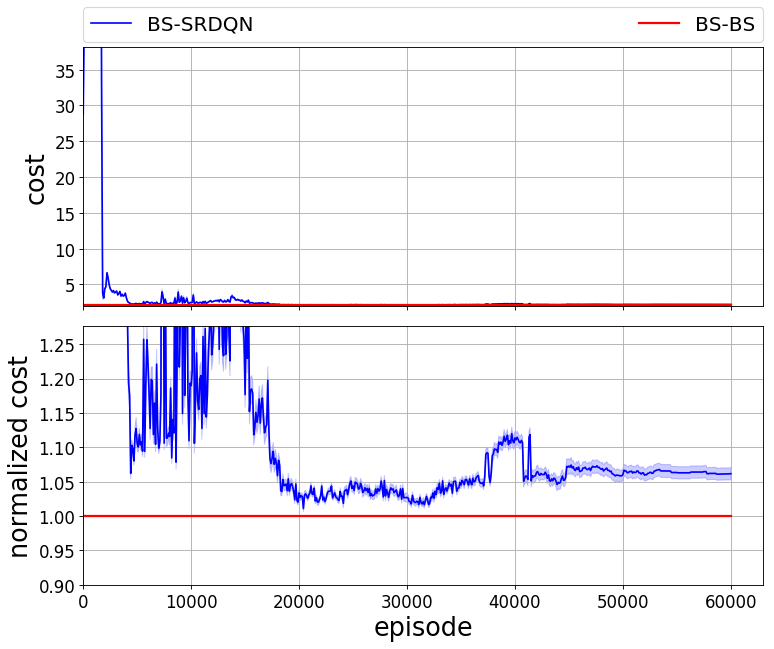}
		\caption{Case 3-4-2}
		\label{fig:agent_action5_5-5-1}		
	\end{subfigure}		
	\begin{subfigure}{0.24\textwidth}
		\centering
		\includegraphics[scale=0.16]{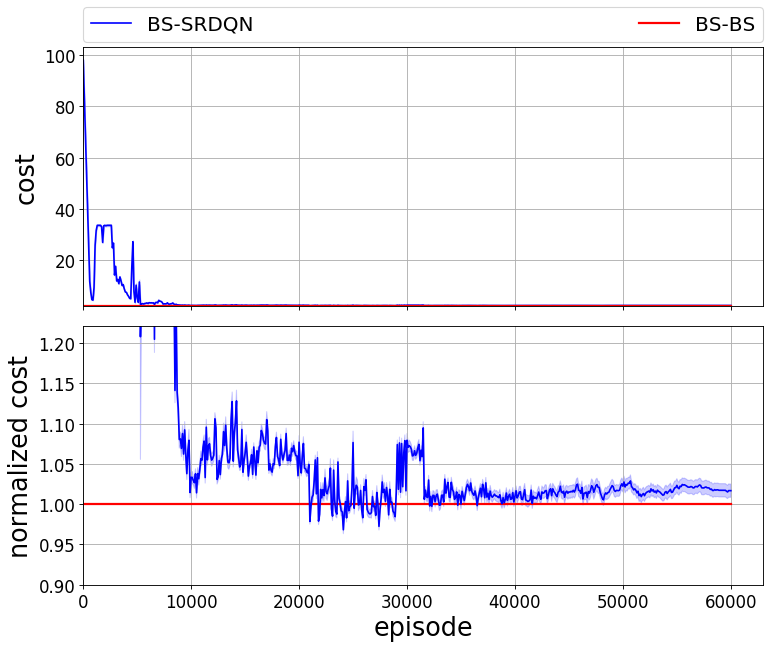}
		\caption{Case 4-2-1}
		\label{fig:agent_action5_6-6-1}		
	\end{subfigure}		
	{}
\end{figure}

\subsection{Transfer Knowledge for Different Action Space, Cost Coefficients, and Demand Distribution}\label{sec:results:transfer_learning:case_5}

This case includes all difficulties of the cases in Sections \ref{sec:results:transfer_learning:same_agents}, \ref{sec:results:transfer_learning:different_cp_ch}, \ref{sec:results:transfer_learning:different_action}, and \ref{sec:results:transfer_learning}, in addition to the demand distributions being different. So, the range of demand, $IL$, $OO$, $AS$, and $AO$ that each agent observes is different than those of the base agent. Therefore, this is a hard case to train, and the average optimality gap is 17.41\%; however, as Figure \ref{fig:case_5} depicts, the cost values decrease quickly and the training noise is quite small. 

\begin{figure}	
	\centering
	\caption{Results of transfer learning for agents with $|\mathcal{A}_1| \neq |\mathcal{A}_2|, (c^j_{p_1}, c^j_{h_1}) \neq (c^j_{p_2}, c^j_{h_2}), D_1 \neq D_2$, and $\pi_1 \neq \pi_2$. \label{fig:case_5}}	
	\vspace{-5pt}			
	\begin{subfigure}{0.24\textwidth}
		\centering
		\includegraphics[scale=0.16]{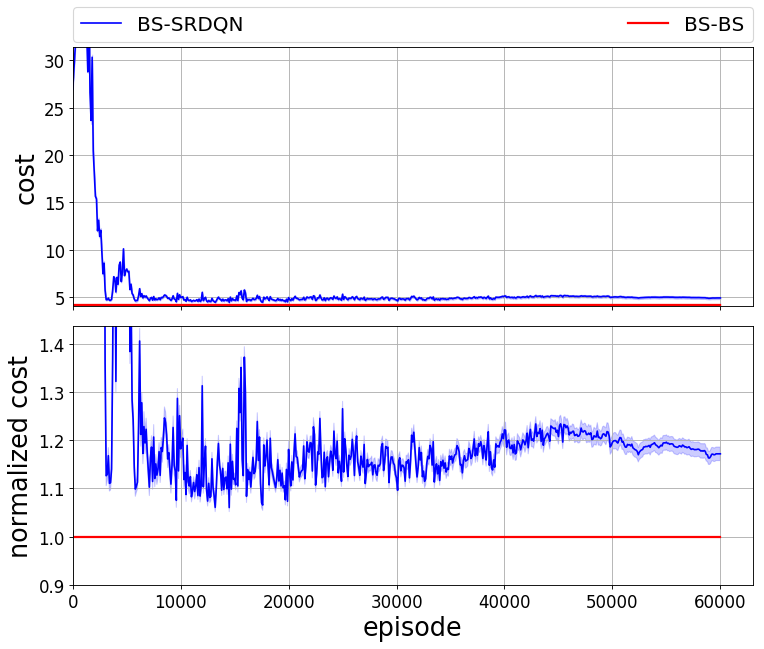}
		\caption{Case 1-3-1}
		\label{fig:agent_case5_3-6-1}		
	\end{subfigure}
	\begin{subfigure}{0.24\textwidth}
		\centering
		\includegraphics[scale=0.16]{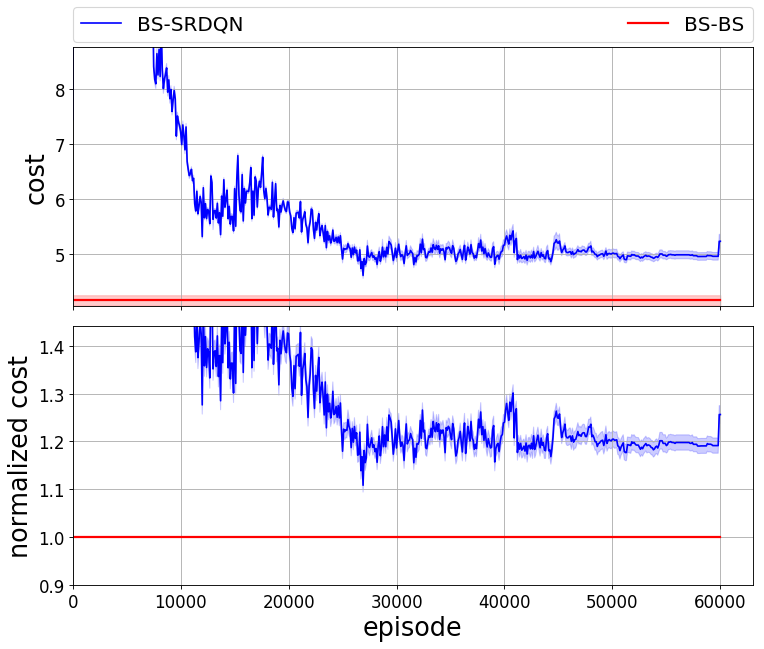}
		\caption{Case 2-3-3}
		\label{fig:agent_case5_4-5-3}		
	\end{subfigure}	
	\begin{subfigure}{0.24\textwidth}
		\centering
		\includegraphics[scale=0.16]{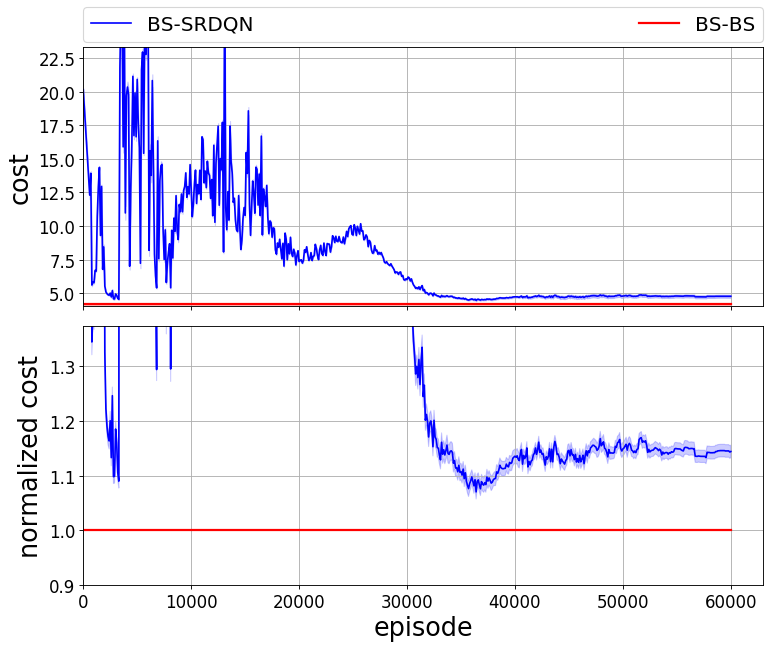}
		\caption{Case 3-2-1}
		\label{fig:agent_case5_5-4-1}		
	\end{subfigure}		
	\begin{subfigure}{0.24\textwidth}
		\centering
		\includegraphics[scale=0.16]{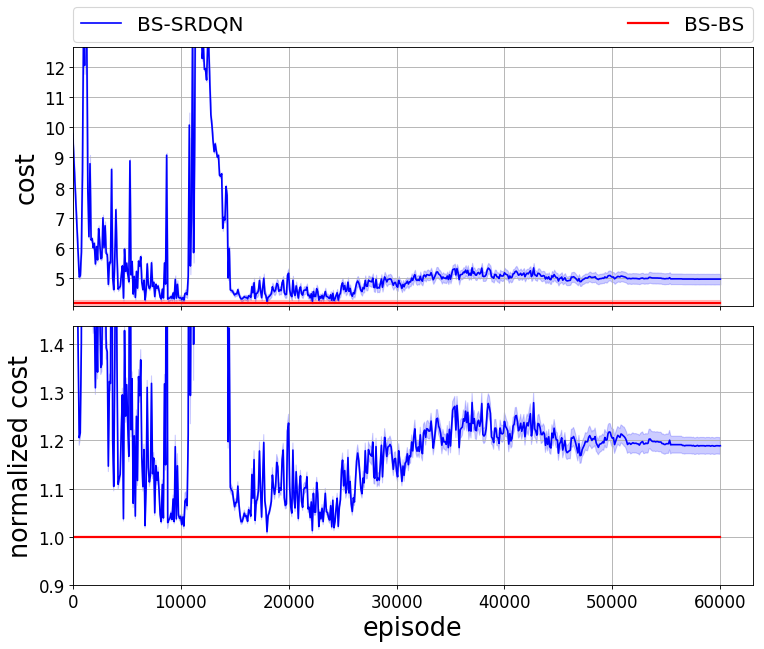}
		\caption{Case 4-3-2}
		\label{fig:agent_case5_6-5-2}		
	\end{subfigure}		
	{}
\end{figure}

\subsection{Transfer Knowledge for Different Action Space, Cost Coefficients, Demand Distribution, and $\pi_2$}\label{sec:results:transfer_learning:case_6}
Figures \ref{fig:case_6_sterman} and \ref{fig:case_6_random} show the results of the most complex transfer learning cases that we tested. Although the SRDQN plays with non-rational (Sterman ) co-players and the observations in each state might be quite noisy, there are relatively small fluctuations in the training, and for all agents after around 40,000 iterations they converge.

\begin{figure}	
	\centering
	\caption{Results of transfer learning for agents with $|\mathcal{A}_1| \neq |\mathcal{A}_2|, (c^j_{p_1}, c^j_{h_1}) \neq (c^j_{p_2}, c^j_{h_2}), D_1 \neq D_2$, and $\pi_1 \neq \pi_2$. \label{fig:case_6_sterman}}	
	\vspace{-5pt}			
	\begin{subfigure}{0.24\textwidth}
		\centering
		\includegraphics[scale=0.16]{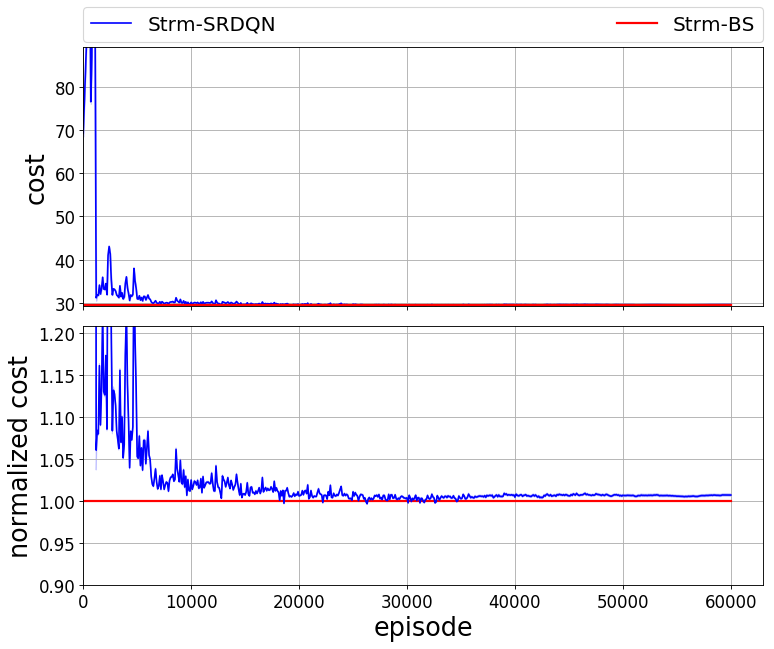}
		\caption{Case 1-1-1}
		\label{fig:agent_case6_7-3-1}		
	\end{subfigure}
	\begin{subfigure}{0.24\textwidth}
		\centering
		\includegraphics[scale=0.16]{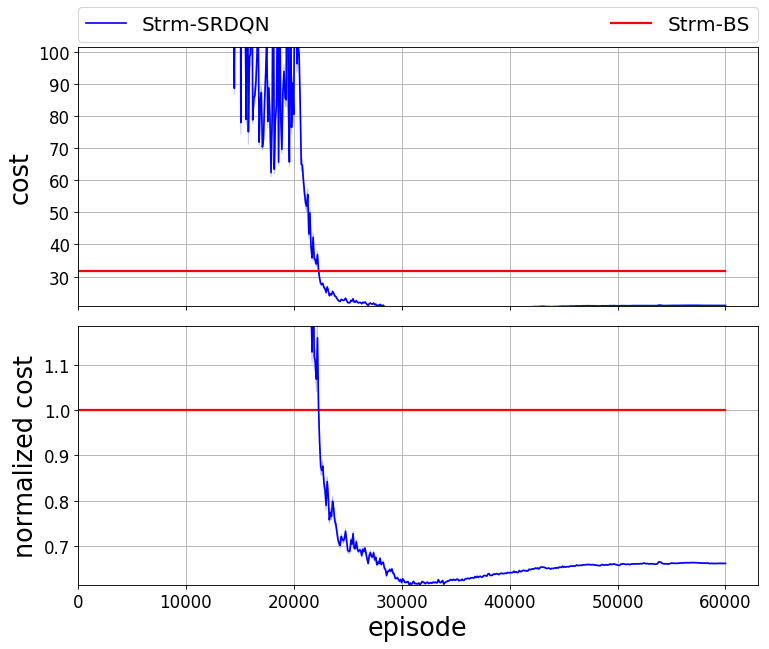}
		\caption{Case 2-1-3}
		\label{fig:agent_case6_8-3-3}		
	\end{subfigure}	
	\begin{subfigure}{0.24\textwidth}
		\centering
		\includegraphics[scale=0.16]{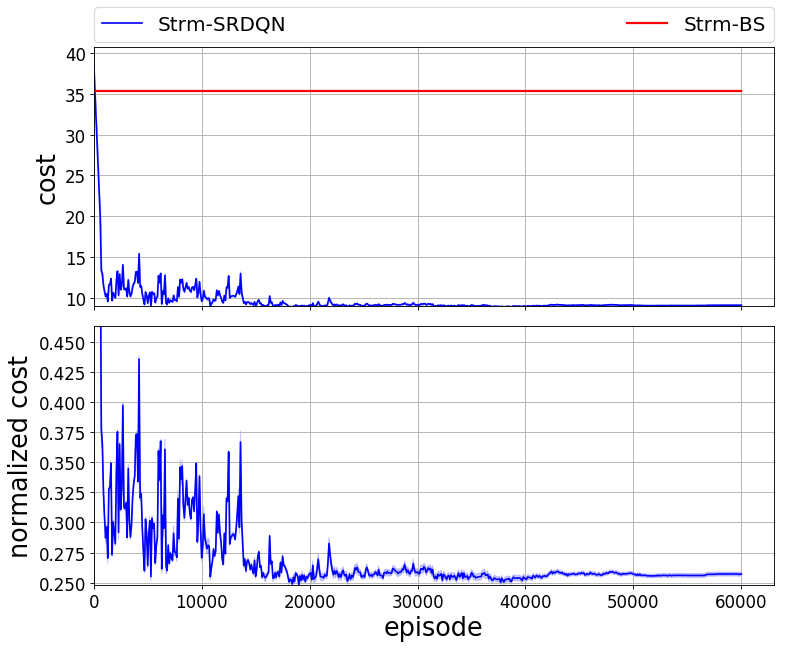}
		\caption{Case 3-1-1}
		\label{fig:agent_case6_9-3-1}		
	\end{subfigure}		
	\begin{subfigure}{0.24\textwidth}
		\centering
		\includegraphics[scale=0.16]{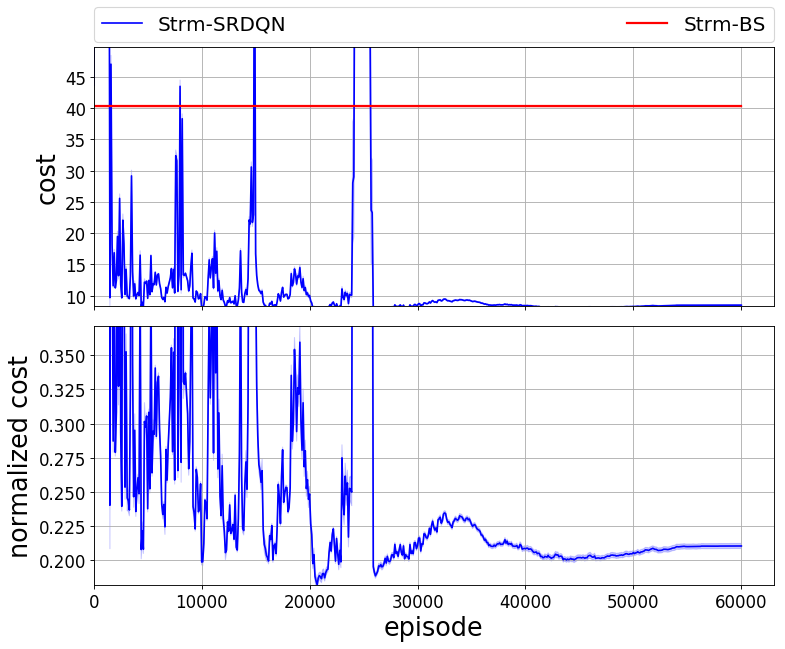}
		\caption{Case 4-1-1}
		\label{fig:agent_case6_10-3-1}		
	\end{subfigure}		
	{}
\end{figure}

\begin{figure}	
	\centering
	\caption{Results of transfer learning for agents with $|\mathcal{A}_1| \neq |\mathcal{A}_2|, (c^j_{p_1}, c^j_{h_1}) \neq (c^j_{p_2}, c^j_{h_2}), D_1 \neq D_2$, and $\pi_1 \neq \pi_2$. \label{fig:case_6_random}}	
	\vspace{-5pt}			
	\begin{subfigure}{0.24\textwidth}
		\centering
		\includegraphics[scale=0.16]{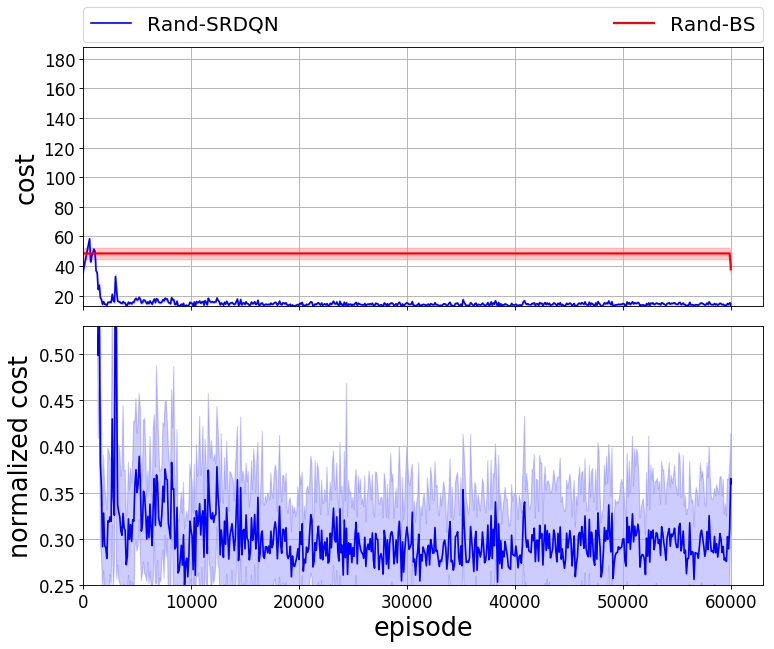}
		\caption{Case 1-2-1}
		\label{fig:agent_case6_11-4-1}		
	\end{subfigure}
	\begin{subfigure}{0.24\textwidth}
		\centering
		\includegraphics[scale=0.16]{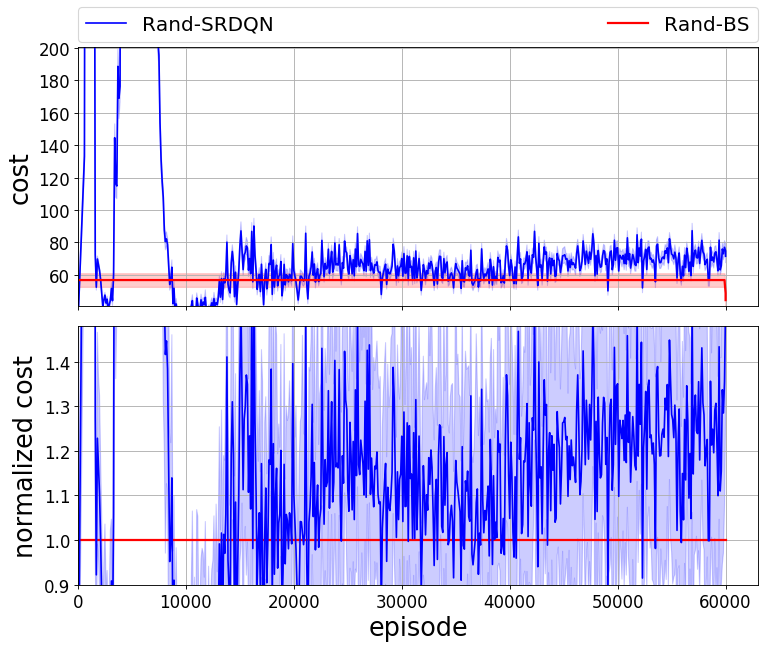}
		\caption{Case 2-1-2}
		\label{fig:agent_case6_12-3-2}		
	\end{subfigure}	
	\begin{subfigure}{0.24\textwidth}
		\centering
		\includegraphics[scale=0.16]{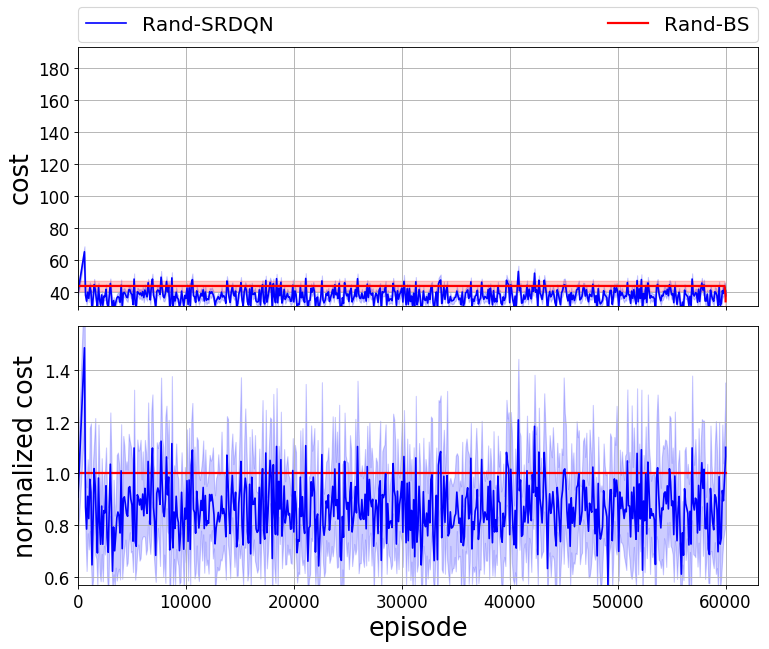}
		\caption{Case 3-3-3}
		\label{fig:agent_case6_13-6-3}		
	\end{subfigure}		
	\begin{subfigure}{0.24\textwidth}
		\centering
		\includegraphics[scale=0.16]{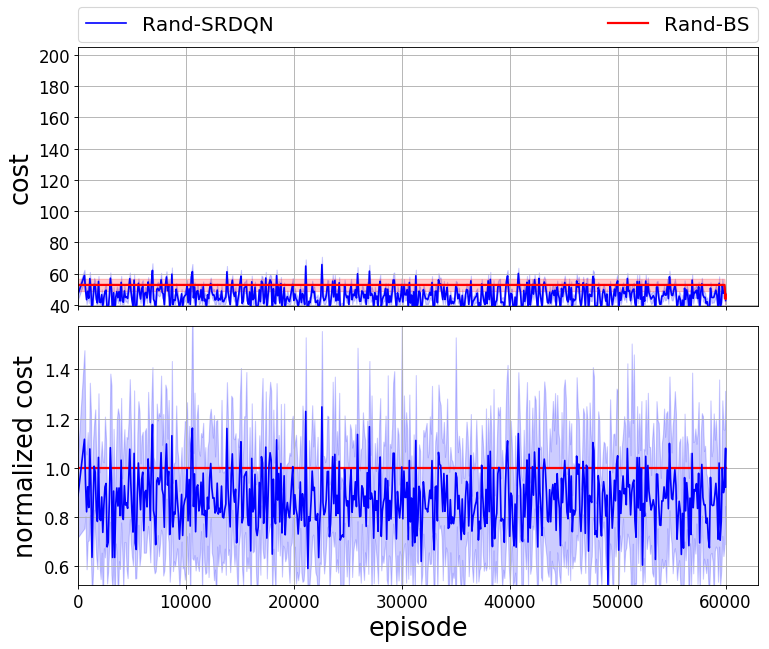}
		\caption{Case 4-1-1}
		\label{fig:agent_case6_14-3-3}		
	\end{subfigure}		
	{}
\end{figure}

\section{Pseudocode of the Beer Game Simulator}\label{sec:appnd:sudocode}

The SRDQN algorithm needs to interact with the environment, so that  for each state and action, the environment should return the reward and the next state. 
We simulate the beer game environment using Algorithm \ref{alg01}. In addition to the notation defined earlier, the algorithm also uses the following notation:\\
$d^{t}$: The demand of the customer in period $t$. \\
$OS_i^t$: Outbound shipment from agent $i$ (to agent $i-1$) in period $t$. \\

{\SingleSpacedXI
	\begin{algorithm}[ht]
		\caption{Beer Game Simulator Pseudocode.}
		\label{alg01}
		\begin{algorithmic}[1]
			\Procedure{playGame} {}
			\State Set $T$ randomly, and $t = 0$, Initialize $IL_i^0$ for all agents, $AO_i^t = 0, AS_i^t = 0, \forall i,t$
			\While{$t \le T$}
			\State $AO_i^{t+l_i^{fi}} += d^{t}$ \Comment {{ \color{purple} set the retailer's arriving order to external demand}}
			\For{$i=1 : 4$} \Comment {{ \color{purple} loop through stages downstream to upstream}}
			\State get action $a_i^t$ \Comment {{ \color{purple} choose order quantity}}	
			\State $OO_i^{t+1} = OO_i^t + a_i^t$ \Comment {{\color{purple} update $OO_i$}}	
			\State $AO_{i+1}^{t+l_i^{fi}} += a_{i}^t$ \Comment {{\color{purple} propagate order upstream}}	
			\EndFor	
			\State $AS_4^{t+l_4^{tr}} += a_4^{t}$ \Comment {{ \color{purple} set manufacturer's arriving shipment to its order quantity}}	
			\For{$i=4 : 1$} \Comment {{ \color{purple} loop through stages upstream to downstream}}	
			\State $IL_i^{t+1} = IL_i^{t} + AS_i^t$ \Comment {{\color{purple} receive inbound shipment}}	
			\State $OO_i^{t+1} -= AS_i^t$ \Comment {{\color{purple} update $OO_i$}}	
			\State current\_Inv = $\max \{0, IL_i^{t+1}\}$ \Comment {{\color{purple} determine outbound shipment}}	
			\State current\_BackOrder = $\max \{0, -IL_i^{t}\}$		
			\State $OS_i^t = \min \{$ current\_Inv, current\_BackOrder + $AO_i^{t}$ $\}$
			\State $AS_{i-1}^{t+l_i^{tr}} += OS_i^{t}$	\Comment {{ \color{purple} propagate order downstream}}	
			\State $IL_{i}^{t+1} -= AO_i^{t}$ \Comment {{\color{purple} update $IL_i$}}	
			\State $c_i^t = c_i^p \max\{-IL_{i}^{t+1}, 0\} + c_i^h \max\{IL_{i}^{t+1}, 0 \}$ \Comment {{\color{purple} calculate cost}}	
			\EndFor
			\State $t += 1$
			\EndWhile
			\EndProcedure
		\end{algorithmic}
	\end{algorithm}
} 

\end{APPENDICES}

\end{document}